%% file: arxiv.tex
\pdfoutput=1
\documentclass{article}

    \PassOptionsToPackage{numbers, sort, compress}{natbib}

\usepackage[preprint]{neurips_2025}

\usepackage[utf8]{inputenc} 
\usepackage[T1]{fontenc}    
\usepackage{url}            
\usepackage{booktabs}       
\usepackage{amsfonts}       
\usepackage{nicefrac}       
\usepackage{microtype}      
\usepackage{xcolor}         
\definecolor{linkblue}{HTML}{6870ae}

\usepackage{amsmath}
\usepackage[pagebackref=true,colorlinks,citecolor=brown]{hyperref}
\hypersetup{
  urlcolor  = linkblue
}
\usepackage{multirow} 
\usepackage{makecell} 
\usepackage{wrapfig}
\usepackage[inkscape=false]{svg} 
\usepackage{cleveref}
\usepackage{caption}
\usepackage{fontawesome5}

\usepackage{graphicx}
\usepackage{float}
\floatstyle{plaintop}
\restylefloat{table}
\usepackage[export]{adjustbox}

\usepackage{amsthm}
\usepackage{subcaption}
\usepackage{etoolbox}

\usepackage[ruled,vlined,linesnumbered,noend]{algorithm2e}
\SetKwInput{Inputs}{\textbf{Inputs}}
\SetKwInput{Hparams}{\textbf{Hparams}}

\usepackage{listings}
\definecolor{psGreen}{rgb}{0,0.6,0}
\definecolor{psGray}{rgb}{0.5,0.5,0.5}
\definecolor{psPurple}{rgb}{0.58,0,0.82}
\definecolor{psBackgroundColour}{rgb}{0.95,0.95,0.92}

    \lstdefinestyle{mystyle}{
      backgroundcolor=\color{psBackgroundColour},   
      commentstyle=\color{psGreen},
      keywordstyle=\color{magenta},
      numberstyle=\tiny\color{psGray},
      stringstyle=\color{psPurple},
      basicstyle=\ttfamily\footnotesize,
      breakatwhitespace=false,         
      breaklines=true,                 
      captionpos=b,                    
      keepspaces=true,                 
      numbersep=5pt,                  
      showspaces=false,                
      showstringspaces=false,
      showtabs=false,                  
      tabsize=2,
    }

\theoremstyle{definition}
\newtheorem{definition}{Definition}[section]

\title{Energy-Based Transformers are Scalable Learners and Thinkers}

\newcommand{\pa}{EBT} 
\newcommand{\parch}{EBT} 

\newcounter{facet}                    
\renewcommand{\thefacet}{\arabic{facet}}  
\newcommand{\facet}[2][]{%
  \refstepcounter{facet}
  \textbf{Facet \thefacet: #2.}
  \ifx&#1&\else\label{#1}\fi            
}

\usepackage{pifont}
\newcommand{\cmark}{\textcolor{green}{\ding{51}}}%
\newcommand{\xmark}{\textcolor{red}{\ding{55}}}%

\usepackage[draft,textsize=footnotesize,textwidth=15mm]{todonotes}

\NewDocumentCommand{\heng}
{ mO{} }{\textcolor{red}{\textsuperscript{\textit{Heng}}\textsf{\textbf{\small[#1]}}}}

\NewDocumentCommand{\yilun}
{ mO{} }{\textcolor{red}{\textsuperscript{\textit{Yilun}}\textsf{\textbf{\small[#1]}}}}

\NewDocumentCommand{\ember}
{ mO{} }{\textcolor{purple}{\textsuperscript{\textit{Ember}}\textsf{\textbf{\small[#1]}}}}

\NewDocumentCommand{\carl}
{ mO{} }{\textcolor{blue}{\textsuperscript{\textit{Carl}}\textsf{\textbf{\small[#1]}}}}

\newcommand{\blfootnote}[1]{%
  \begingroup
    \renewcommand\thefootnote{}
    \footnote{#1}%
    \addtocounter{footnote}{-1}
  \endgroup}

\author{%
  \textbf{Alexi Gladstone}\textsuperscript{1,2},
  \textbf{Ganesh Nanduru}\textsuperscript{1},
  \textbf{Md Mofijul Islam}\textsuperscript{1,3},
  \textbf{Peixuan Han}\textsuperscript{2}, \\
  \textbf{Hyeonjeong Ha}\textsuperscript{2},
  \textbf{Aman Chadha}\textsuperscript{3,4},
  \textbf{Yilun Du}\textsuperscript{5},
  \textbf{Heng Ji}\textsuperscript{3},
  \textbf{Jundong Li}\textsuperscript{1},
  \textbf{Tariq Iqbal}\textsuperscript{1} \\[0.6em]
  \textsuperscript{1}UVA \quad
  \textsuperscript{2}UIUC \quad
  \textsuperscript{3}Amazon GenAI\textsuperscript{$\dagger$} \quad
  \textsuperscript{4}Stanford University \quad
  \textsuperscript{5}Harvard University \\[0.6em]
  \textbf{\href{https://energy-based-transformers.github.io}
     {\faGlobe\enspace{energy-based-transformers.github.io}}
    \quad
  \href{https://github.com/alexiglad/ebt}
     {\faGithub\enspace{github.com/alexiglad/EBT}}}
}

\begin{document}

\maketitle
\blfootnote{Correspondence to Alexi Gladstone:  \href{mailto:alexigladstone@gmail.com}{\faEnvelope\enspace{alexigladstone@gmail.com}}. \textsuperscript{$\dagger$}Work does not relate to position at Amazon.}

\input{arxiv/main_sec/0_abstract}   
\input{arxiv/main_sec/1_intro}

\input{arxiv/main_sec/2_intuition}
\input{arxiv/main_sec/4_approach}
\input{arxiv/main_sec/5_experimentation}

\input{arxiv/main_sec/7_discussion}

\input{arxiv/main_sec/3_related_work}

\input{arxiv/main_sec/8_conclusion}
\input{arxiv/main_sec/9_acknowledgements}


{
    \small
    \bibliographystyle{unsrtnat}
    \bibliography{references}
}

\appendix

\renewcommand{\thesection}{\Alph{section}}          
\renewcommand{\thefigure}{\Alph{section}\arabic{figure}} 
\renewcommand{\thetable}{\Alph{section}\arabic{table}}   

\setcounter{section}{0}
\counterwithin{figure}{section}   
\counterwithin{table}{section}

\clearpage
\input{arxiv/supp_sec/_Appendix_Description}
\input{arxiv/supp_sec/A_broader_impact_future_work}

\input{arxiv/supp_sec/B_additional_experiments}

\input{arxiv/supp_sec/C_Approach_Details}
\input{arxiv/supp_sec/D_Experimental_Details}

\input{arxiv/supp_sec/E_Additional_Related_Works}
\input{arxiv/supp_sec/F_Additional_Facets}

\input{arxiv/supp_sec/G_Counterarguments}
\input{arxiv/supp_sec/H_EBM_Intro}
\input{arxiv/supp_sec/I_How_to_Train}


\end{document}

%% file: arxiv/main_sec/0_abstract.tex
\vspace{-25pt}

\begin{abstract}
\label{sec:abstract}

Inference-time computation techniques, analogous to human System 2 Thinking, have recently become popular for improving model performances. 
However, most existing approaches suffer from several limitations: they are modality-specific (e.g., working only in text), problem-specific (e.g., verifiable domains like math and coding), or require additional supervision/training on top of unsupervised pretraining (e.g., verifiers or verifiable rewards). In this paper, we ask the question \textit{``Is it possible to \textbf{generalize} these System 2 Thinking approaches, and develop models that learn to think solely from unsupervised learning?}'' Interestingly, we find the answer is \textbf{yes}, by learning to explicitly \textbf{verify} the compatibility between inputs and candidate-predictions, and then re-framing prediction problems as optimization with respect to this verifier. 
Specifically, we train \textbf{Energy-Based Transformers (EBTs)}---a new class of Energy-Based Models (EBMs)---to assign an \textbf{energy} (unnormalized probability) value to every input and candidate-prediction pair, enabling predictions through gradient descent-based energy minimization until convergence. 
This formulation enables System 2 Thinking to emerge from unsupervised learning, making it modality and problem agnostic. 
Across both discrete (text) and continuous (visual) modalities, we find EBTs scale faster than the dominant Transformer++ approach during training, achieving an up to 35\% higher scaling rate with respect to data, batch size, parameters, FLOPs, and depth. 
During inference, EBTs improve performance with System 2 Thinking (i.e., extra computation) by 29\% more than the Transformer++ on language tasks, and EBTs outperform Diffusion Transformers on image denoising while using fewer forward passes. Further, we find that System 2 Thinking with EBTs yields larger performance improvements on data that is farther out-of-distribution, and that EBTs achieve better results than existing models on most downstream tasks given the same or worse pretraining performance, suggesting that EBTs generalize better than existing approaches.
Consequently, EBTs are a promising new paradigm for scaling both the \textbf{learning} and \textbf{thinking} capabilities of models.

\end{abstract}

%% file: arxiv/main_sec/1_intro.tex
\section{Introduction}
\label{sec:intro}

In psychology, human thinking is often classified into two different types: System 1 (thinking fast) and System 2 (thinking slow)~\cite{kahneman2011thinking, evans2011dual, kahneman2002representativeness, frankish2010dual}. System 1 thinking is characterized by quick, intuitive and automatic responses, relying on previous experience to solve simple or familiar problems. Alternatively, System 2 Thinking is slow, deliberate and analytical, requiring conscious effort and logical reasoning to process more complex information. System 2 Thinking is essential for complex problems that go beyond automatic pattern recognition, such as in mathematics, programming, multistep reasoning, or novel out-of-distribution situations~\cite{neys2006dual, goel2000dissociation}, where precision and depth of understanding are important. Although current models perform well on tasks suitable for System 1 thinking~\cite{li2025system}, they continue to struggle with tasks that demand System 2 capabilities~\cite{mirzadeh2024gsm, yan2025phd, EscapeBench2025}.

Consequently, the recent pursuance of System 2 Thinking capabilities has become a growing interest among AI researchers, leading to the rise of several foundation models such as O1~\cite{jaech2024openai}, R1~\cite{guo2025deepseek}, Grok3~\cite{xai2025grok3}, and Claude 3.7 Sonnet~\cite{anthropic2025claude37sonnet}. These ``reasoning models'' excel on math and coding benchmarks, particularly by increasing the time models spend thinking. However, publicly available information on training methods, particularly from the open-source R1 model~\cite{guo2025deepseek}, suggests that the Reinforcement Learning (RL) based training approach for these models only works in domains where rule-based rewards can easily verify answers, such as math and coding. This limitation reduces applicability to a small range of problem types, and as a consequence, often deteriorates performance in other tasks such as writing~\cite{openai2024learning, su2025expanding, illusionofthinking}. Further, recent evidence suggests this RL-based approach may not induce new reasoning patterns, but rather just increase the probability of reasoning patterns already known to the base model~\cite{yue2025does}, which limits performance on problems requiring exploration.

Along similar lines, there has been strong interest in achieving System 2 Thinking in both diffusion models and Recurrent Neural Networks (RNNs). Diffusion models support iterative inference through denoising steps, where increasing denoising steps can improve performance. However, they typically fail to benefit from denoising steps beyond what they were trained on~\cite{ma2025inference}, and aside from increasing denoising steps, diffusion models require an external verifier to improve System 2 Thinking capabilities~\cite{ma2025inference, liu2025video, singhal2025general}. RNNs also offer iterative computation via recurrent state updates~\cite{gu2023mamba, peng2023rwkv, hochreiter1997long}, however, most modern RNNs only update their internal state with new information, meaning they cannot be used for thinking longer. Additionally, RNNs that do support recurrent depth still lack mechanisms for explicit verification~\cite{geiping2025scaling}, resulting in limited adoption for System 2 Thinking.

\begin{figure}[!t]
    \centering
    \begin{subfigure}[t]{0.24\textwidth}
        \centering
        \includegraphics[height=3.5cm,keepaspectratio]{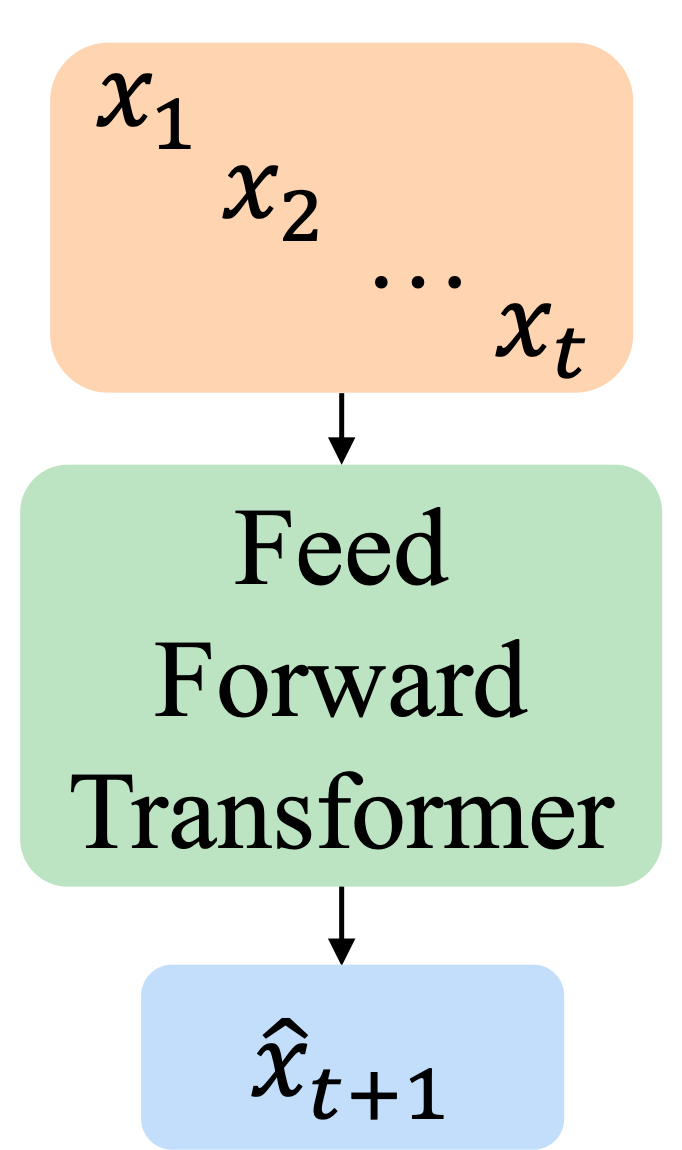}
        \caption{AR Transformer}\label{fig:transformer_arch}
        \label{subfig:ar_transformer}
    \end{subfigure}%
    \hfill
    \begin{subfigure}[t]{0.24\textwidth}
        \centering
        \includegraphics[height=3.5cm,keepaspectratio]{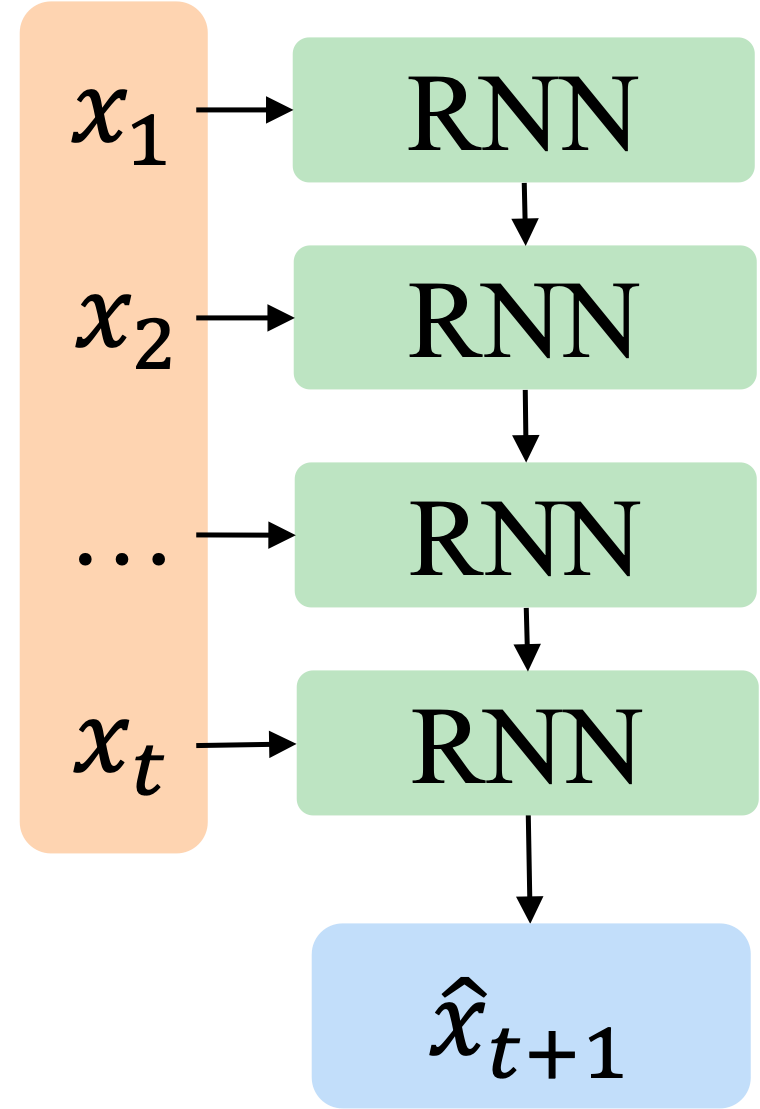}
        \caption{RNN}
        \label{subfig:rnn_arch}
    \end{subfigure}%
    \hfill
    \begin{subfigure}[t]{0.24\textwidth}
        \centering
        \includegraphics[height=3.5cm,keepaspectratio]{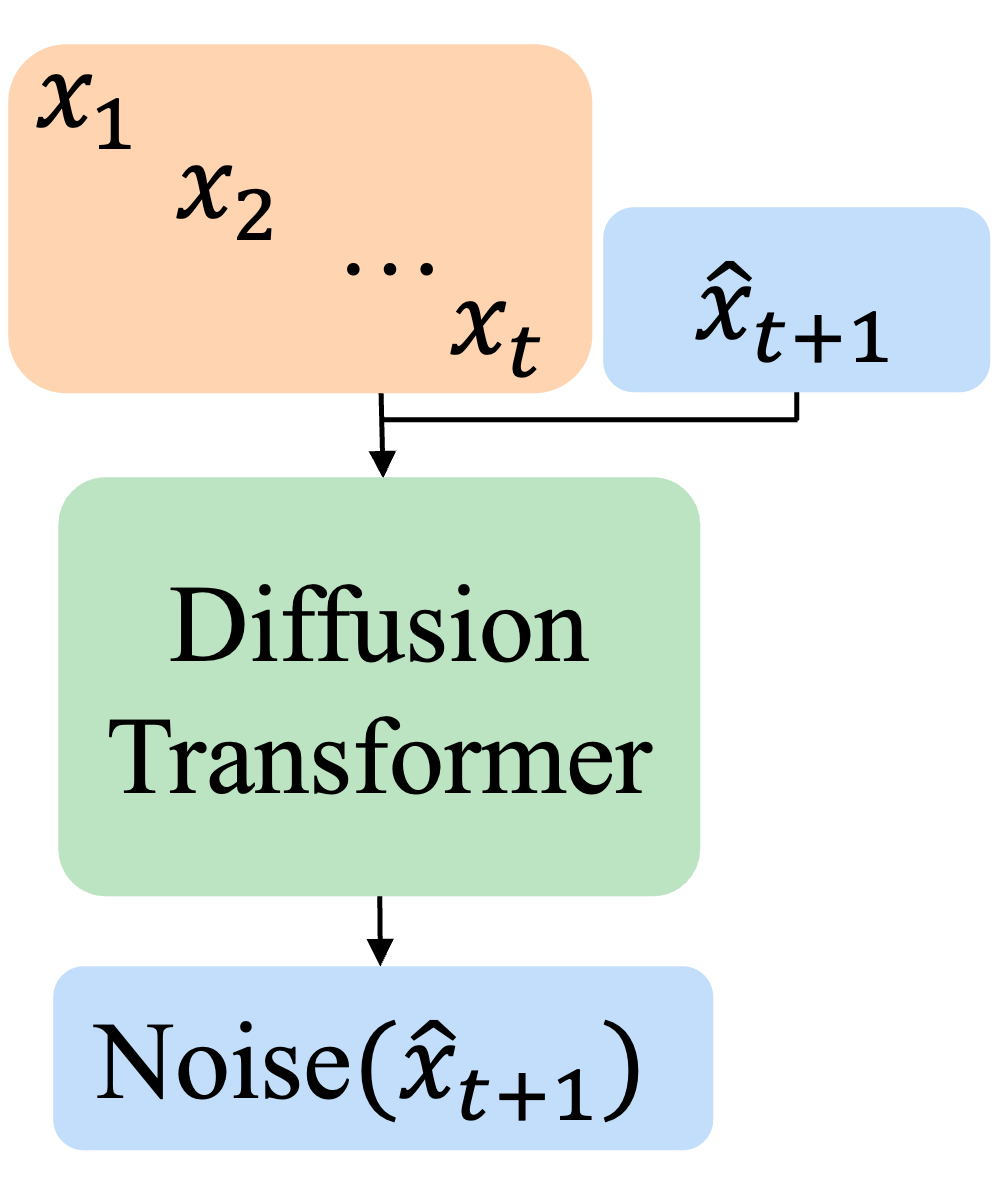}
        \caption{Diffusion Transformer}\label{fig:diffusion_arch}
        \label{subfig:diffusion_arch}
    \end{subfigure}%
    \hfill
    \begin{subfigure}[t]{0.24\textwidth}
        \centering
        \includegraphics[height=3.5cm,keepaspectratio]{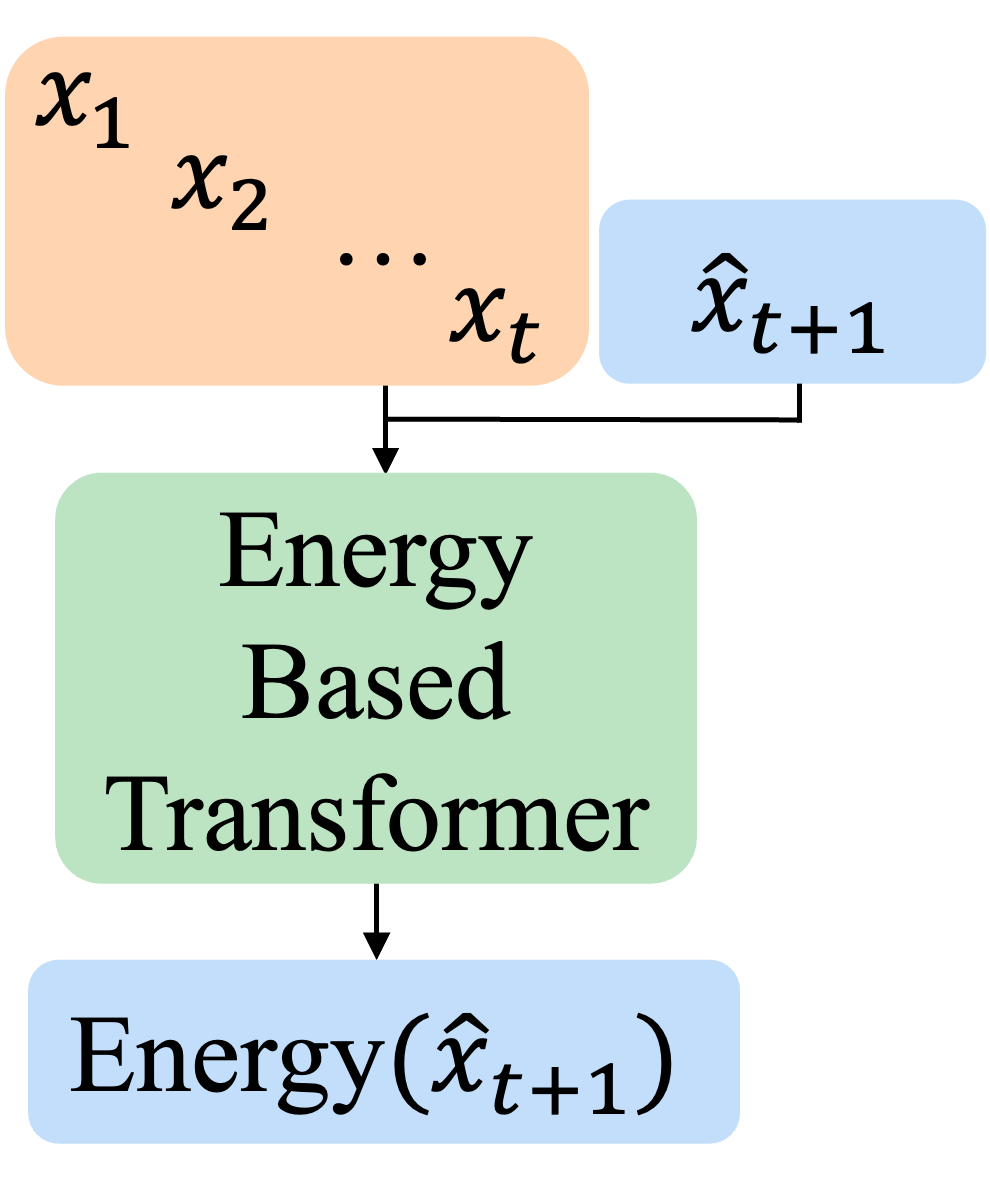}
        \caption{\pa}\label{fig:ebwm_arch}
    \end{subfigure}

    \begin{flushleft}
      \hspace*{0.02\textwidth}
      \begin{minipage}{.725\textwidth}
        \centering
        \rule{\textwidth}{0.4pt}\\[-3pt]
        Existing Autoregressive Approaches
      \end{minipage}
    \end{flushleft}
    \vspace{-5pt}

    \caption{\textbf{Autoregressive Architecture Comparison.} (a) Autoregressive (AR) Transformer is the most common, with (b) RNNs becoming more popular recently~\cite{gu2023mamba, peng2023rwkv}. (c) Diffusion Transformers~\cite{peebles2023scalable} (which are often bidirectional but can also be autoregressive) are the most similar to {\pa}, being able to dynamically allocate computation during inference, but predict the noise rather than the energy~\cite{li2025autoregressive, deng2024causal}. Consequently, diffusion models cannot give unnormalized likelihood estimates at each step of the thinking process, and are not trained as explicit verifiers, unlike {\pa}s.}
    \label{fig:model_comparison}
    \vspace{-20pt}
\end{figure}

As one of the primary goals of AI is to figure out how we can create systems that \textbf{learn to think on their own} on any type of problem, these approaches ultimately bring about the following core research question: ``\textit{Can we \textbf{rely entirely} on \textbf{unsupervised learning} to develop \textbf{System 2 Thinking?}}'' Such a capability would enable \textit{generalization} of current System 2 Thinking approaches to \textit{any problem}, \textit{any modality}, and \textit{avoid the reliance on external human, reward, or model supervision}.

In this work, we argue and demonstrate empirically that the answer to this core research question is \textbf{yes}, but that there are several limitations in existing model architectures that prevent this type of unsupervised System 2 Thinking from emerging. Particularly, when comparing the qualities of human System 2 Thinking with the current modeling paradigms (Figure~\ref{fig:model_comparison},  Table~\ref{tab:cognitive_facets}), we observe several key differences, outlined below as three key Facets of System 2 Thinking\footnote{We acknowledge that these are not comprehensive for achieving System 2 Thinking, but rather, a good first step (more info is in Section~\ref{sec:thinking_formalization}).}: \\

\begin{table*}[t]
\centering
\caption{\textbf{Architectures and Cognitive Facets.} For each prediction, Feed Forward (FF) Transformers and RNNs generally\protect\footnotemark~have a finite amount of computation. While diffusion models have potentially more computation during inference by increasing the number of denoising steps, they do not learn to explicitly verify or estimate uncertainty for predictions. EBMs can use a dynamic amount of computation during inference by iterating for any number of steps, and give an energy scalar that can be used to evaluate uncertainty and verify the strength of predictions.}
\small
	\begin{tabular}{lccc}
            \toprule
            \multirow{2}{*}{\textbf{Architecture}} & \multirow{2}{*}{\makecell{\textbf{Dynamic Compute} \\ \textbf{Allocation (Facet~\ref{facet:dynamic-computation})}}} & \multirow{2}{*}{\makecell{\textbf{Modeling} \\ \textbf{Uncertainty (Facet~\ref{facet:prediction-uncertainty})}}} & \multirow{2}{*}{\makecell{\textbf{Prediction} \\ \textbf{Verification (Facet~\ref{facet:prediction-verification})}}} \\
            & & & \\
            \toprule
            FF Transformers & \xmark & \xmark & \xmark \\
            RNNs & \xmark & \xmark & \xmark \\
            Diffusion Transformers & \cmark & \xmark & \xmark \\
            {\pa}s & \cmark & \cmark & \cmark \\
            \bottomrule
        \end{tabular}
 \label{tab:cognitive_facets}
 \vspace{-10pt}
\end{table*}
\footnotetext{Recent works attempt to enable dynamic computation per prediction~\cite{hao2024training, geiping2025scaling, goyal2023think}, but these approaches are generally not modality agnostic and have not been widely adopted. While RNNs can theoretically support dynamic computation, they are typically updated only with new state information, except in specialized cases such as in~\cite{geiping2025scaling}.}

\facet[facet:dynamic-computation]{Dynamic Allocation of Computation}
Humans naturally allocate varying amounts of effort to different tasks depending on difficulty, which is widely supported by psychology and neuroscience~\cite{kahneman2011thinking, ditterich2006evidence, rougier2005prefrontal}.
As the difficulty of tasks humans face varies widely, the ability to adjust the magnitude of computational resources allocated towards a task is fundamental to success.\footnote{It's important to note that we refer to this dynamic computation capability at the granularity of \textbf{each} prediction being made, meaning current LLMs built with traditional AR transformers/RNNs cannot dynamically allocate compute per token generated as they have a finite depth/width and are only updated with new text tokens. Please check Section~\ref{sec:counterarguments} for more information.} For example, a decision regarding whether to change careers generally takes people much more time than deciding what to eat for lunch.

\facet[facet:prediction-uncertainty]{Modeling Uncertainty in Continuous State Spaces} 
While thinking longer is important for improving performance, humans also weigh how uncertain they are before committing to a decision. In language, LLMs can simulate this through token-level probabilities~\cite{tomani2024uncertainty}.
In the context of continuous state spaces, such as in vision, without the usage of discretization schemes such as Vector Quantization~\cite{van2017neural} or pseudo losses/objectives (such as ELBO~\cite{kingma2013auto}), standard implementations of the most successfully used approaches with Transformers, RNNs, or Diffusion models generally do not provide strong or reliable uncertainty estimates~\cite{sankararaman2022bayesformer, heng2024out, nalisnick2018deep, serra2019input}.\footnote{We acknowledge that there are approaches to achieve uncertainty with the models discussed, such as Mixture Density Networks~\cite{bishop1994mixture} as well as score-based diffusion models~\cite{song2020score}. However, these approaches have seen less widespread success and scalability than the current dominant approaches.} EBMs can naturally model uncertainty without having to model exact likelihoods by modeling the relative unnormalized likelihoods of predictions~\cite{dawid2024introduction}, as demonstrated in Figure~\ref{fig:energy_landscape_minimization}. As the real world often contains many inherently unpredictable elements, for instance, when a pedestrian might emerge from behind a parked vehicle, the ability to express uncertainty in predictions is essential to being cautious, and is a natural capability of humans~\cite{Peters2017Uncertainty, Vilares2012Differential, Sarinopoulos2010Uncertainty}.

\facet[facet:prediction-verification]{Verification of Predictions} 
In addition to allocating computation and modeling uncertainty, effective thinking also benefits from the ability to verify predictions, which can guide decisions about when to stop thinking or to select the most accurate predictions.
There is strong evidence that such verification is a core component of human thinking~\cite{loesche2018paving, alkouri2016using}. In fact, it can be shown that verifying solutions is exponentially easier than generating solutions~\cite{du2022learning}, which means that verifiers can often generalize better than explicit generators~\cite{du2022learning}. Training an explicit verifier means that at each step of the thinking process an estimate of the current prediction's quality can be extracted. This supports more dynamic inference time behavior such as early stopping when a prediction is known to be correct or allocating more compute when a problem is difficult, which can intuitively be seen as adjusting resources according to the difficulty of a problem~\cite{kahneman2011thinking}. Training an explicit verifier also allows for capabilities such as Monte Carlo Tree Search~\cite{silver2017mastering} or sampling many times and choosing the best prediction, which traditionally have involved training new models on more data~\cite{lightman2023let, ma2025inference}.

For more information on additional Facets, please refer to Section~\ref{sec:cognitive_facet_details}.

\begin{figure}[t]
    \centering
    \includegraphics[width=\textwidth]{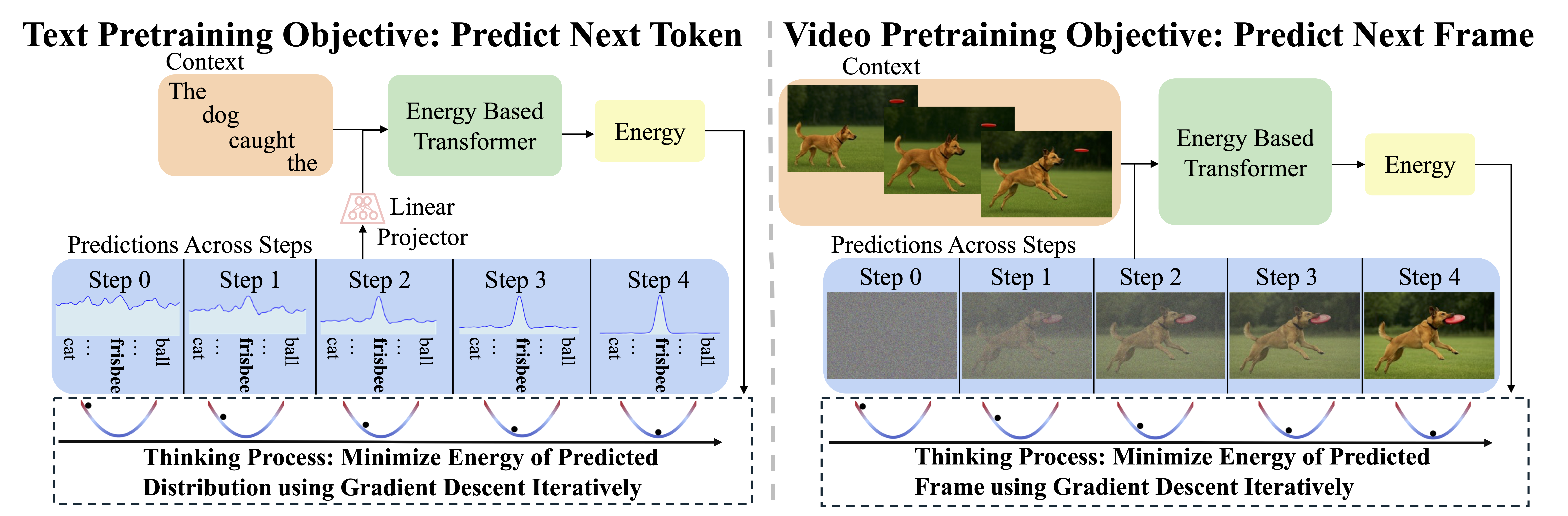}
    \caption{\textbf{{\pa} for Autoregressive Modeling.} Each blue box corresponds to a different prediction based on the current step of the thinking process, where the initial prediction starts as random. At each step, a new prediction is fed into the model, which gives an energy scalar for the prediction's current \textit{compatibility} (unnormalized likelihood) with the context (Facets~\ref{facet:prediction-uncertainty} and~\ref{facet:prediction-verification}). Then, the gradient of this energy with respect to the prediction is calculated and used to update the prediction. This gradient descent update is done iteratively to refine the prediction until convergence of the predicted energy, which allows for dynamic use of computation (Facet~\ref{facet:dynamic-computation}).}
    \label{fig:model_architecture}
    \vspace{-10pt}
\end{figure}

To achieve all facets described, we propose viewing thinking as an optimization procedure with respect to a learned verifier, which evaluates the compatibility (unnormalized probability) between an input and candidate prediction (Figure~\ref{fig:model_architecture}). More concretely, we train Energy-Based Models (EBMs) to learn an energy (unnormalized probability) landscape over all possible input-prediction pairs, where lower energy indicates higher compatibility. Then, thinking corresponds to starting from an initial random prediction, and progressively refining it through energy minimization until convergence (visualized in Figure~\ref{fig:energy_landscape_minimization}). Optimizing over a learned energy landscape naturally allows for dynamic compute allocation (Facet~\ref{facet:dynamic-computation}), allowing models to think for longer on harder problems by iteratively performing more steps. Additionally, at each step of this thinking process, EBMs act as verifiers in the forward pass, giving an energy scalar which represents the \textit{compatibility} of the context with the prediction. This energy scalar directly addresses Facets~\ref{facet:prediction-uncertainty} and~\ref{facet:prediction-verification} by serving as an unnormalized likelihood as well as the score that verifies whether a prediction is correct (Figure~\ref{fig:energy_landscape_minimization}).

While viewing thinking through the lens of inference-time energy minimization is a promising perspective, EBMs have traditionally struggled with scalability~\cite{du2019implicit}, with zero publicly known Foundation EBMs. This can be attributed to issues with EBM training stability~\cite{du2019implicit, du2020improved, li2023learning, arbel2020generalized} and long training times~\cite{li2023learning, arbel2020generalized}. To address these challenges and establish a scalable EBM training paradigm, we introduce Energy-Based Transformers (EBTs), Transformer implementations specifically for EBMs. We release two variants: a decoder-only EBT inspired by the GPT architecture~\cite{radford2018improving}, which parallelizes all predictions simultaneously; and a bidirectional EBT, enabling bidirectional attention across entire sequences, similar to BERT~\cite{devlin2019bert} and Diffusion Transformers (DiT)~\cite{peebles2023scalable}. 

Additionally, we identify practical enhancements that improve the training efficiency of EBTs and provide theoretical insights into why our optimization-based training approach achieves strong scalability. Finally, we introduce \textit{energy landscape regularization} techniques, which are methods that enhance the convexity and smoothness of learned energy landscapes, thereby fostering the emergence of strong System 2 Thinking capabilities during pretraining.

To investigate the learning and thinking scalability of EBTs, we compare EBTs to the Transformer++ for autoregressive modeling and the Diffusion Transformer (DiT) for bidirectional modeling, across both discrete and continuous modalities. During pretraining, we find that EBTs achieve an up to $35\%$ higher scaling rate than the Transformer++ across several axes, including data, batch size, parameters, FLOPs, and depth. Notably, EBTs are the first approach to out-scale the standard feed-forward Transformer++ recipe across several modalities and axes, including improved data efficiency. At inference, EBTs outperform existing paradigms on System 2 Thinking, or solving more challenging problems with additional computation. For example, EBTs can improve language modeling performance $29\%$ more than the Transformer++, and for image denoising EBTs exhibit a higher performance-compute scaling rate and improved performance over DiTs with $99\%$ fewer forward passes. In investigations of the effect of System 2 Thinking on generalization, we consistently observe two phenomena. First, with the same or worse pretraining performance, {\pa}s still outperform existing models at inference by performing System 2 Thinking, demonstrating its importance and how EBTs often generalize better than existing models. Second, System 2 Thinking with EBTs yields more substantial performance gains on data that is further out-of-distribution, aligning with observations in psychology where humans use System 2 Thinking on challenging, unseen tasks. 

We believe the EBT implementations, along with novel techniques for EBMs to maximize the learning and thinking scalability, will advance the EBM paradigm by addressing key challenges in stable, parallelizable, and efficient training.

%% file: arxiv/main_sec/2_intuition.tex
\section{Energy-Based Transformers ({\pa}) Intuition}

\label{sec:intuition}
Energy-Based Models (EBMs) learn an energy function that assigns a scalar value to each input configuration, with lower energy indicating higher compatibility or likelihood between input variables~\cite{du2019implicit}, and high energy indicating lower compatibility or likelihood. Accordingly, the energy function acts as a \textbf{verifier} of input data coherence. Leveraging this principle, Energy-Based Transformers ({\pa}s) are trained to \textbf{verify} the compatibility of a given context-prediction pair (give an energy value), and then make predictions by optimizing with respect to this verifier (minimizing the energy), which is shown in Figure~\ref{fig:model_architecture}. Starting from an initial prediction, such as random noise, {\pa}s iteratively refine their output by progressively minimizing the energy, ultimately converging to predictions that are consistent with the given context. Performing energy minimization through this process simulates the thinking process during pretraining, unlike with traditional models, enabling each prediction (e.g., a token for LLMs) to have its own thinking process.

\begin{figure}[t]
    \centering
    \includegraphics[width=.75\textwidth]{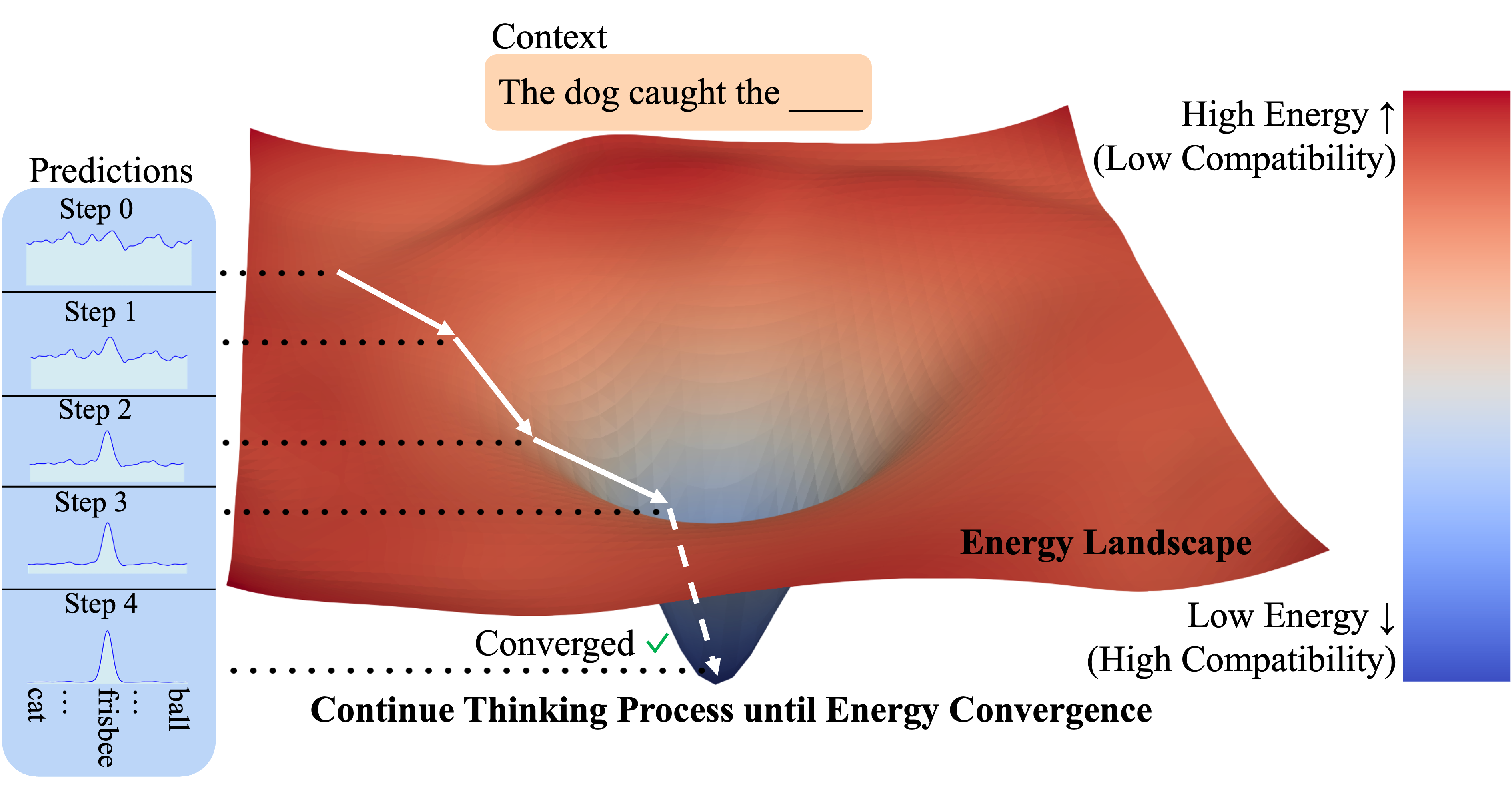}
    \caption{\textbf{Thinking Process Visualization.} A learned energy landscape and its optimization through gradient descent, interpreted as a thinking process. In this example, the model predicts a distribution over text tokens, progressively shifting from an initial random distribution toward the target distribution. At each step, the EBM assigns an energy scalar indicating how \textbf{compatible} the current prediction is with the context, visualized as the landscape's height (Facet~\ref{facet:prediction-verification}). This scalar's convergence allows the model to determine whether the prediction is adequate or if further thinking is necessary. Uncertainty (Facet~\ref{facet:prediction-uncertainty}) can be represented by landscapes that are harder to optimize or by landscapes with many local minima, allowing the model to know when it requires more steps to think (Facet~\ref{facet:dynamic-computation}). Adapted from~\cite{li2018visualizing}.} 
    \label{fig:energy_landscape_minimization}
    \vspace{-10pt}
\end{figure}

\subsection{Learning to Verify}
Verifying solutions is often substantially more tractable than generating them, a distinction well-established in complexity theory and foundational to developments in knowledge proofs and learning algorithms~\cite{cook2023complexity, goldwasser2019knowledge, godel1956letter}. For example, in solving a maze, verifying the correctness of a given path is significantly easier than discovering such a path. This asymmetry has been recognized and utilized for several decades, notably in the field of cryptography~\cite{goldwasser2019knowledge, lavin2024survey, rivest1978method}. EBMs are built on this principle that \textit{verification is easier than generation}: rather than learning to generate directly, as in most existing paradigms, EBMs learn to generate by optimizing predictions with respect to this learned verification (energy function); this is depicted in Figure~\ref{fig:energy_landscape_minimization}. 

Recent works have attempted to leverage this characteristic of verifiers~\cite{Team_2023, yao2023tree, ma2025inference, ouyang2022training}, but these approaches decouple the verifier and generator, resulting in adversarial dynamics~\cite{ma2025inference} and challenges in scalability~\cite{yao2023tree}. For example, researchers combining tree search and LLMs required thousands or even millions of samples to achieve optimal performance~\cite{Team_2023}.
In contrast, EBMs \textbf{combine} the verifier and generator into a \textbf{single model}, where the generator is defined implicitly by the gradient of the verifier~\cite{du2019implicit}. We show that this coupling resolves scalability and adversarial issues (Figures~\ref{subfig:thinking_scaling} and~\ref{subfig:self_verification_non_adversarial}).

An additional advantage of verifiers is generalization. Because verification is usually easier than generation~\cite{swamy2025all}, \textit{prediction verification on Out-Of-Distribution (OOD) data is often easier than explicit prediction generation for OOD data}~\cite{du2022learning}. This characteristic often results in better generalization of verifiers than explicit generators~\cite{du2022learning}. As an example, models trained to explicitly solve mazes on a small grid may not generalize to larger grids, whereas verifiers for maze solutions learn whether a path is correct, which more easily transfers to larger mazes. This characteristic may be why EBMs often generalize better than existing models~\cite{du2022learning, du2024learning}, which we further support in our larger-scale experiments (Figure~\ref{subfig:ood_thinking_comparison} and Table~\ref{tab:img_denoising_performance}).

\subsection{Learning to Understand}

This verifier-centric perspective also relates to a deeper limitation in contemporary generative models, referred to as ``The Generative AI Paradox~\cite{west2023generative}.'' Although current generative models achieve strong generative capabilities, they frequently lack basic discrimination skills, such as the ability to assess the plausibility or coherence of their own predictions~\cite{west2023generative, stojnic2023commonsense}. These limitations impede their ability to engage in reasoning, planning, and decision-making~\cite{yan2025phd, kambhampati2024position}. In contrast, EBMs offer a potential solution to this challenge: as EBMs generate by learning a verifier (which is conceptually similar to a discriminator), they develop strong discrimination skills~\cite{wang2023energy}. Experimental results further support this observation (Table~\ref{tab:img_denoising_performance}).

%% file: arxiv/main_sec/4_approach.tex
\section{Energy-Based Transformers ({\pa}) Approach}

\subsection{Energy-Based Models (EBM) Background}
\label{sec:learning_approach}
Energy-Based Models (EBMs) assign a scalar energy value to each configuration of input variables, enabling them to model the \textbf{compatibility} and \textbf{interactions} between variables, such as between a context and candidate-prediction. In the case of probabilistic EBMs, this defines a probability distribution using a Boltzmann distribution $p_\theta(x) = \frac{e^{-E_{\theta}(x)}}{Z(\theta)}$ where $Z(\theta) = \int e^{-E_{\theta}(x)} dx$ is the intractable partition function involving an integral over all possible values of $x$. Due to the negative exponential, lower energy corresponds to higher probability, and higher energy corresponds to lower probability. To avoid the intractability of the partition function, it is common to work with \textbf{unnormalized EBMs}, which dispense of the partition function in favor of representing relative unnormalized probabilities. This formulation shifts the focus from addressing the partition function, to simply assigning low energy to the true data manifold and high energy elsewhere~\cite{du2019implicit, dawid2024introduction}, offering benefits such as scalability to spaces where the true data manifold is thin and therefore a probabilistic EBMs would have an infinite score~\cite{dawid2024introduction}. 
In supervised or predictive self-supervised learning (e.g., classification, autoregressive modeling, masked modeling), unnormalized EBMs can be formulated as: $p_\theta(x, \hat{y}) \propto e^{-E_{\theta}(x, \hat{y})}$, where the goal of the EBM is to learn to predict $\hat{y}$ given $x$.\footnote{In the case of self-supervised learning, $x$ is some unmasked portion of the original $x$ and $\hat{y}$ is the masked portion.} For an accessible and beginner-friendly introduction to EBMs, including intuitive explanations and pseudocode, please refer to Section~\ref{sec:ebm_intro}.

\begin{figure}[t]
  \centering
  \begin{minipage}[t]{0.48\linewidth}
    \begin{algorithm}[H]
      \small
      \caption{Training}\label{alg:ebt_training}
      \Inputs{Context $x$, Target $y$, EBM $E_\theta(x,\hat y)$}
      \Hparams{Steps $N$, Step Size $\alpha$, Loss $J(\cdot)$}
      Sample $\hat y_0 \sim \mathcal{N}(0,I)$\;
      \For{$i = 0, \dots, N-1$}{
        $\hat y_{i+1} \gets \hat y_i - \alpha\nabla_{\hat y_i}E_\theta(x,\hat y_i)$\;
      }
      $\mathcal{L} \gets J(\hat y_N,y)$\;
      \Return{$\mathcal{L}$, update $E_\theta$}\;
    \end{algorithm}
  \end{minipage}\hfill
  \begin{minipage}[t]{0.48\linewidth}
    \begin{algorithm}[H]
      \small
      \caption{Inference with Verification}\label{alg:ebt_inference}
      \Inputs{Context $x$, EBM $E_\theta(x,\hat y)$}
      \Hparams{Steps $N$, Step Size $\alpha$, Samples $M$}
      \For{$j = 1, \dots, M$}{
        Sample $\hat y_{0, j} \sim \mathcal{N}(0,I)$\;
        \For{$i = 0, \dots, N-1$}{
          $\hat y_{i+1, j} \gets \hat y_{i, j} - \alpha \nabla_{\hat y_{i, j}}E_\theta(x,\hat y_{i, j})$\;
        }
      }
      \Return {$\hat y^* = \operatorname*{argmin}_j E_\theta\!\bigl(x,\hat y_{N, j}\bigr)$}\;
    \end{algorithm}
  \end{minipage}
  \vspace{-10pt}
\end{figure}

\subsection{Scalable EBM Learning}
While EBMs offer a flexible modeling framework, training them scalably remains an open research problem. Two primary training approaches exist---contrastive and regularized methods~\cite{lecun2022path}. Contrastive methods increase the energy of negative samples while decreasing the energy of positive samples. Due to the curse of dimensionality, where the volume of spaces grows exponentially with their dimension, contrastive methods struggle to scale because they must increase the energy of an exponentially higher number of negative samples~\cite{dawid2024introduction}.

An alternative is to frame EBM learning as an optimization problem~\cite{du2022learning, wang2023energy}, which avoids the curse of dimensionality by implicitly regularizing the energy landscape, enabling scalable learning. In this approach, EBMs are trained to optimize an initial prediction to the ground truth solution through gradient descent, as shown in Figure~\ref{fig:energy_landscape_minimization}. This pushes the energy landscape to be convex surrounding the ground truth solution, thereby regularizing the energy landscape to only have low energy on the true data manifold. Intuitively, this training approach is similar to GANs~\cite{goodfellow2014generative}. During the forward pass, EBMs can be seen as a GAN discriminator by giving an energy ``verification''; on the backward pass they can be seen a GAN generator by optimizing predictions through energy minimization to try and fool the discriminator.  

Training EBMs to perform optimization can be formalized as follows. We begin with an EBM $E_{\theta}$, a prior (initial prediction) $\hat{y}_0$, an input (context) for the model $x$, and seek to predict $y$. We aim to find the minimum energy (most compatible/likely) $\hat{y}$ given an $x$, which we search for using gradient descent:
\begin{align}
\hat{y}_{i+1} = \hat{y}_i - \alpha \nabla_{\hat{y}_i}E_{\theta}(x, \hat{y}_i),
\end{align}
where $\alpha$ is the step size (this is formalized in Algorithm~\ref{alg:ebt_training}). 
Then, the loss can be computed using any standard objective function. For this work, we use the same loss functions as existing papers to simplify experiments, so we use categorical cross-entropy for language modeling~\cite{bengio2003neural} and mean squared error for image denoising~\cite{ho2020denoising}. Importantly, this loss is backpropagated through the entire optimization process, requiring second-order derivatives (i.e., gradients of gradients). These are computed efficiently via Hessian-vector products, which scale linearly with model size~\cite{dagréou2024howtocompute}, similar to standard first-order backpropagation in feed-forward models. More details and pseudocode can be found in Section~\ref{sec:how_to_learn_ebt}.

\subsection{Scalable EBM Thinking}
\label{sec:thinking_approach}
While this training approach is scalable, achieving smooth and convex energy landscapes, such as the landscape visualized in Figure~\ref{fig:energy_landscape_minimization}, remains challenging on real-world problems. Because $y$ is high-dimensional, the energy landscape spans a high-dimensional space, and must remain well-shaped throughout. To address this, we found three key \textit{energy landscape regularization} techniques to be essential in ensuring the smoothness and convexity of learned energy landscapes, enabling strong thinking capabilities to be learned during training. 

First, we found a replay buffer (following existing EBM works~\cite{du2022learning, du2019implicit}) helps simulate longer optimization trajectories, enabling energy landscapes to be well defined near their minimum.
Second, a variant of Langevin Dynamics~\cite{du2019implicit} (random noise), was found to be helpful for encouraging \textbf{exploration} of the energy landscape:
\begin{align}
\hat y_{i+1}
&= \hat y_i - \alpha\nabla_{\hat y_i}E_\theta\bigl(x,\hat y_i\bigr) + \eta_i,\quad
   \eta_i \sim \mathcal{N}(0, \sigma),
\end{align}
where $\sigma$ is the magnitude of the noise $\eta$. Without this random noise term, exploration is often limited to paths leading directly to the energy minimum, leaving other regions poorly defined.
Third, varying the paths taken towards predicting solutions, by randomizing the gradient descent step size $\alpha$ and number of optimization steps, significantly improved generalization. 
Together, these techniques substantially improved the System 2 Thinking capabilities of models, as confirmed by ablation experiments in Table~\ref{tab:system_2_ablations}.

With these energy landscape regularization techniques established, to understand the System 2 Thinking capabilities of {\pa}s, we explored two main thinking approaches, corresponding to two of the cognitive facets described. First, corresponding to dynamic computation allocation (Facet~\ref{facet:dynamic-computation}), we conduct experiments that involve changing the number of steps taken for optimization of a single prediction. This is conceptually similar to increasing the denoising steps performed with a diffusion model. Second, corresponding to the ability to verify predictions (Facet~\ref{facet:prediction-verification}), we generate N predictions from an EBM and then choose the minimum energy prediction. This is conceptually similar to Best of N (BoN) sampling using language models~\cite{stiennon2020learning}. However, EBMs generalize this approach to both discrete and continuous modalities, don't require additional supervision (e.g., an external reward/verification model), and perform it on \textit{every single prediction}, not just to entire sequences. We demonstrate performance improvements gained from both of these techniques in several Thinking experiments (Figures~\ref{fig:best_result},~\ref{fig:ood_generalization_thinking}, and~\ref{fig:thinking_fwd_passes_img}), which further confirm the importance of the described cognitive facets. This thinking process is formalized in Algorithm~\ref{alg:ebt_inference} and we more formally define and justify usage of the term thinking in Section~\ref{sec:thinking_formalization}.

\vspace{-2pt}
\subsection{Energy-Based Transformers ({\pa}s) Architecture}
\vspace{-2pt}
\label{sec:ebt_intro}

Transformers have demonstrated exceptional performance across numerous domains~\cite{openai2023gpt4, guo2025deepseek, oquab2023dinov2, he2021masked, borsos2023audiolm}. Three primary advantages of Transformers include their parallelizability~\cite{radford2019language, vaswani2017attention}, their stability~\cite{grattafiori2024llama}, and their scalability~\cite{kaplan2020scaling}. Because Energy-Based Models (EBMs) have encountered difficulties with all three of these characteristics~\cite{du2019implicit, du2020improved, li2023learning, arbel2020generalized}, Transformers represent a particularly suitable architecture for scaling EBMs. Consequently, to advance the EBM paradigm, we introduced Energy-Based Transformers (EBTs), which are Transformer implementations designed for EBMs. We developed two variants: a decoder-only EBT, inspired by the GPT architecture~\cite{radford2018improving} for autoregressive modeling, as well as a bidirectional EBT with bidirectional attention across sequences, enabling capabilities such as infilling and masked modeling~\cite{peebles2023scalable, devlin2019bert}. Although the bidirectional EBT implementation is relatively straightforward, the autoregressive EBT presents greater implementation challenges, primarily due to the potential for information leakage in naïve implementations. Comprehensive details of this implementation are discussed in Section~\ref{sec:ebt_full}.

%% file: arxiv/main_sec/5_experimentation.tex
\vspace{-1pt}
\section{Experimentation and Results}
\vspace{-1pt}

\label{sec:experimentation}
We experiment with {\pa} across both Autoregressive (AR)~\cite{radford2019language} as well as Bidirectional models~\cite{devlin2019bert} in discrete and continuous spaces.\footnote{Here, Autoregressive and Bidirectional refer to the procedure for generation. It's worth noting that autoregressive models are compatible with bidirectional attention, as in~\cite{li2025autoregressive}.} In discrete spaces, we focus primarily on the language modeling objective. In continuous spaces, we focus on vision tasks of next frame prediction and image denoising. We perform the most comprehensive experiments using Autoregressive Language Models trained to predict the next token, as this is the most extensively studied pretraining approach~\cite{li2025mis}. All models are pretrained from scratch under a tightly constrained compute budget, as the architecture of {\pa}s is incompatible with existing foundation models, making them incompatible for adaptation via fine-tuning. We focus on two primary types of results. First, we examine \textbf{learning scalability}, investigating how quickly models can fit the pretraining data, which is standard in the pretraining literature~\cite{gu2023mamba, touvron2023llama, li2025mis, alabdulmohsin2022revisiting, kaplan2020scaling, henighan2020scaling, hoffmann2022training}.

Second, we study \textbf{thinking scalability}, or how the performance of models changes as we scale the System 2 Thinking of models (Definition~\ref{def:system_2_thinking}). To measure the thinking scalability, we use the Number of Function Evaluations (NFEs)~\cite{chen2018neural, ma2025inference}, which we deem to be one for every forward pass completed (for EBMs this is one function evaluation per optimization step). By scaling the number of forward passes during inference, we can determine whether model performance improves with increased thinking.


\vspace{-1pt}
\subsection{Autoregressive Language Modeling Experiments}
\vspace{-1pt}

In this section, we detail and discuss the results for all Natural Language Processing (NLP) experiments using Autoregressive (AR) Language Models trained to predict the next discrete token in a text sequence~\cite{radford2019language}. All language models are pretrained on the RedPajamaV2 text corpus~\cite{weber2024redpajama, together2023redpajama} $100B$ sample from HuggingFace using the GPT-NeoX tokenizer~\cite{black2022gpt} (following~\cite{gu2023mamba}) to predict the next token. Following existing pretraining work, we compare AR {\pa} with the standard Transformer++ recipe~\cite{gu2023mamba, touvron2023llama, sun2024learning}. We manually created a training and validation split of $66$ million and $33$ thousand samples, respectively. 

\begin{figure}
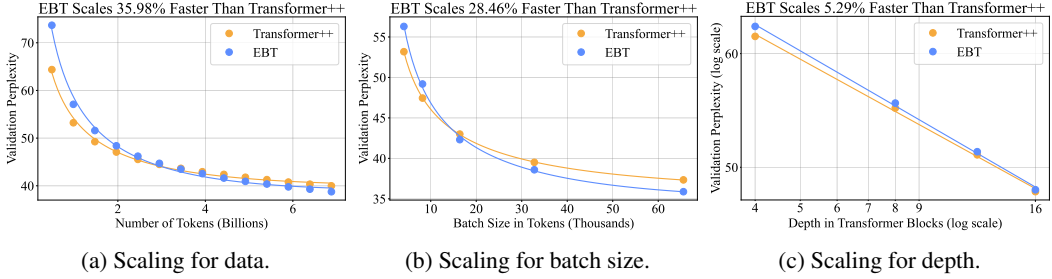

  \centering
  \scriptsize
  \begin{subfigure}[b]{0.33\columnwidth}
    \includesvg[width=\linewidth]{images/scaling_learning_nlp_ar_data.svg}
    \caption{Scaling for data.}
    \label{subfig:data_scaling}
  \end{subfigure}\hfill
  \begin{subfigure}[b]{0.33\columnwidth}
    \includesvg[width=\linewidth]{images/scaling_learning_nlp_ar_bs.svg}
    \caption{Scaling for batch size.}
  \end{subfigure}\hfill
  \begin{subfigure}[b]{0.33\columnwidth}
    \includesvg[width=\linewidth]{images/scaling_learning_nlp_ar_depth.svg}
    \caption{Scaling for depth.}
  \end{subfigure}
  \caption{\textbf{Language Learning Scalability---Data, Batch Size, and Depth.} A comparison between the scaling of the Transformer++ recipe~\cite{touvron2023llama} and EBTs across data, batch size, and depth during pretraining. On all of these axes, EBTs out-scale the Transformer++ recipe significantly, indicating improved data efficiency and suggesting potential benefits for generalization. Additionally, the improved depth scaling offers promise for reasoning, where depth is crucial~\cite{ye2024physics}. These results suggest that if these scaling trends persist, {\pa}s would likely outperform Transformer++ models at foundation model data scale.}
  \label{fig:nlp_ar_learn_scale_1}
  \vspace{-15pt}
\end{figure}

For downstream evaluation, we used four key datasets in addition to the pretraining dataset, spanning reasoning, question answering, and syntax understanding. Ordered roughly by increasing perplexity difficulty, these include GSM8K~\cite{cobbe2021training}, SQuAD~\cite{rajpurkar2016squad}, BigBench Elementary Math QA ~\cite{srivastava2022beyond}, and BigBench Dyck Languages~\cite{srivastava2022beyond}. We intentionally design the evaluation towards reasoning benchmarks due to their alignment with System 2 Thinking. Additionally, we focus on reporting perplexity as our relatively small models trained from scratch, when compared to current foundation models, do not achieve high accuracies on many of these benchmarks. Furthermore, perplexity often functions as a more linear metric than accuracy~\cite{schaeffer2023emergent, gu2023mamba}, enabling a more comparable analysis of downstream performance as we scale compute during inference. Further details on the exact hyperparameters and setup used for all experiments are in Section~\ref{sec:experimental_details}.

\vspace{-4pt}
\subsubsection{Autoregressive Language Model Learning Scalability}
\begin{figure}
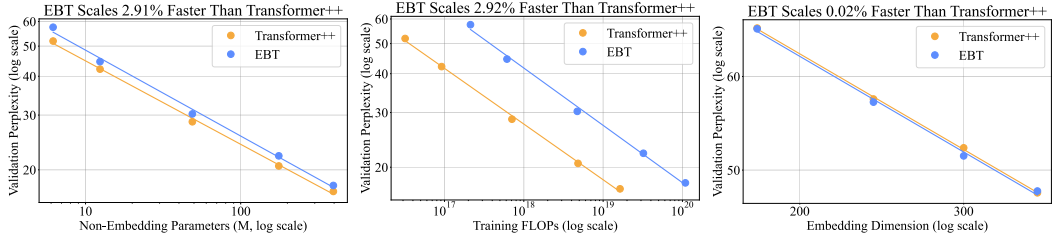

  \centering
  \scriptsize
  \begin{subfigure}[b]{0.33\columnwidth}
    \includesvg[width=\linewidth]{images/scaling_learning_nlp_ar_params.svg}
    \caption{Scaling for number of Parameters.}
  \end{subfigure}\hfill
  \begin{subfigure}[b]{0.33\columnwidth}
    \includesvg[width=\linewidth]{images/scaling_learning_nlp_ar_flops.svg}
    \caption{Scaling for number of FLOPs.}
  \end{subfigure}\hfill
  \begin{subfigure}[b]{0.33\columnwidth}
    \includesvg[width=\linewidth]{images/scaling_learning_nlp_ar_width.svg}
    \caption{Scaling for the embed. dimension.}
  \end{subfigure}
  \caption{\textbf{Language Learning Scalability---Parameters, FLOPs, and Width.} Pretraining scaling comparisons between the Transformer++ recipe~\cite{touvron2023llama} and EBTs across model size (parameters), compute (FLOPs), and width (embedding dimension). EBTs slightly out-scale the Transformer++ in FLOP and parameter scaling, becoming the first approach to achieve a higher \textbf{scaling rate} without modifying the tokenizer~\cite{pagnoni2024byte} to our knowledge. These results suggest that {\pa}s offer high promise as a pretraining paradigm in both parameter and FLOP efficiency as scale increases.}
  \label{fig:nlp_ar_learn_scale_2}
  \vspace{-5pt}
\end{figure}

The scaling trends related to the learning speed of models, commonly referred to as ``scaling laws,'' are challenging to measure. For example, a recent survey~\cite{li2025mis} found that the ``scaling laws'' observed often depend on several implementation details and axes to measure, which can result in several different conclusions.\footnote{For more information on this perspective please refer to Section~\ref{sec:scaling_law_extra_info}} Therefore, to be as comprehensive as possible in determining how {\pa}s scales compared to the Transformer++, we conduct scaling experiments for six different axes--including data, batch size, depth, parameters, FLOPs,\footnote{The FLOP calculation is nuanced and depends on specific hyperparameters. For more information please refer to Section~\ref{sec:flop_calc}.} and embedding dimension. The results for the data, batch size, and depth scaling are shown in Figure~\ref{fig:nlp_ar_learn_scale_1}; and the results for parameters, FLOPs, and embedding dimension are visualized in Figure~\ref{fig:nlp_ar_learn_scale_2}. Across all axes {\pa} consistently out-scales (has a higher scaling rate) the Transformer++ recipe, becoming the first model to achieve such a feat without using a different tokenizer (\cite{pagnoni2024byte} was the first and only work to our knowledge to out-scale the Transformer++ recipe, but they use a different tokenizer, and do not out-scale the Transformer++ across multiple axes unlike {\pa}s). These results suggest that {\pa}s are more data efficient, batch size efficient, parameter efficient, depth efficient, and compute efficient than the Transformer++ recipe. Thus, at the scale of modern foundation models trained on $1,000\times$ more data with models $1,000\times$ larger (following~\cite{grattafiori2024llama}), we expect the pretraining performance of {\pa}s to be significantly better than that of the Transformer++ recipe.

\begin{table}[t]
\caption{\textbf{System 2 Thinking Ablations.} All energy landscape regularization techniques described in Section~\ref{sec:thinking_approach} and their impact on System 2 Thinking performance, measured by percent perplexity improvement. Thinking Longer denotes more optimization steps and Self-Verification denotes optimizing many predictions and choosing the best. Bolded highlights the default System 2 Hyperparameters, leveraging all energy landscape regularization techniques described. This configuration results in the best performance when thinking longer and doing self-verification. Removing regularization, such as Langevin Dynamics, results in less energy landscape exploration, which improves single path performance (thinking longer) at the expense of self-verification performance.}
\centering
\small
\begin{tabular}{lcc}
\toprule
\textbf{Model} & \textbf{Thinking Longer $\uparrow$} & \textbf{Thinking Longer and Self-Verification $\uparrow$}\\
\midrule
No Random Step Size  & -1.47 & 0.19 \\
No Random Num. Steps  & 0.00 & 9.65 \\
No Langevin Dynamics  & \textbf{17.2} & 17.0 \\
No Replay Buffer  & 14.8 & 17.8 \\
\textbf{Full System 2 Configuration} & 7.19 & \textbf{18.7} \\
\bottomrule
\label{tab:system_2_ablations}
\vspace{-20pt}
\end{tabular}

\end{table}

\vspace{-4pt}
\subsubsection{Autoregressive Language Model Thinking Scalability}
Building on the learning results, we investigate {\pa}s for thinking at inference time. We found that the thinking capabilities of {\pa} emerge with a sufficiently large data scale, and therefore, due to limited resources, we focus on conducting thinking experiments with smaller models trained on substantial amounts of data. We test thinking capabilities along two axes: thinking longer, which denotes more optimization steps, and self-verification, which denotes generating many candidate predictions and selecting the minimum energy prediction.
In Table~\ref{tab:system_2_ablations}, we conduct ablation studies to confirm the benefits of our \textit{energy landscape regularization} techniques for System 2 Thinking on Out-of-Distribution Data from the BigBench Dyck Languages benchmark~\cite{srivastava2022beyond}. 
We find that using all techniques yields the best System 2 Thinking performance when combining extended thinking and self-verification. Additionally, the results show that randomizing the step size is critical---removing it nearly eliminates Thinking gains, while disabling Langevin Dynamics degrades combined performance but improves results without verification, offering a performance-compute tradeoff.

\begin{figure}[t]
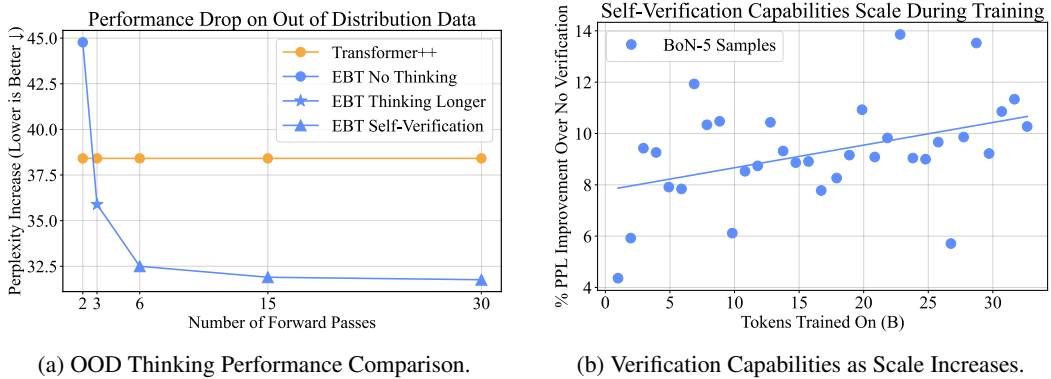

  \centering
  \scriptsize
  \begin{subfigure}[b]{0.48\columnwidth}
    \includesvg[width=\linewidth]{images/scaling_thinking_nlp_ar_fwd_advanced.svg}
    \caption{OOD Thinking Performance Comparison.}
    \label{subfig:ood_thinking_comparison}
  \end{subfigure}\hfill
  \begin{subfigure}[b]{0.48\columnwidth}
    \includesvg[width=\linewidth]{images/scaling_thinking_nlp_ar_data_ood_linear.svg}
    \caption{Verification Capabilities as Scale Increases.}
    \label{subfig:thinking_scaling}
  \end{subfigure}
  \caption{\textbf{EBT Thinking Analysis.} (a) Mean performance degradation of the standard Transformer++ recipe~\cite{touvron2023llama} and the Energy-Based Transformer (EBT) on four Out of Distribution (OOD) datasets. While the Transformer++ cannot reduce perplexity at a \textit{per-token level}, EBTs can by performing more forward passes over a single token/sample (Thinking Longer) as well as generating many samples and choosing the minimum energy one (Self-Verifying/BoN in addition to longer thought). (b) The Self-Verification capabilities of EBTs using BoN-5 compared to not doing any self-verification. As data scale increases, the benefit from doing self-verification increases. These results suggest EBTs generalize OOD better than the Transformer++ because of their System 2 Thinking capabilities, and that the thinking capabilities of EBTs scale during training.}
  \label{fig:best_result}
  \vspace{-15pt}
\end{figure}

Having established the importance of these landscape regularization techniques, in Figure~\ref{fig:best_result}, we analyze the scalability of thinking with {\pa}s, where the results yield two main insights. First, as shown in Figure~\ref{subfig:ood_thinking_comparison}, EBTs are able to improve performance by as much as $29\%$ by increasing the amount of forward passes (thinking time), whereas the Transformer++ cannot improve performance at all.\footnote{Because we pretrained language models from scratch, and are unable to train models the size of modern foundation models, we find models did not benefit from inference time techniques such as Chain-of-Thought. However, we expect both {\pa} and Transformer++ models to benefit equally from existing techniques.} This aligns with our claims that because traditional feed-forward Transformers cannot dynamically allocate additional computation for each prediction being made, they are unable to improve performance for each token by thinking for longer. 

\begin{figure}[t]
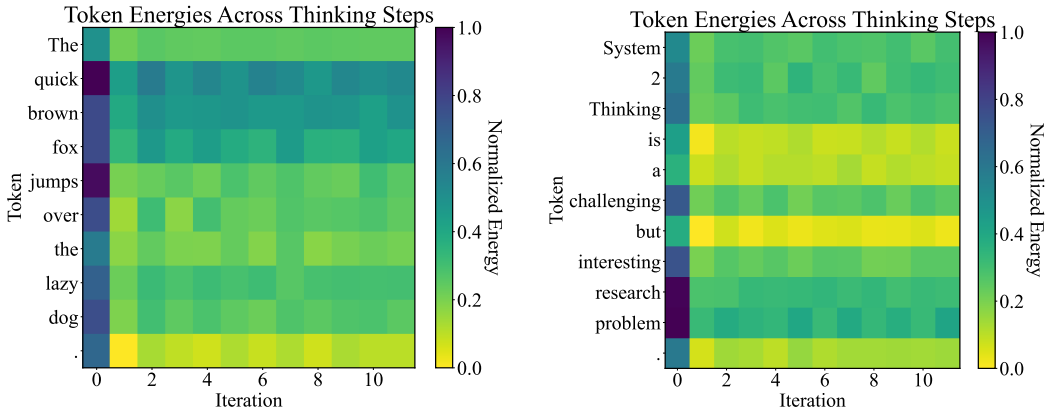

  \centering
  \scriptsize
  \begin{subfigure}[b]{0.48\columnwidth}
    \includesvg[width=\linewidth]{images/token_energy_seq5.svg}
    \caption*{}
  \end{subfigure}\hfill
  \begin{subfigure}[b]{0.48\columnwidth}
    \includesvg[width=\linewidth]{images/token_energy_seq3.svg}
    \caption*{}
  \end{subfigure}
  \vspace{-20pt}
  \caption{\textbf{Learning Uncertainty on Text Results.} EBTs learn to vary uncertainty across text tokens without any explicit supervision. As an example, in both (a) and (b), simple tokens such as ``.'', ``is'', ``a'', ``but'', or ``the'' have lower energies across inference-time optimization (thinking) steps, indicating lower uncertainty. On the other hand, harder to predict tokens such as ``quick'', ``brown'', ``research'', and ``problem'' have higher energies across optimization steps, and more difficulty in achieving energy convergence, meaning the model is more uncertain. Inspired by~\cite{geiping2025scaling}.}
  \label{fig:qual_nlp_uncertainty}
  \vspace{-15pt}
\end{figure}

Second, as demonstrated in Figure~\ref{subfig:thinking_scaling}, the thinking capabilities of {\pa}s scale, showing that as {\pa}s are trained for longer, their ability to achieve improvements from verification improves, increasing up to $10\%-14\%$ from $4\%-8\%$. This suggests that {\pa}s trained at the same scale as modern foundation models, such as the $15$T tokens Llama3~\cite{grattafiori2024llama} was trained on ($\approx1000 \times$ the current data scale), would have significantly more substantial results from self-verifying. Lastly, we visualize results from {\pa} at representing uncertainty while predicting tokens in Figure~\ref{fig:qual_nlp_uncertainty}. The results demonstrate that for easier to predict tokens, such as ``the'' or ``but'', {\pa}s optimize to lower energies faster, whereas for harder to predict tokens, such as ``fox'' or ``problem'' {\pa}s have higher energy that does not converge across steps. This suggests that during pretraining {\pa}s learn to capture uncertainty regarding which tokens are harder or easier to predict, achieving Facet~\ref{facet:prediction-uncertainty}. 

\subsubsection{Autoregressive Language Model Generalization}

\begin{wrapfigure}[25]{r}{0.5\textwidth}
  \centering
  \includesvg[width=\linewidth]{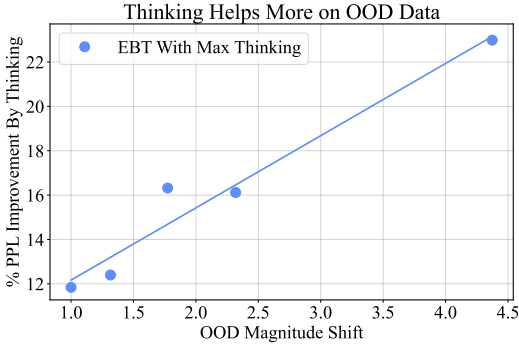}
  \caption{\textbf{OOD Thinking Performance.} As the data becomes more OOD, thinking leads to greater performance improvements, with a roughly linear trend. These findings highlight that {\pa}s thinking is especially critical for robust generalization to OOD data. Performance is measured on $5$ different datasets varying in Out-of-Distribution (OOD) magnitude shift, which is measured as the ratio of downstream dataset perplexity to pretraining perplexity. Max Thinking denotes combining thinking longer and self-verification. 
  }
  \label{fig:ood_generalization_thinking}
\end{wrapfigure}

As System 2 Thinking in humans is associated with generalization to novel scenarios, we conduct experiments directly aimed at measuring the effects of System 2 Thinking on generalization. In Figure~\ref{fig:ood_generalization_thinking}, we visualize the performance of {\pa}s on the datasets described, which have varying levels of Out-of-Distribution (OOD) shift (measured as the ratio of downstream task perplexity to pretraining perplexity). We observe a strong linear trend: as the data becomes more OOD, thinking leads to greater performance improvements. Therefore, these findings suggest that the benefits of EBTs' thinking are not uniform across all data but scale positively with the magnitude of distributional shifts, highlighting thinking as a critical mechanism for robust generalization beyond training distributions. These findings align with observations in psychology where humans rely on deliberate System 2 Thinking to tackle challenging OOD tasks.

Next, we investigate the relation between OOD generalization and pretraining performance. Pretraining performance is strongly correlated with downstream task performance in language models~\cite{gadre2024language, gururangan2020don, thrush2024improving}, and can even be a predictor of downstream performance~\cite{chen2024scaling, isik2024scaling, grattafiori2024llama}. Because we know that EBTs scale at a faster rate than the Transformer++, as demonstrated in Figures~\ref{fig:nlp_ar_learn_scale_1} and~\ref{fig:nlp_ar_learn_scale_2}, it is reasonable to hypothesize that they may also perform better on downstream tasks at scale.
To investigate this, we compare models with identical training setups, where EBTs have slightly worse pretraining perplexity than Transformer++ models. As shown in Table~\ref{tab:generalization_exps}, despite achieving a higher pretraining perplexity, EBTs achieve lower (better) perplexity on most downstream tasks, suggesting stronger generalization, particularly to Out-of-Distribution (OOD) data. Together, with the better learning scalability results, and knowing that improved pretraining performance usually leads to improved downstream task performance~\cite{chen2024scaling, isik2024scaling, grattafiori2024llama}, these results suggest that at scale {\pa}s would outperform the Transformer++.

\begin{table}[t]
\caption{\textbf{Language Model Task Generalization Comparison.} Despite exhibiting a slightly higher pretraining perplexity, {\pa}s usually achieve lower perplexity on downstream tasks than the Transformer++. This suggests that {\pa}s generalize better than the Transformer++. Additionally, because {\pa}s scale better than the Transformer++ during pretraining (Figure~\ref{fig:nlp_ar_learn_scale_1}), these findings suggest that {\pa}s would outperform Transformer++ at foundation model scale. BB stands for BigBench.}
\centering
\small
\begin{tabular}{lccccc}
\toprule
\textbf{Model}  & \textbf{Pretrain} & \textbf{GSM8K ↓} & \textbf{SQuAD ↓} & \textbf{BB Math QA ↓} & \textbf{BB Dyck ↓} \\
\midrule
Transformer++  & \textbf{31.36} & 49.6 & \textbf{52.3} & 79.8 & 131.5 \\
EBT            &  33.43         & \textbf{43.3} & 53.1 & \textbf{72.6} & \textbf{125.3} \\
\bottomrule
\label{tab:generalization_exps}
\end{tabular}
\vspace{-10pt}
\end{table}

\subsection{Autoregressive Video Experiments}

To assess how {\pa}s scale in continuous domains, we train models to predict the next image in a video conditioned on all previous frames---a common pretraining objective for video models~\cite{deng2024autoregressive, weissenborn2019scaling, rakhimov2020latent, ye2023video, gu2025long}. Unlike in the NLP experiments, where models see each sample only once due to the dataset size, current popular video datasets are relatively small, requiring models to train repeatedly on the same data. As a result, this setting probes a different question: \textit{``how well can models fit a fixed dataset?''} rather than how efficiently models scale under non-data-bound regimes. This distinction is especially important given the recent scarcity of high-quality datasets~\cite{villalobos2022will, Data-centricML-benchmarking} and the perspective that data will increasingly become a bottleneck.

\begin{figure}[t]
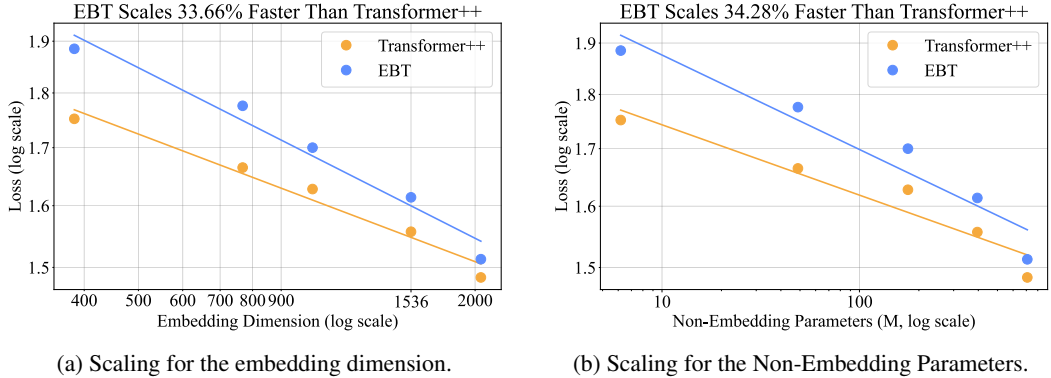

  \centering
  \scriptsize
  \begin{subfigure}[b]{0.48\columnwidth}
    \includesvg[width=\linewidth]{images/scaling_learning_vid_ar_embed.svg}
    \caption{Scaling for the embedding dimension.}
  \end{subfigure}\hfill
  \begin{subfigure}[b]{0.48\columnwidth}
    \includesvg[width=\linewidth]{images/scaling_learning_vid_ar_params.svg}
    \caption{Scaling for the Non-Embedding Parameters.}
  \end{subfigure}
  \caption{\textbf{Video Learning Scalability---Width and Parameters.} The minimum validation loss achieved on the Something Something V2 (SSV2) dataset. While {\pa}s achieve higher validation loss than the Transformer++ at smaller scales, the scaling rate is more than $33\%$ higher, suggesting that at foundation model scale with hundreds of billions of parameters {\pa}s would perform much better than the Transformer++. Notably, scaling with respect to the embedding dimension behaves more linearly than for the number of parameters, likely due to the embedding dimension serving as a bottleneck for the image representation.}
  \label{fig:cv_scaling_ssv2}
\end{figure}

For experiments, we encode all $224 \times 224$ images into $3136$ dimensional features with the frozen SD-XL VAE~\cite{rombach2022high, stabilityai/sd-vae-ft-mse}. Then, all models are trained using a Smooth L1 loss with $\beta = 1.0$ on the Something Something V2 dataset~\cite{goyal2017something}, where we report the minimum validation loss achieved. 
In Figure~\ref{fig:cv_scaling_ssv2} we report scaling results for the embedding dimension and non-embedding parameter count, as we found these axes behaved the most linearly. The results demonstrate that, despite achieving a higher initial loss, {\pa}s scale at a more than $33\%$ faster rate than the Transformer++. This suggests that at foundation model scale {\pa}s would achieve significantly better performance than the Transformer++. 

\begin{wrapfigure}[32]{r}{0.5\textwidth}
    \centering
    \includesvg[width=\linewidth]{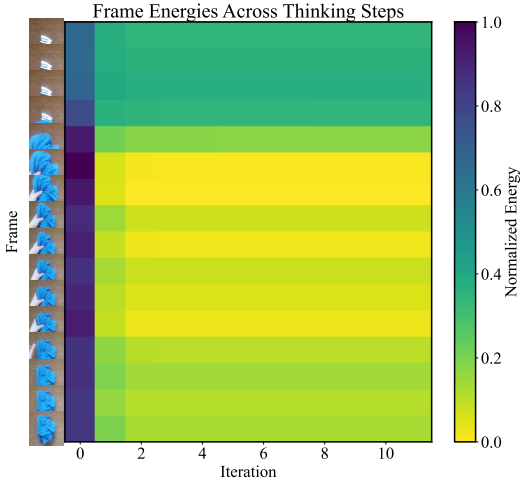}
    \caption{\textbf{Learning Uncertainty on Video Results.} In line with cognitive Facet~\ref{facet:prediction-uncertainty}, EBTs learn to express uncertainty across continuous video frames without supervision. At the start of the video, uncertainty is high (high energy) because the frame is mostly empty and the scene is highly unpredictable. As a blue garment is placed into the frame, uncertainty decreases (low energy), reflecting the greater predictability of the scene. When the blue garment is removed from the scene, uncertainty increases again, indicating a return to higher unpredictability. Such a capability is significantly more difficult to achieve in continuous spaces with traditional feed-forward transformers without discretization schemes~\cite{pei2022transformer}.}
    \label{fig:qual_vid_uncertainty}
\end{wrapfigure}

We believe this large scaling rate gap can be linked to the fact that {\pa}s more seamlessly model continuous distributions than standard feed forward transformers due to being able to express uncertainty (Facet~\ref{facet:prediction-uncertainty}) through their energy scalar. To confirm this, we visualize results for different energies when predicting video frames in Figure~\ref{fig:qual_vid_uncertainty}. The results demonstrate that {\pa}s successfully learn to capture uncertainty---where frames earlier on in the video have higher energy (higher uncertainty) due to no large objects being within the frame, and then as the major object in the scene becomes revealed more {\pa} predicts lower energy (lower uncertainty). {\pa}s learn to exhibit this behavior without any supervision using a Smooth L1 loss, whereas the standard feed-forward Transformer++ would require discretization schemes such as Vector Quantization~\cite{islam2023eqa} with a categorical loss, or other tricks to achieve the same effect.


\begin{table}[t]
\captionsetup[table]{position=top}
\caption{\textbf{Image Denoising and Classification Comparison}. For image denoising, EBTs significantly outperform DiTs~\cite{peebles2023scalable} in Peak Signal to Noise Ratio (PSNR), as well as MSE, on both in-distribution and Out-Of-Distribution (OOD) data, while using $99\%$ less forward passes. This suggests that EBTs generalize better than DiTs while using less computation. On image classification, EBTs also perform better than DiTs, yielding around $10\times$ higher accuracy, suggesting that EBTs learn better image representations and therefore understand images better than DiTs.}
\centering
\small
\begin{tabular}{lcc cc cc}
\toprule
 & \multicolumn{2}{c}{In Distribution Noise $\sigma = 0.1$} & \multicolumn{2}{c}{OOD Noise $\sigma = 0.2$} & \multicolumn{2}{c}{ImageNet-1k Classification} \\
\cmidrule(r){2-3} \cmidrule(r){4-5} \cmidrule(r){6-7}
Model & PSNR $\uparrow$ & MSE Pixel $\downarrow$ & PSNR $\uparrow$ & MSE Pixel $\downarrow$ & Top 1 Acc. $\uparrow$ & Top 5 Acc. $\uparrow$ \\
\midrule
DiT & 26.58 & 142.98 & 19.56 & 718.7 & 0.31\% & 1.36\%\\
EBT & \textbf{27.25} & \textbf{122.55} & \textbf{23.29} & \textbf{305.2} & \textbf{5.32\%} & \textbf{13.2\%}\\
\bottomrule
\vspace{-20pt}
\label{tab:img_denoising_performance}
\end{tabular}
\end{table}

\subsection{Bidirectional Image Experiments}
In addition to investigating autoregressive {\pa}s, we also explore the performance of {\pa}s trained bidirectionally. These experiments allow for a fairer comparison with diffusion models, which are not commonly trained autoregressively. Following~\cite{du2022learning, chen2020learning}, models are trained to denoise images that have been noised. To make these denoising experiments compatible with diffusion models, we deviate from the original noising schemes performed in these works and use a noising scheme based on the noising schedule from diffusion models. Specifically, we follow~\cite{peebles2023scalable}, and use a linear variance schedule ranging from $1\times10^{-4}$ to $2\times10^{-2}$. To control the noise level, we use a hyperparameter denoted $\sigma$ representing the percentage of the diffusion schedule to noise samples; $\sigma$ was set to $0.1$ during training, and $0.2$ during testing to test generalization. We use the COCO 2014 dataset~\cite{lin2014microsoft, AbdoTW/COCO_2014DatasetsHF} with $128$ by $128$ images, a patch size of $16$, and the Diffusion Transformer implementation from~\cite{peebles2023scalable}.

\begin{figure}[t]
    \centering
    \includegraphics[width=0.95\textwidth]{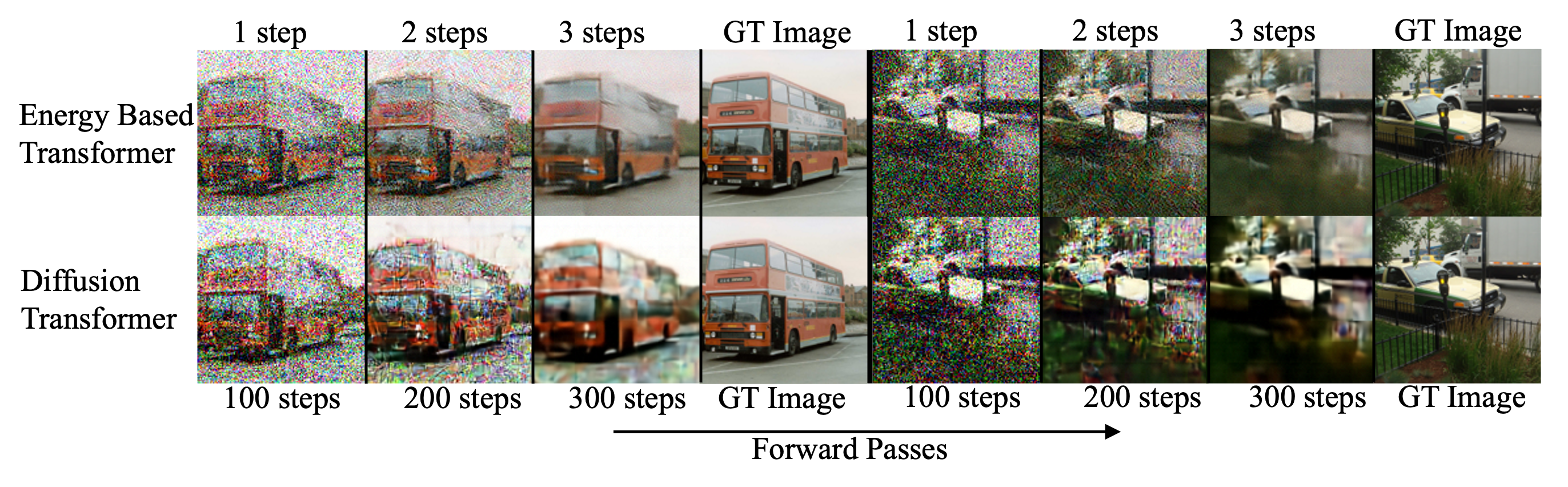}
    \vspace{-5pt}
    \caption{\textbf{Qualitative OOD Image Denoising.} {\pa}s achieve better denoising quality during inference while using one step for every $100$ denoising steps of a DiT. The overall image quality of {\pa} denoised images is less blurry than images denoised by DiT.} 
    \label{fig:qualitative_image_denoising}
    \vspace{-10pt}
\end{figure}

Following~\cite{du2022learning, chen2020learning}, during inference we investigate denoising on the same level of noise performed during training ($\sigma = 0.1$), as well as at a higher noise level ($\sigma = 0.2$), representing Out-of-Distribution (OOD) noisier images. The results are in Table~\ref{tab:img_denoising_performance}, where we observe that {\pa}s perform better than DiTs at both in and out of distribution image denoising across various metrics, by as much as $3.5$ in Peak Signal to Noise Ratio (PSNR). Following~\cite{chen2018neural}, we also plot the performance based on the number of forward passes (Number of Function Evaluations NFEs) in Figure~\ref{fig:thinking_fwd_passes_img}. These results demonstrate that {\pa}s perform better at denoising while using $99\%$ less denoising steps than DiTs, and that the System 2 Thinking scaling rate for {\pa}s is higher than for DiTs. Lastly, qualitative results for denoised out-of-distribution images for {\pa} compared to the DiT baseline are shown in Figure~\ref{fig:qualitative_image_denoising}. These results further demonstrate that the visual quality of EBT denoised images is much better than the visual quality of DiT denoised images, while using less compute.

In an effort to understand whether the representations learned from denoising captured useful visual features, we perform a linear probe evaluation on ImageNet-1k~\cite{russakovsky2015imagenet}, following common practice in visual representation learning~\cite{oquab2023dinov2, he2021masked}. For both models, we take the average of all the final patch tokens, and for DiTs we feed in $T = 0$. The results are shown in Table~\ref{tab:img_denoising_performance}, where the performance of EBTs is much higher than DiTs, achieving a Top-1 and Top-5 accuracy around $10 \times$ higher than that of DiTs. This suggests that EBTs learn better image representations than DiTs, and therefore that EBTs offer promise in generative models with an \textbf{improved understanding} of what they generate.






%% file: arxiv/main_sec/7_discussion.tex
\section{Discussion}
Across both discrete (text) and continuous (video) autoregressive models, the results show that {\pa}s scale at a faster rate than the standard Transformer++ approach during pretraining across all measured axes, including data, batch size, depth, parameters, FLOPs, and width. This is especially apparent with data and batch scaling for text, as well as width and parameter scaling for video, where the scaling rate was over $30\%$ higher. These results are particularly important for two reasons: first, they suggest that at the scale of modern foundation models, {\pa}s would outperform the current Transformer++ approach \textit{even without their System 2 Thinking capabilities}. Second, {\pa}s appear to be the first approach that has \textit{better data efficiency} than the Transformer++. As data has become one of the major limiting factors in further scaling~\cite{openaiYouTubegpt4_5}, this makes {\pa}s especially appealing.

With System 2 Thinking, the EBT-Transformer++ performance gap would be expected to increase, as we found that System 2 Thinking could improve performance by as much as $29\%$ for text (Figure~\ref{subfig:ood_thinking_comparison}). Further, we found that the thinking capabilities of EBTs scale well, as they improve during pretraining (Figure~\ref{subfig:thinking_scaling}) and perform better for data that is more OOD (Figure~\ref{fig:ood_generalization_thinking}). 
These findings imply that OOD generalization benefits from System 2 Thinking will further increase at larger scales, paving the way for principled generalization to OOD data. In addition, EBTs become increasingly robust to self-generated errors during verification (Figure~\ref{subfig:self_verification_non_adversarial}), indicating that their self-verification process scales reliably and sidesteps the adversarial instabilities reported in prior works~\cite{ma2025inference,huang2025best}.

\begin{wrapfigure}[24]{r}{0.50\textwidth}
  \centering
  \includesvg[width=\linewidth]{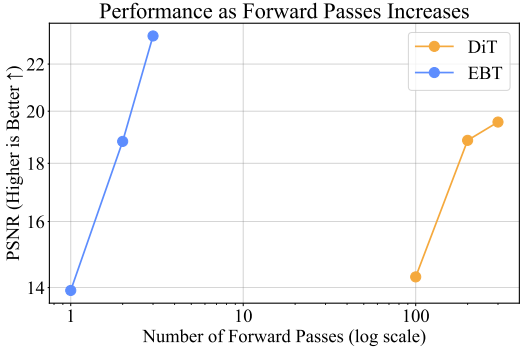}
  \caption{\textbf{Image Denoising Thinking Scalability.} A comparison between EBT and DiT on image denoising given a different number of forward passes. EBTs require only $1\%$ of the forward passes used by DiT to achieve comparable or better PSNR. Further, the scaling rate of PSNR improvement given more forward passes is much higher for EBTs than it is for DiTs. These results suggest EBTs have superior thinking capabilities than DiTs on OOD data.}
  \label{fig:thinking_fwd_passes_img}
  \vspace{-10pt}
\end{wrapfigure}

We hypothesize that the superior scaling of {\pa}s compared to the Transformer++ can be attributed to {\pa}s learning to verify (Facet~\ref{facet:prediction-verification}) rather than solely learning to predict. As discussed in Section~\ref{sec:intuition}, verification often generalizes better than amortized generation, which we believe leads to improved learning efficiency. The results in our generalization experiments further support this idea of verification leading to improved generalization, as we found that \textit{given a slightly worse pretraining performance}, {\pa}s still outperform the Transformer++ recipe on downstream tasks. Additionally, corresponding to Facet~\ref{facet:prediction-uncertainty} regarding prediction uncertainty, we find that the energy values learned by {\pa}s correlate strongly with more challenging to predict data, as shown in Figures~\ref{fig:qual_nlp_uncertainty} and~\ref{fig:qual_vid_uncertainty}. This is a promising characteristic of {\pa}s, as it enables uncertainty estimation within continuous state spaces, which would allow for principled inference-time behavior adaptation (Facet~\ref{facet:dynamic-computation}) when models determine a problem is more challenging.

Lastly, the results from our bidirectional experiments indicate that {\pa}s hold strong promise in bidirectional modeling. Particularly, we find that bidirectional EBTs outperform the DiT baseline in image denoising on both In and Out-of-Distribution performance, while using significantly fewer forward passes. This suggests that EBTs offer promise in tasks such as image generation or bidirectional text generation. Further, when evaluating the representations from DiTs and EBTs, we find that EBTs perform significantly better, achieving up to a $10 \times$ improvement in accuracy, suggesting EBTs offer promise in developing a \textit{better understanding} of what is being generated when performing generative modeling.

%% file: arxiv/main_sec/3_related_work.tex
\section{Related Work}

\subsection{Traditional Transformers} The Transformer architecture~\cite{vaswani2017attention} has become ubiquitous across various domains~\cite{radford2019language, touvron2023llama, latif2023transformers, oquab2023dinov2}. The most commonly used transformer variant of today makes predictions directly in the output space with a single forward pass, demonstrated in Figure~\ref{subfig:ar_transformer}. Because these models have a finite depth and width, and make predictions in a single forward pass, they are unable to dynamically allocate more computation to \textit{each} prediction being made. Furthermore, they cannot model uncertainty in continuous state spaces in the same way they can in discrete state spaces, because the normalization process for continuous state spaces is not as well-defined as it is for discrete spaces using softmax~\cite{dawid2024introduction}. Rather, training these models to express uncertainty relies on tricks such as Vector Quantization~\cite{van2017neural} or pseudo losses/objectives (e.g., ELBO~\cite{kingma2013auto}). Finally, because these models are not trained to explicitly verify samples, improving inference-time performance at a per-prediction level often requires external models~\cite{lightman2023let}.

\subsection{RNNs} Recently, several RNN variants (Figure~\ref{subfig:rnn_arch}) have emerged to alleviate memory bottlenecks and achieve faster inference~\cite{gu2023mamba, peng2023rwkv}. These approaches have scaled similarly to Transformers in autoregressive sequence modeling and achieve better memory efficiency and reduced latency. However, traditional RNNs that update their internal state based solely on new information/data~\cite{gu2023mamba, peng2023rwkv} are not capable of allocating additional computation during inference, and thus suffer from the same flaws as traditional transformers in achieving human-like System 2 Thinking.

To resolve these issues, people have equipped RNNs with the ability to allocate computation dynamically, with architectures such as the Universal Transformer~\cite{dehghani2018universal}. Recently, this type of RNN has also been applied to LLMs~\cite{geiping2025scaling, saunshi2025reasoning}, allowing LLMs to reason using additional computation in a continuous latent space through the depth of an unrolled RNN. However, like Diffusion models, these models learn to amortize gradient prediction of the energy function~\cite{geiping2025scaling}, meaning they cannot model uncertainty or explicitly verify predictions. Consequently, EBMs generalize these RNN-based architectures by offering explicit prediction verification capabilities~\cite{ma2025inference}. Further discussion on this relationship is provided in Section~\ref{sec:additional_related_work}.

\subsection{Dynamic Computation with LLMs}
The ability to leverage a dynamic amount of computation in LLMs has been emulated using chain-of-thought prompting~\cite{wei2022chain} and continuous latent space reasoning~\cite{hao2024training}. While these approaches can improve performance, they don't seamlessly transfer to continuous modalities, and LLM chain-of-thought has been shown to be unreliable for reasoning~\cite{lin2025implicit, turpin2023language, agarwal2024faithfulness}. More recently, models have been explicitly trained to perform reasoning using Reinforcement Learning~\cite{jaech2024openai, guo2025deepseek, xai2025grok3, anthropic2025claude37sonnet}. These approaches allow LLMs to simulate additional computational depth based on the number of tokens decoded before making a prediction, and as a result, significantly improve performance~\cite{jaech2024openai, guo2025deepseek}. 
The main limitations of these approaches are that they currently apply only to discrete domains (i.e., LLMs), are effective on a narrow set of problems that are easily verifiable (e.g., math and coding), and require additional supervision, typically in the form of reward signals, making them incompatible with purely unsupervised pretraining~\cite{guo2025deepseek}.

\subsection{Dynamic Computation with Diffusion}
The most common instance of a model architecture specifically created to leverage dynamic computation is diffusion models (Figure~\ref{subfig:diffusion_arch}), where using multiple forward passes to generate a prediction is a core aspect of both training and inference~\cite{höppe2022diffusion, rombach2022high}. 
Although diffusion models implicitly define a likelihood through the reverse process~\cite{ho2020denoising}, which could theoretically be used to verify predictions, in practice an external verifier is necessary to improve performance at inference time beyond increasing denoising steps~\cite{ma2025inference, liu2025video, singhal2025general}. This requirement limits the generalizability and scalability of diffusion models as an approach for System 2 Thinking, as they do not have two of the cognitive facets discussed in Table~\ref{tab:cognitive_facets}: the ability to model uncertainty in continuous state spaces and the ability explicitly verify predictions without additional models~\cite{ma2025inference, liu2025video, singhal2025general}. 
Furthermore, diffusion models rely on a fixed denoising schedule, which restricts their ability to adaptively halt or extend computation---unlike EBMs.
Additionally, diffusion models can be seen as predicting the gradient of the data density/energy function~\cite{du2023reduce}, and therefore EBMs are a generalization of diffusion models that learn to explicitly verify predictions. More on this connection is in Section~\ref{sec:additional_related_work}, and a side-by-side comparison of diffusion models and EBMs is in Figure~\ref{fig:diffusion_vs_ebm}.


\subsection{Energy-Based Models (EBMs)}
The perspective of energy minimization as thinking/reasoning has been known for some time~\cite{lecun2006tutorial}. Therefore, the most similar approaches to {\pa}s also train EBMs to do reasoning/thinking~\cite{du2022learning, du2024learning}. While these works achieved impressive generalization results, they only focus on small-scale problems, and did not scale EBMs to high-dimensional real-world problems such as language or video. Additionally, these works did not perform an in-depth analysis on the types of System 2 Thinking that emerge with EBMs, more complex inference time procedures beyond just increasing the number of gradient descent steps, approaches towards improving EBM scalability, and required techniques for enhancing System 2 Thinking in EBMs.

%% file: arxiv/main_sec/8_conclusion.tex
\section{Limitations and Conclusion}
In this work, we proposed Energy-Based Transformers ({\pa}s), a new paradigm that frames System 2 Thinking as an optimization procedure with respect to a learned verifier (an Energy-Based Model), enabling System 2 Thinking to emerge across any problem or modality entirely from unsupervised learning.

\paragraph{Limitations.} Despite demonstrating strong performance, {\pa}s have several current limitations. First, because {\pa}s generate predictions through an optimization process, they introduce additional hyperparameters, such as the optimization step size and the number of optimization steps. Tuning these hyperparameters is crucial for training stability, as we found poorly chosen values often lead to unstable training. Second, training and inference are more computationally expensive than standard feed-forward models, requiring additional gradient computations. Third, while {\pa}s scale well up to 800M parameters, we have not explored larger models due to resource constraints. However, experimental scaling trends demonstrate that {\pa}s scale faster than existing paradigms during pretraining, suggesting that {\pa}s would perform better at foundation-model scale. Finally, {\pa}s currently struggle with data distributions that have many modes, such as class conditional image generation, likely due to the convex energy landscape assumption made during training.

\paragraph{Conclusion.} {\pa}s are the first instance of an approach that scales at a \textit{faster rate} than the Transformer++ during pretraining across both continuous and discrete modalities. Additionally, our results suggest that {\pa}s scale better than existing approaches during inference at System 2 Thinking by dynamically allocating computational resources and self-verifying their own predictions; these System 2 Thinking capabilities enable improved generalization to out-of-distribution data. Ultimately, the superior scaling in both training and inference, coupled with improved generalization, positions EBTs as a promising new paradigm shift for advancing the capabilities of future foundation models.


%% file: arxiv/main_sec/9_acknowledgements.tex
\section{Acknowledgements}
We extend special thanks to Jeonghwan Kim and Cheng Qian for their helpful discussions.
This research used the Delta and DeltaAI advanced computing and data resources, which are supported by the National Science Foundation (award OAC 2320345 and award OAC 2005572) and the State of Illinois. Delta and DeltaAI are joint efforts of the University of Illinois Urbana-Champaign and its National Center for Supercomputing Applications.

%% file: arxiv/supp_sec/_Appendix_Description.tex
In this appendix, we provide additional insight and details on EBTs. First, we provide more insight on the broader impact/future work of EBTs in Section~\ref{sec:future_broader}. Next, in Section~\ref{sec:additional_exp}, we include additional experiments. Then, we include additional approach details in Section~\ref{sec:training_details}, as well as additional experimental details in Section~\ref{sec:experimental_details}. After that, we include a comprehensive related work in Section~\ref{sec:additional_related_work}, additional facets of cognition in Section~\ref{sec:cognitive_facet_details}, and a discussion of counterarguments in Section~\ref{sec:counterarguments}. Finally, in the hopes of making EBTs more accessible to general audiences, in Section~\ref{sec:ebm_intro} we contain a general easier-to-understand intro to EBMs, and in Section~\ref{sec:tutorial} we describe a tutorial for getting started with EBTs.

%% file: arxiv/supp_sec/A_broader_impact_future_work.tex
\section{Future Works and Broader Impact }
\label{sec:future_broader}

{\pa}s, being qualitatively different from existing approaches, open several future research directions.


\subsection{Reversal Curse}
Recently, a phenomenon known as ``The Reversal Curse'' has been observed where LLMs fail to learn a symmetric mapping~\cite{berglund2023reversal} ``B is A'' despite learning ``A is B''. For example, LLMs trained on an example such as ``Q: Who is Tom Cruise’s mother? A: Mary Lee Pfeiffer'' often fail to generalize to know the answer to the reverse question of ``Who is Mary Lee Pfeiffer’s son?'' Remarkably, the Reversal Curse has manifested itself in LLMs regardless of the size or scale~\cite{berglund2023reversal}---probing researchers to investigate whether there are fundamental limitations to traditional feed-forward LLMs. One predicted cause of The Reversal Curse is the nature of gradient updates, where backpropagation only updates tokens within context. That is, while learning the mapping ``A is B'', none of B's tokens are within context, meaning they do not receive gradient updates when updating A's tokens. We hypothesize that LLMs trained with {\pa}s rather than standard feed-forward Transformers could help reduce this phenomenon, as with {\pa}s the tokens of A and B are within context during gradient updates due to predictions being made in the input space. Therefore, an exciting research direction would be investigating whether this hypothesis is correct in LLMs trained with {\pa}s, allowing improved generalization.

\subsection{Improved Stability}
In this work, we primarily trained {\pa}s with either two or three optimization steps. While these parameters worked well, we suspect that increasing the number of steps would improve the System 2 Thinking capabilities and the scaling during pretraining of {\pa}s, as more steps enable a longer ``thinking process'' before needing to converge. However, because of challenges in stability when training with more steps, due to a larger gradient computation graph, we were unable to successfully increase past two or three steps. Future work could focus more on extending the number of steps by studying ways to improve the training stability of EBMs.

\subsection{World Models}
In this work, we focus on autoregressive and bidirectional models over just state information (no actions). {\pa}s offer high promise in modeling states and actions due to the nature of EBMs learning a distribution over all possible inputs. Particularly, given a model trained to estimate the unnormalized joint distribution of the current context, future, as well as future actions, such world models could implicitly be used as policies to generate actions to achieve a specific state, similar to~\cite{janner2022planning, chi2023diffusion, zhou2024dino}. This would involve holding the current context (past states) constant, and minimizing the energy by propagating the gradient back to the action inputs and future state predictions. Thus, world models trained in this manner become capable of more than just predicting the future, but also in decision making to achieve a specific goal state.

\subsection{{\pa}s as Complementary Models}
\label{sec:complementary}
As demonstrated in~\cite{bakhtin2021residual, bhattacharyya2020energy}, EBMs can be used to improve the quality of generated text from language models. It's possible {\pa}s could be used in a similar manner for a broad variety of tasks, serving as the verifier of predictions initialized by standard feed-forward models. Therefore, although we do a side-by-side comparison to existing models in this work, {\pa}s could be \textbf{complementary} to existing model paradigms---being used as the System 2 Backbone for helping lighter models that perform System 1 thinking. 

There exist several current real-world use cases, such as low-latency LLM serving, where doing a single forward pass is sufficient, and where the added inference overhead of gradients with {\pa}s would not be worth the extra computation. However, we also envision a world in which people use {\pa}s for long-term System 2 Thinking to solve challenging problems. How much computation would it be worth dedicating to prove a long-standing mathematical conjecture, or figuring out a cure to cancer?

\subsection{Recurrent Energy-Based Models}
While {\pa}s scale well, for latency-driven use cases, Transformers require significantly more memory than Recurrent Neural Networks. Additionally, there is strong evidence for recurrence in the human brain~\cite{douglas2007recurrent}. Therefore, we anticipate that recurrent Energy-Based Models, possibly leveraging the Mamba architecture~\cite{gu2023mamba}, will eventually become common.

\subsection{Improved Thinking Algorithms}
The EBM thinking algorithms described have strong connections to or are derived from Markov Chain Monte Carlo (MCMC) sampling. Therefore, we broadly expect known MCMC samplers with more advanced techniques for traversing the energy landscape to be successful, such as Hamiltonian Monte Carlo~\cite{betancourt2017conceptual} or annealed Langevin dynamics~\cite{du2019implicit}. Additionally, we did not explore more advanced search algorithms such as Monte Carlo Tree Search, which we suspect could offer performance improvements and leave for future work.

\subsection{Multimodal Energy-Based Models}
We did not experiment with multimodal EBMs, however, EBMs offer several advantages for learning over multiple modalities. For example, multimodal EBMs would enable a single energy scalar to represent the alignment between modalities, and would simplify joint training across modalities by providing a unified objective that naturally captures inter‑modal dependencies.

\subsection{Thinking Scalability}
\label{sec:search}
Due to a lack of computational resources, we were unable to train models with more than $10^{21}$ FLOPs ($\approx 1300$ A100 GPU Hours). Therefore, training and thinking with {\pa}s remains untested at larger foundation model scale. We leave it to future work to scale with more GPUs and investigate the qualitative differences in training and thinking with {\pa}s.

\subsection{Learning Multimodal Distributions}
We found that {\pa}s, with the current training approach, struggle to capture distributions with many modes (e.g., unconditional image generation). Therefore, future work could explore approaches to improve the learning of distributions with many modes. More info is in Section~\ref{sec:failure_cases}.

\subsection{Understanding Predictions}
We believe that there exists a fundamental distinction between the internal representations associated with model inputs and outputs. Specifically, models generate internal representations \textbf{of} inputs, as these serve as the foundation that dictate the model's behavior at any given point in time. Conversely, as outputs do not affect a model's behavior at a given point in time, we contend that models construct representations \textbf{for predicting} (and not necessarily understanding) outputs. This distinction leads us to an interesting insight: \textit{models may not achieve a genuine understanding of outputs in the same way they understand inputs.} This implies that while models may develop an intricate understanding of input data, such understanding does not naturally extend to predictions that are made in the output space. Therefore, existing feed-forward models primarily making predictions in the output space may not \textbf{understand} their predictions in the same way they understand their inputs. This intuition further supports the principles behind {\pa}, where predictions are made in the input space, enabling representations \textbf{of predictions} to be developed. We leave the investigation of this hypothesis to future work.


%% file: arxiv/supp_sec/B_additional_experiments.tex
\section{Additional Experimentation}
\label{sec:additional_exp}

\subsection{Additional Natural Language Processing Experiments}

\begin{figure}[t]
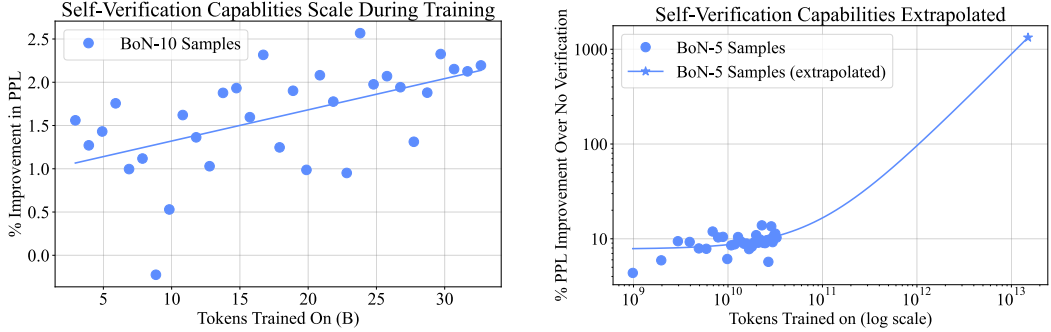

  \centering
  \scriptsize
  \begin{subfigure}[b]{0.48\columnwidth}
    \includesvg[width=\linewidth]{images/scaling_thinking_nlp_ar_data_linear.svg}
    \caption{Self-verification with BoN-10 versus BoN-2.}
    \label{subfig:self_verification_non_adversarial}
  \end{subfigure}\hfill
  \begin{subfigure}[b]{0.48\columnwidth}
    \includesvg[width=\linewidth]{images/scaling_thinking_nlp_ar_data_ood_linear_projected.svg}
    \caption{Results in Fig.~\ref{subfig:thinking_scaling} projected to Llama3 scale~\cite{grattafiori2024llama}.}
    \label{subfig:thinking_scaling_extrapolated}
  \end{subfigure}
  \caption{\textbf{{\pa} Thinking Analysis for Data Scaling.} (a) Self-verification of BoN-2 compared to BoN-10. EBTs become less adversarial and thus benefit more from verifying an increasing number of samples during training. (b) A projection of the results from Figure~\ref{subfig:thinking_scaling} to the data scale of Llama3~\cite{grattafiori2024llama}, demonstrating that as data scale increases, improvements from self-verification can lead to potentially massive performance increases from System 2 Thinking.}
  \label{fig:nlp_ar_think_scale}
  \vspace{-20pt}
\end{figure}

We conduct experiments to confirm hypotheses on thinking results obtained in the main paper. 
First, we confirm that {\pa}s become less adversarial with scale by comparing the performance of using BoN with $2$ versus $10$ samples. The results in Figure~\ref{subfig:self_verification_non_adversarial} demonstrate that when models are trained on less tokens, there is little performance improvement by verifying $10$ samples instead of just $2$. In fact, verifying $10$ samples occasionally leads to \textit{worse} performance than verifying $2$ samples, likely because the EBT found an adversarial sample (a sample with low energy that is in fact not a good prediction). However, as data scale increases we observe that performance improvements from BoN-10 versus BoN-2 increase, and that these adversarial dynamics decrease. Together, these results suggest that with scale {\pa}s become less adversarial due to an improved energy landscape.
In an effort to understand the impacts of thinking at the scale of modern foundation models, we project results from Figure~\ref{subfig:thinking_scaling} to the scale of modern foundation models~\cite{grattafiori2024llama} to extrapolate a projected performance gain from self-verification based thinking. The results are visualized in Figure~\ref{subfig:thinking_scaling_extrapolated}, where they demonstrate that, because of the $1000 \times$ data scale of modern foundation models, the performance improvement from self-verification increases drastically.

\begin{figure}[t]
    \centering
    \includesvg[width=0.95\textwidth]{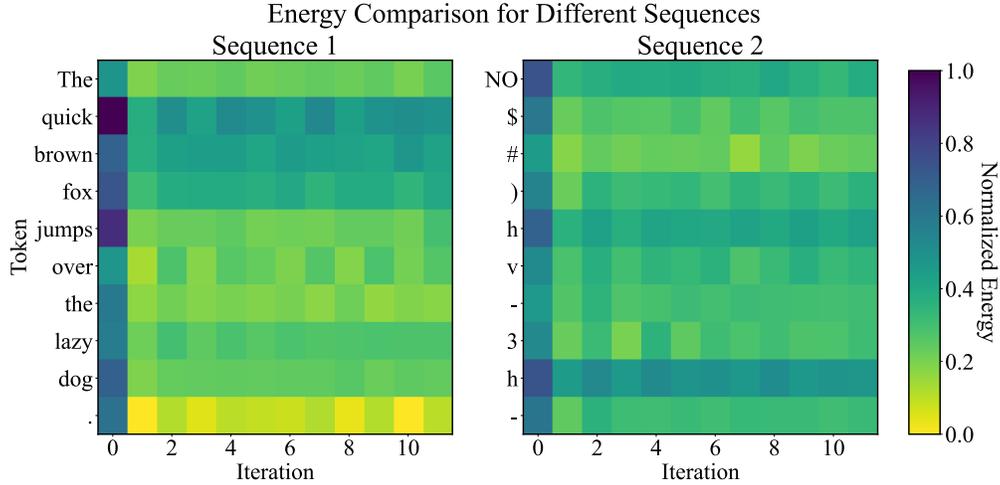}
    \caption{\textbf{Epistemic Uncertainty Comparison.} EBTs learn to express epistemic uncertainty (uncertainty related to lack of knowledge) on unseen data. Particularly, the sequence on the left, which is a text sequence likely seen during training, has consistently lower energy (uncertainty) for tokens than the sequence on the right, which is a random text sequence not from the training distribution. This demonstrates that {\pa}s learn to ``know what they don't know.''}
    \label{fig:qualitative_epistemic_uncertainty}
    \vspace{-10pt}
\end{figure}

Additionally, in an effort to understand whether {\pa}s can capture epistemic uncertainty (uncertainty related to knowledge), in addition to aleatoric uncertainty (uncertainty related to inherent data randomness), we visualize the energies of different token sequences that are in versus Out-Of-Distribution (OOD) in Figure~\ref{fig:qualitative_epistemic_uncertainty}. The results demonstrate that, for a more in-distribution sequence, {\pa}s have lower energy (less uncertainty), than for an OOD sequence. This suggests that {\pa}s learn to \textit{know what they don't know}, as they learn to have higher energy for OOD sequences signifying harder predictability.

\begin{figure}[t]
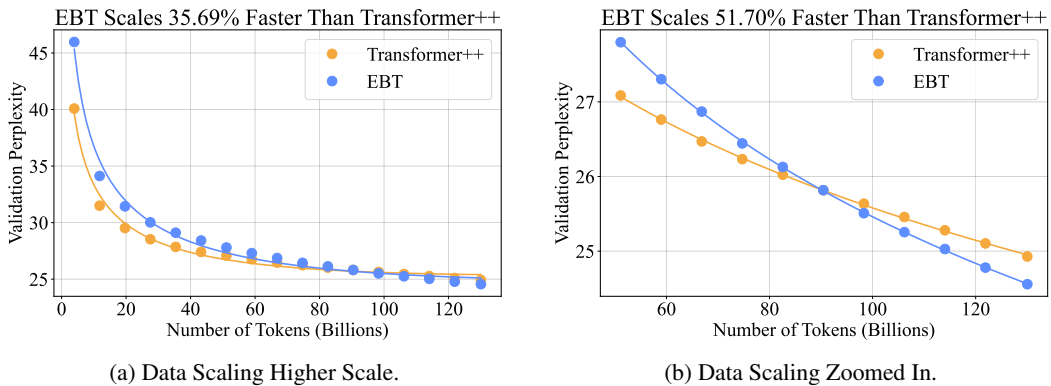

  \centering
  \scriptsize
  \begin{subfigure}[b]{0.48\columnwidth}
    \includesvg[width=\linewidth]{images/scaling_learning_nlp_ar_data_smallm1.svg}
    \caption{Data Scaling Higher Scale.}
  \end{subfigure}\hfill
  \begin{subfigure}[b]{0.48\columnwidth}
    \includesvg[width=\linewidth]{images/scaling_learning_nlp_ar_data_smallm2.svg}
    \caption{Data Scaling Zoomed In.}
  \end{subfigure}
  \caption{\textbf{EBT Larger Scale Data Scaling.} (a) Data scaling results from Figure~\ref{subfig:data_scaling} scaled up to a larger model size, higher context length, increased batch size, and a higher number of training steps. (b) Results from (a) zoomed in.}
  \label{fig:data_scaling_larger_scale}
  \vspace{-10pt}
\end{figure}

To confirm some of the scaling trends in the paper, as well as experiment with additional datasets, we conduct larger-scale experiments on the FineWeb dataset~\cite{penedo2024fineweb}. Particularly, for Figure~\ref{subfig:data_scaling}, we confirm that the generally better data scaling of EBTs compared to the Transformer++ holds at increased scale. We confirm this by training small-sized models with a batch size of $256$, a context length of $1024$, and for $500,000$ training steps. These results are visualized in Figure~\ref{fig:data_scaling_larger_scale}, where we observe that EBTs still continue to outscale the Transformer++ by as much as $35\%$ in scaling rate. These results further reinforce that EBTs are more data-efficient than the Transformer++.

\subsection{{\pa} Failure Cases}
\label{sec:failure_cases}
During experimentation, along with image denoising experiments, we also conducted small-scale experiments with text-to-image generation. We found that, in datasets such as COCO with many different modes for a single condition (e.g., hundreds of images with a caption similar to ``giraffe with a long neck''), that {\pa}s did not learn to generate high-quality novel images. Instead, {\pa}s often generated blurred images similar to the training distribution. We believe this is caused by the training approach pushing the energy landscape to be convex surrounding the training examples (modes). Therefore, when there are many different modes within the same region (same condition), this convex energy landscape ``merges'' to one landscape averaged around the different modes, resulting in blurriness. We believe that this is not a fundamental limitation of {\pa}s, and that future work could address this issue, such as by adding more conditioning.

%% file: arxiv/supp_sec/C_Approach_Details.tex
\section{Additional {\pa} Details}
\label{sec:training_details}

\subsection{Formalizing Thinking}
\label{sec:thinking_formalization}
Due to the recent surge of interest in scaling the performance of models during inference/test time, there are several common terms used to refer to these ideas. These terms include scaling the thinking capabilities of models~\cite{jaech2024openai}, inference-time scaling~\cite{ma2025inference}, inference-time compute~\cite{manvi2024adaptive}, and test-time compute~\cite{jaech2024openai, snell2024scaling}. Therefore, to reduce confusion stemming from a wide variety of terminology and unite the community, in this work we broadly define these concepts as \textbf{System Two Thinking} or more concisely \textbf{Thinking}. We formalize improvements made by Thinking as the following: \\

\begin{definition}[System 2 Thinking]
\label{def:system_2_thinking}
\textit{Given a problem with data $x$, a model $\theta$, and additional computational resources in the form of function evaluations $F$ greater than the minimum number of function evaluations to get a valid prediction from the model $F_0$, \textbf{System Two Thinking} $\text{STT}(\cdot)$ quantifies the expected percentage improvement in performance as $F$ increases. Let $P(x, \theta, F)$ be the performance on input $x$ when the model $\theta$ uses $F$ function evaluations:}
\[
    \text{STT}(x, \theta, F) \;=\; \mathbb{E}_x \left[
        \frac{P(x, \theta, F)}{P(x, \theta, F_0)} \;-\; 1
    \right],
\]
\end{definition}

This formalization is compatible with any type of metric (e.g., Accuracy, Perplexity, FID, etc), and uses more psychology-aligned terminology~\cite{kahneman2011thinking}. Further, avoiding terms such as ``inference'' or ``test-time'' makes the idea of Thinking more compatible with domains where the line between inference and training is blurry, such as real-world continual learning, domain adaptation, or actual human learning/thinking processes~\cite{parr2022active}. Just as \textbf{learning} has become a flexible term across machine learning representing several different ideas, we intend for \textbf{thinking} to similarly unify many diverse ideas under a common framework. For a greater justification on this perspective, please see Section~\ref{sec:counterarguments}

\subsection{Energy-Based Transformer ({\pa}) Thinking Types}
\label{sec:ebt_types}

\begin{figure}[t]
    \begin{center}
    \scriptsize
    \includesvg[width=0.48\columnwidth]{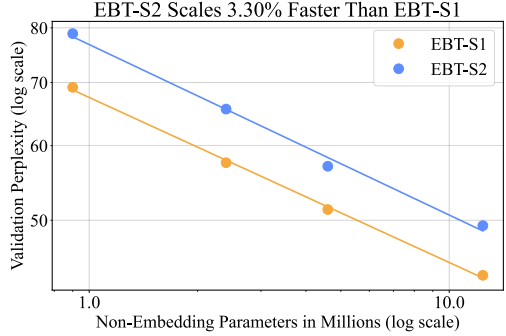} \\
    \caption{\textbf{EBT S1 and S2 Scaling Comparison.} Scaling rate of System 1 (S1) compared to System 2 (S2) models. System 2 models have a higher Y-intercept but scale slightly faster than System 1 models. Therefore, as the scaling rate ultimately dominates asymptotic scaling behavior, and not the Y-intercept, scaling results in the main paper that hold for S1 models should generally hold for S2 models.}
    \label{fig:s1_vs_s2}
    \end{center}
\end{figure}

All experiments in the paper are conducted with two main variants of {\pa}s, which we call System 1 (S1) and System 2 (S2) {\pa}s. S1-{\pa}s have hyperparameters specifically optimized for stability and learning convergence, whereas S2-{\pa}s have hyperparameters optimized for System 2 Thinking capabilities. Many of the pretraining scaling experiments conducted in Section~\ref{sec:experimentation} are with S1 models as to reduce the computational resources required for experimentation. To confirm that these results hold for S2 models, we plot the scaling trends of S1 and S2 models side by side; in Figure~\ref{fig:s1_vs_s2}, we find that S2 models scale \textit{at the same or a higher rate} than S1 models during training, but have a higher Y-intercept. This Y-intercept offset does not affect asymptotic scaling behavior (as asymptotically the scaling rate dominates), and hence, scaling trends that hold for S1 models should generally hold for S2 models. In fact, S2 models may even perform better than S1 models during pretraining asymptotically because of the higher scaling rate. Intuitively, switching from S1 to S2 models allows for a compute trade-off between the model's pretraining performance and the model's System 2 Thinking capabilities.

S1 models have the gradient of predictions detached between optimization steps to increase training stability. Conversely, following~\cite{du2024learning}, the S2 models truncate backpropagation and avoid detaching prediction tensors between optimization steps. Additionally, the S2 models have all the energy landscape regularization techniques described in Section~\ref{sec:thinking_approach}, whereas the S1 models have none. We also find that S2 models require a different value for the optimization step size, that the optimization step size not be learned, and to perform a minimum number of optimization steps greater than one.

\subsection{Autoregressive Causal Energy-Based Transformers Efficient Implementation}
\label{sec:ebt_full}
GPT-style decoder-only autoregressive Transformers are parallelizable due to making all next-state predictions simultaneously~\cite{radford2019language}. In this section, we detail the implementation of decoder-only autoregressive EBTs, which also parallelize all next-state predictions simultaneously, but involve more complexity due to EBMs making predictions in the input space rather than the output space.\footnote{For simplicity we often use the term matrix to refer to slices of tensors.} 
To demonstrate why this poses a challenge, for the decoder-only autoregressive Transformer, consider the case of the \(N \times N\) attention scores matrix after the causal mask has been applied:
\[
\text{scores} = 
\begin{bmatrix}
\alpha_{z_1,z_1} & 0 & \ldots & 0 \\
\alpha_{z_2,z_1} & \alpha_{z_2,z_2} & \ldots & 0 \\
\ldots & \ldots & \ldots & \ldots \\
\alpha_{z_n,z_1} & \alpha_{z_n,z_2} & \ldots & \alpha_{z_n,z_n} \\
\end{bmatrix},
\]
where \( \alpha_{z_i,z_j} \) represents the attention score (probability mass) from observed state \( z_i \) to observed state \( z_j \). Now, in the case of an EBM, where predictions are made in the input space, the desired \(N \times (N + 1) \) attention scores matrix would look like the following:
\begin{equation}
\text{scores} = 
\begin{bmatrix}
\alpha_{z_1,z_1} & \alpha_{z_1,\hat{z}_2} & 0 & \ldots & 0 \\
\alpha_{z_2,z_1} & \alpha_{z_2,z_2} & \alpha_{z_2,\hat{z}_3} & \ldots & 0 \\
\ldots & \ldots & \ldots & \ldots & \ldots \\
\alpha_{z_n,z_1} & \alpha_{z_n,z_2} & \alpha_{z_n,z_3} & \ldots & \alpha_{z_n,\hat{z}_{n+1}} \\
\end{bmatrix},      \label{math:scores_ebm}
\end{equation}
where \(\hat{z}_j\) denotes a predicted state (as opposed to an observed state). This is challenging to compute because each \(\hat{z}_j\) along the superdiagonal is unique for its row, as we chose for each prediction to be made independently from each other prediction.\footnote{We note that it's possible to make each prediction not be independent, however, not making every prediction independent would mean that there is stochasticity for each prediction caused not only by its initial value but also by all initial values of all previous predictions.} Consequently, unlike standard attention in feed-forward Transformers, this attention matrix cannot be computed with a matrix multiplication (\(\text{softmax}\left( \frac{QK^T}{\sqrt{d_k}} \right)\)), as every element of the superdiagonal is a prediction and not an observed state. 

Additionally, in traditional transformers, if the context length is $N$, the size of manipulated tensors will generally be $B \times N \times D$ where $B$ is the batch size and $D$ is the embedding dimension. However, because EBMs make predictions in the input space, the input tensor needs to be different to allow for inputting predictions. Therefore, to distinguish these tensors, for a context length of $N$ tokens (meaning we are given $N$ tokens), we define a sequence of the first $N$ elements as \(z_o\), or the observed sequence representations, and the final \(N\) elements as \(z_p\), or the predicted representations (for each next element in the sequence). This means that the manipulated tensors within the EBT are of size $B \times 2 N \times D$ for a context length of size $N$.

The values for \(z_o\), or the observed states, are computed identically as in the original transformer~\cite{vaswani2017attention}, as the attention scores of the observed states do not depend on the predicted states. This can be formalized as the following:
\[
\text{Attention}(Q_o, K_o, V_o)_{z_o} = \text{softmax}\left(\frac{Q_oK_o^T}{\sqrt{d_k}}\right)V_o,
\]
where $Q_o$, $K_o$, and $V_o$ are the Query, Key, and Value matrices of the observed states \(z_o\). In every block of the transformer, the representations of all observed states are updated in this manner, independent of the representations of the predictions. \\

For the computation of prediction representations, three matrices are computed, which we notate as $Q_p$, $K_p$, and $V_p$.\footnote{In practice, we shared the weights for the Q/K/V matrices for both observations and predictions to enable a one-to-one comparison to existing feed-forward transformers. It would be interesting to experiment with using different weight matrices to determine if that improves convergence and generalization at the cost of more parameters.} First, we compute the self-attention scores of all predicted representations to all observed representations:
\[
\tilde{S}_{z_o\leftarrow z_p} = \frac{Q_pK_o^T}{\sqrt{d_k}},
\]
where \(\tilde{S}_{z_o\leftarrow z_p}\) denotes the raw unnormalized attention scores of the predicted states to the observed states. Note, however, that the self-attention scores of each prediction with itself have not yet been calculated---due to the key matrix being from the observed states. Therefore, the superdiagonal of the \(\tilde{S}_{z_o\leftarrow z_p}\) matrix needs to be replaced with the self-attention scores of each prediction with itself to achieve the attention score matrix shown in Equation~\ref{math:scores_ebm}.

To achieve this matrix, we first need to append a column to the right side of the \(\tilde{S}_{z_o\leftarrow z_p}\) matrix, as the size of the matrix is currently $N \times N$, but the matrix needs to be $N \times (N+1)$ ($N$ for the observed states and then one for the predicted states along the second dimension). After doing this, we first mask out the superdiagonal with zeros to ensure that the probabilities in the superdiagonal of the score matrix only correspond to the values of predicted states with themselves. To make this operation differentiable, this masking operation is done through elementwise multiplication of a matrix with ones everywhere except the superdiagonal, which has zeros. Then, we compute the self-attention scores of each predicted state with itself, using the following equation: \\
\[
\tilde{S}_{z_p \leftarrow z_p} = \frac{\text{sum}(Q_p * K_p)}{\sqrt{d_k}},
\]
where the $*$ indicates the Hadamard product and the sum is across the feature dimension. Using a superdiagonal mask again, we set the diagonal of the \(\tilde{S}_{z_o\leftarrow z_p}\) to these values. We denote the updated matrix as \(\tilde{S}_{z_p}\), representing that this matrix is still unnormalized scores but takes into account interaction between $z_p$ and $z_o$ as well as from $z_p$ to themselves. Now, after applying the softmax: \\
\[
S_{z_p} = \text{softmax}\left( \tilde{S}_{z_p} \right),
\]
we have the intended scores matrix shown in Equation~\ref{math:scores_ebm}. Note that for this softmax operation we adjust the standard causal attention mask to ensure all future information is masked out and the superdiagonal we added is not masked out. 

One more barrier towards finally extracting all updated \(z_p\) representations is the fact that we cannot simply multiply this resulting scores matrix by the values matrix, as each element of the superdiagonal corresponds to a different predicted state. Thus, using similar techniques to before, we first clone and then extract the superdiagonal from this scores matrix using a diagonal mask.

After extracting out the superdiagonal, zeroing it out after extraction, and removing the earlier appended right-most column, we can multiply the resulting scores matrix by the $V_o$ matrix to get all of the representations summed together of each predicted state with all observed states. This is computed using the following matrix multiplication: \\
\[
z_p = S_{z_p} \cdot V_o.
\]
As we also need to add the representation of each predicted state weighted with its own attention score (what was extracted on the superdiagonal), we perform another Hadamard product of the \(V_p\) matrix with the cloned superdiagonal to get these values, and then add these element wise to the \(z_p\) representations. Now, we have computed the intended representations involving the scores matrix shown in Equation~\ref{math:scores_ebm}. Now, \(z_o\) and \(z_p\) are updated by multiplying these tensors by the shared output weight matrix $W_o$.

\subsection{Autoregressive Energy-Based Transformers Simplified Implementation}
A more simplified implementation involves the entire attention matrices and a generalized causal mask, as described in~\cite{deng2024causal}. However, because this implementation involves a matrix multiplication with $2$ times the sequence length, this results in $4$ times the number of FLOPs as normal attention, which is around double the number of FLOPs of our more efficient implementation.

%% file: arxiv/supp_sec/D_Experimental_Details.tex
\section{Experimentation Details}
\label{sec:experimental_details}

Tables~\ref{tab:hparams_baseline} and~\ref{tab:hparams_ebt} specify general model information and hyperparameters. We utilized the Llama 2 transformer implementation~\cite{touvron2023llama} for the Transformer++ and used this implementation as the backbone upon which we built causal decoder-only {\parch}s. We seed all libraries using PyTorch Lightning~\cite{falcon2019pytorch} for all experiments with a seed of $33$. For the Diffusion Transformer, we use the implementation from~\cite{peebles2023scalable}---for the bidirectional EBT we build upon this implementation.

Unless otherwise stated, for all scaling experiments we copy the popular Mamba paper pretraining settings for the model configuration (we use different batch sizes/data due to computational constraints), shown in Table~\ref{tab:scaling_model_sizes}. Because we are compute-constrained, we also include two extra model sizes in this table, extra extra small (xxs) and extra small (xs), which allow us to collect more data points for the extrapolation of scaling trends. Because the hyperparameters used in these experiments were tuned for feed-forward Transformers in~\cite{kaplan2020scaling}, we broadly expect that hyperparameters tuned for EBTs will further increase the performance gap between EBTs and the Transformer++ in autoregressive modeling. Future research should investigate the optimal width/depth/batch sizes for EBTs.

\begin{table}[t]
\captionsetup[table]{position=top}
\caption{\textbf{Model sizes and hyperparameters for scaling experiments.} For applicable model sizes we follow Mamba~\cite{gu2023mamba}.}
\centering
\small
\begin{tabular}{l c c c c}
\toprule
Size  & Non-Embedding Params & \# layers & embed.\ dim & \# heads \\
\midrule
xxs    & 6.18M  & 6  & 384  & 6  \\
xs     & 12.4M  & 12 & 384  & 6  \\
small  & 48.8M  & 12 & 768  & 12 \\
medium & 176M   & 24 & 1024 & 16 \\
large  & 396M   & 24 & 1536 & 16 \\
xl     & 708M   & 24 & 2048 & 32 \\
\bottomrule
\label{tab:scaling_model_sizes}
\end{tabular}
\end{table}

\begin{table*}[t]

\caption{\textbf{Hyperparameters for Transformer++.}}
\centering
\small
\begin{tabular}{lcc}
\toprule
\textbf{Hyperparameter} & \textbf{CV} & \textbf{NLP} \\
\toprule
Optimizer & \multicolumn{2}{c}{AdamW} \\
Optimizer Momentum & \multicolumn{2}{c}{$\beta_1,\beta_2=0.9,0.999$} \\
LR Schedule & \multicolumn{2}{c}{Linear warm up cosine decay} \\
Warmup steps & \multicolumn{2}{c}{$1e4$} \\
Minimum LR Scale & \multicolumn{2}{c}{$10$} \\
Gradient Clip Value & \multicolumn{2}{c}{$1$} \\
Weight Decay & \multicolumn{2}{c}{$0.01$} \\
Context Length & $16$ & $256$ \\
Encoder & SD-XL VAE~\cite{rombach2022high, stabilityai/sd-vae-ft-mse} & - \\
Image Dimension & $224$x$224$ & - \\
Tokenizer & - & EleutherAI/gpt-neox-20b~\cite{black2022gptneox20b} \\
Vocab Size & - & $50277$ \\
\bottomrule
\end{tabular}
\label{tab:hparams_baseline}
\end{table*}

\subsection{Autoregressive Language Modeling Experimental Details}

\subsubsection{Autoregressive Language Modeling Learning Scalability Details}
\label{sec:scaling_law_extra_info}
Conducting thorough scaling experiments is extremely challenging---a recent survey on ``scaling laws''~\cite{li2025mis} showed just how fragile many of these ``scaling laws'' are to hyperparameters, data, and other parameters, and how changing these variables slightly can lead to vastly different conclusions. Being bottlenecked by a limited set of computing resources further exacerbates the issue of comparing two models. Therefore, we sought out to conduct controlled experiments that revealed the most information possible regarding the scaling of {\pa}s compared to different models.

Most existing works study scaling by varying several factors simultaneously, including depth (number of transformer blocks)~\cite{kaplan2020scaling}, width (embedding dimension)~\cite{kaplan2020scaling}, possibly batch size~\cite{chen2024scaling, hu2023predicting}, and the amount of data~\cite{kaplan2020scaling}. Therefore, to be more comprehensive in determining when {\pa}s scale differently than the Transformer++, we decided to conduct normal scaling experiments over all of these factors at the exact same time (as is standard), \textit{as well as ablating over changing just one of these factors at a time}. Notably, conducting experiments in this manner allows for controlling a single independent variable at a time (i.e., just changing the number of Transformer Blocks), which allows for stronger conclusions regarding what aspects of model scaling different models perform better over (i.e., {\pa}s scale better then the Transformer++ when increasing the number of Transformer Blocks). Scaling all factors at once does not allow for such insight, as there are many independent variables changing at once. We believe experiments that involve changing a single independent variable are in broader alignment with traditional scientific principles.

For the parameter and FLOP scalability experiments, all models are pretrained for $105k$ steps. For each model size, following~\cite{chen2024scaling, hu2023predicting}, we scale the batch size. For NLP experiments we use the following batch sizes for the xxs, xs, small, medium, and large models respectively: $32$, $46$, $90$, $170$, and $256$. These batch sizes were determined by scaling the batch size with the square root of the number of parameters and then rounding to an even number, similar to batch size trends in~\cite{hu2023predicting}.
All model sizes use the same learning rate as in the Mamba paper~\cite{gu2023mamba} ($0.0006$, $0.0003$, $0.00025$, and $0.0002$ for small, medium, large, and xl models respectively), where for the xxs and xs models we use a learning rate of $0.0012$ and $0.0009$ respectively. For the data scaling and batch size scaling experiments conducted in Figure~\ref{fig:nlp_ar_learn_scale_1}, we use xxs models (we also report small models in Section~\ref{sec:additional_exp}). The data scaling experiment used a batch size of $128$ and the batch size experiments ran for $105k$ steps. All models were trained with a context length of $256$ and no FFN multiplier (FFN dimension being equal to the embedding dimension) due to limited resources.

\subsubsection{Autoregressive Language Modeling Thinking Scalability Details}
We train xxs S2-EBT and Transformer++ models with the same setup as above, with the exception that models are trained with a batch size of $128$ for 1M training steps (Table~\ref{tab:hparams_ebt_thinking} contains more hyperparameter details). Increasing the data scale enables us to better understand how thinking scales during pretraining. It's worth noting that since we are training small language models, they could not benefit from modern techniques such as Chain of Thought (CoT) in improving performance. Figures~\ref{fig:best_result},~\ref{fig:nlp_ar_think_scale},~\ref{fig:ood_generalization_thinking} and Table~\ref{tab:generalization_exps} use the best checkpoints of these two models. Figures~\ref{subfig:ood_thinking_comparison},~\ref{fig:ood_generalization_thinking} both use all four downstream datasets shown in Table~\ref{tab:generalization_exps}. Figure~\ref{fig:ood_generalization_thinking} also uses the pretraining dataset in addition to these downstream datasets, where every dataset is represented as a separate OOD point. 

Figure~\ref{subfig:thinking_scaling} was solely from the Dyck Languages benchmark. We did not observe this trend in other benchmarks, possibly due to perplexity not being a completely linear metric. Figure~\ref{subfig:self_verification_non_adversarial} was on the RedPajamaV2 validation dataset.

\subsection{Autoregressive Video Experimental Details}
We use the same model parameter scaling, shown in Table~\ref{tab:scaling_model_sizes}, as the NLP experiments. We also used a batch size of $256$ for all models, as we found that it did not affect the scaling performance due to models training for many epochs. We also processed videos with $0.25$ seconds between frames. For the Transformer++ baseline, we use the same learning rates as the NLP experiments. For EBTs, we found that it was necessary to use a lower learning rate by a factor of $3$ for proper training stability. We use the standard SSV2 train and validation split for experiments. Other hyperparameters are shown in Figure~\ref{tab:hparams_baseline} and Figure~\ref{tab:hparams_ebt}.

\begin{table*}[t]
\caption{\textbf{Hyperparameters for \textbf{\pa} scaling experiments.}}
\centering
\small
\begin{tabular}{lcc}
\toprule
\textbf{Hyperparameter} & \textbf{CV} & \textbf{NLP} \\
\toprule
Optimizer & \multicolumn{2}{c}{AdamW} \\
Optimizer Momentum & \multicolumn{2}{c}{$\beta_1,\beta_2=0.9,0.999$} \\
LR Schedule & \multicolumn{2}{c}{Linear warm up cosine decay} \\
Warmup Steps & \multicolumn{2}{c}{$1e4$} \\
Minimum LR Scale & \multicolumn{2}{c}{$10$} \\
Gradient Clip Value & \multicolumn{2}{c}{$1$} \\
Weight Decay & \multicolumn{2}{c}{$0.01$} \\
Context Length & $16$ & $256$ \\

Encoder & SD-XL VAE~\cite{rombach2022high, stabilityai/sd-vae-ft-mse} & - \\
Image Dimension & $224$x$224$ & - \\
Tokenizer & - & EleutherAI/gpt-neox-20b~\cite{black2022gptneox20b} \\
Vocab Size & - & $50277$ \\
Normalize Input Distribution & - & \cmark \\

Optimization Steps & $2$ & $2$ \\
Optimization Step Size & $30,000$ & $500$ \\
Optimization Step Size LR Multiplier & $90,000$ & $1,500$ \\
Learnable Optimization Step Size & \multicolumn{2}{c}{\cmark} \\

\bottomrule
\end{tabular}
\label{tab:hparams_ebt}
\end{table*}

\subsection{Bidirectional Image Denoising Experimental Details}
We use the COCO 2014 dataset~\cite{lin2014microsoft, AbdoTW/COCO_2014DatasetsHF} with $128$ by $128$ images, its train/validation split, a patch size of $16$, and the Diffusion Transformer implementation from~\cite{peebles2023scalable}. All models were trained using the large model size described in Table~\ref{tab:scaling_model_sizes}, with a learning rate of $1e-4$ for $100,000$ steps. For the DiT baseline, we used the same hyperparameters from~\cite{peebles2023scalable}, changing only the batch size to $128$ from $256$. We based our bidirectional EBT implementation on the code from this repository. We experimented with several different diffusion inference strategies to ensure a fair comparison, including DDPM, DDIM, increasing the number of diffusion steps at inference, and recursively applying the diffusion model on its own denoised output.\footnote{This recursive application was only performed during testing due to the distribution shift.} Ultimately, we found that the combination of DDIM recursively applied on its own output performed best, hence we used this as the baseline in all experiments. We used the default denoising schedule from the DiT codebase~\cite{peebles2023scalable}. 


For both DiTs and EBTs, we found that models performed best on OOD noise levels when denoising their own outputs twice, that is applying the model to denoise the image three times recursively. This is how we are able to achieve the results in Figure~\ref{fig:thinking_fwd_passes_img} demonstrating the performance for $300$ forward passes from DiTs and $3$ forward passes from EBTs. We found that for image denoising, it was not necessary to train EBTs with the S2 hyperparameters for System 2 Capabilities to emerge, although its possible these would further improve performance. Additionally, for image classification, for both models, we take the average of all the final patch tokens, and for DiTs we feed in $T = 0$.

\begin{table*}[t]
\caption{\textbf{Hyperparameters for {\pa} System 2 Thinking experiments.}}
\centering
\small
\begin{tabular}{l c}
\toprule
\textbf{Hyperparameter} & \textbf{NLP} \\
\midrule
Optimizer & AdamW \\
Optimizer Momentum ($\beta_1,\beta_2$) & 0.9, 0.999 \\
Model Size & Small \\
Learning Rate (LR) & 0.0012 \\
Learning Rate (LR) Schedule & Linear warm-up + cosine decay \\
Warmup Steps & $1\times10^4$ \\
Minimum LR Scale & 10 \\
Gradient Clip Value & 1 \\
Weight Decay & 0.01 \\
Context Length & 256 \\
Tokenizer & EleutherAI/gpt-neox-20b~\cite{black2022gptneox20b} \\
Vocab Size & 50,277 \\
Normalize Input Distribution & \cmark \\
Optimization Steps & 2-3 (randomized) \\
Optimization Step Size & 5 \\
Optimization Step Size Multiplicative Random Factor & 2 \\
Langevin Dynamics Noise & 3 \\
Learnable Optimization Step Size & \xmark \\
No Detach Between Optimization Steps & \cmark \\
Truncate Optimization & \cmark \\
Replay Buffer & \cmark \\
\bottomrule
\end{tabular}
\label{tab:hparams_ebt_thinking}
\end{table*}

\subsection{Computational Resources}
All experiments were conducted on either Nvidia A100s, H100s or GH200s, with the largest scale experiment requiring approximately $\approx 1300$ A100 GPU Hours. The runtime for each experiment was dependent on the model sizes used as well as the amount of data trained on.

\subsection{FLOP Calculations}
\label{sec:flop_calc}

We adopt the standard estimate of \(6N\) FLOPs per token for the AR Transformer++~\cite{casson2023transformerflops}, where \(N\) denotes the number of non‐embedding parameters. For AR EBTs, however, the per‐token cost varies with the number of training optimization steps and chosen hyperparameters.

To derive the FLOPs for EBT training, we follow~\cite{dagréou2024howtocompute} for the Hessian‐vector product (HVP), which EBTs require to backpropagate through a first‐order derivative. Since an HVP has the same theoretical complexity as a gradient computation, we express the per‐step FLOPs as
\begin{align*}
\text{FLOPs} = F + B + B.
\end{align*}
Based on~\cite{casson2023transformerflops}, the forward and backward passes require approximately \(2N\) and \(4N\) FLOPs per token, respectively, where \(N\) denotes the count of non‑embedding parameters in the Transformer. In the autoregressive {\pa} implementation, the effective sequence length becomes twice that of the original Transformer (formally \(2S - 2\) for an original sequence length \(S\)). Owing to the efficient scheme of Section~\ref{sec:ebt_full}, this doubling of sequence length translates roughly into a two‑fold increase in FLOPs, rather than a four-fold increase in FLOPs. Hence, each second‑order optimization step demands roughly 
\begin{align*}
(F + B + B) \times 2 = (2N + 4N + 4N)\times 2 = 10N \times 2,
\end{align*}
making it \(\approx3.33\times\) more expensive than a standard feed‑forward Transformer step.

The overall FLOP count also depends on one’s choice of hyperparameters. For S1 models, where the loss is evaluated at every iteration without gradient truncation, the total FLOPs simply multiply by the number of steps. Therefore, for our pretraining experiments using two optimization steps, we get that {\pa}s used \(6.66\times\) the FLOPs of a comparable Transformer++ during training. In contrast, for S2 models, a random number of optimization steps is used, the gradient is truncated, the loss is only calculated at the last step following~\cite{du2022learning}, and a Replay Buffer is used. Therefore, the FLOP count varies and can both decrease (as truncating uses less FLOPs for earlier steps) as well as increase (as using more steps and a replay buffer both use more FLOPs). These numbers also vary during inference, where the full EBT implementation parallelizing all predictions at once is not necessary.

Given the scarcity of published methods for computing higher‑order derivative FLOPs and our inability to leverage existing libraries for Hessian‑vector products, these estimates remain approximate. We welcome corrections or additional insights from readers familiar with FLOP calculations for second‑order methods.

%% file: arxiv/supp_sec/E_Additional_Related_Works.tex
\section{Additional Related Works}
\label{sec:additional_related_work}

\subsection{Energy-Based Models (EBMs) as a Generalization of Diffusion Models and Recurrent Depth Models}

Both diffusion models~\cite{du2023reduce} and RNNs~\cite{saunshi2025reasoning} can be seen as predicting the score, or the gradient of the energy function/data density, $\nabla_x E_{\theta}(x)$, where diffusion models do this with an additional time condition~\cite{liu2022compositional}. Thus, the largest benefit of explicit EBMs over these approaches, which can be seen as implicit EBMs (due to only implicitly defining an energy function), is that using an explicit EBM allows for explicit verification/likelihood modeling. We show that this enables the use of self-verification to improve predictions, whereas with diffusion models and RNNs an additional verifier model is necessary to achieve this capability~\cite{ma2025inference}. 

\begin{figure}
    \centering
    \scriptsize
    \begin{subfigure}[t]{0.48\columnwidth}
      \includegraphics[valign=t,width=\linewidth]{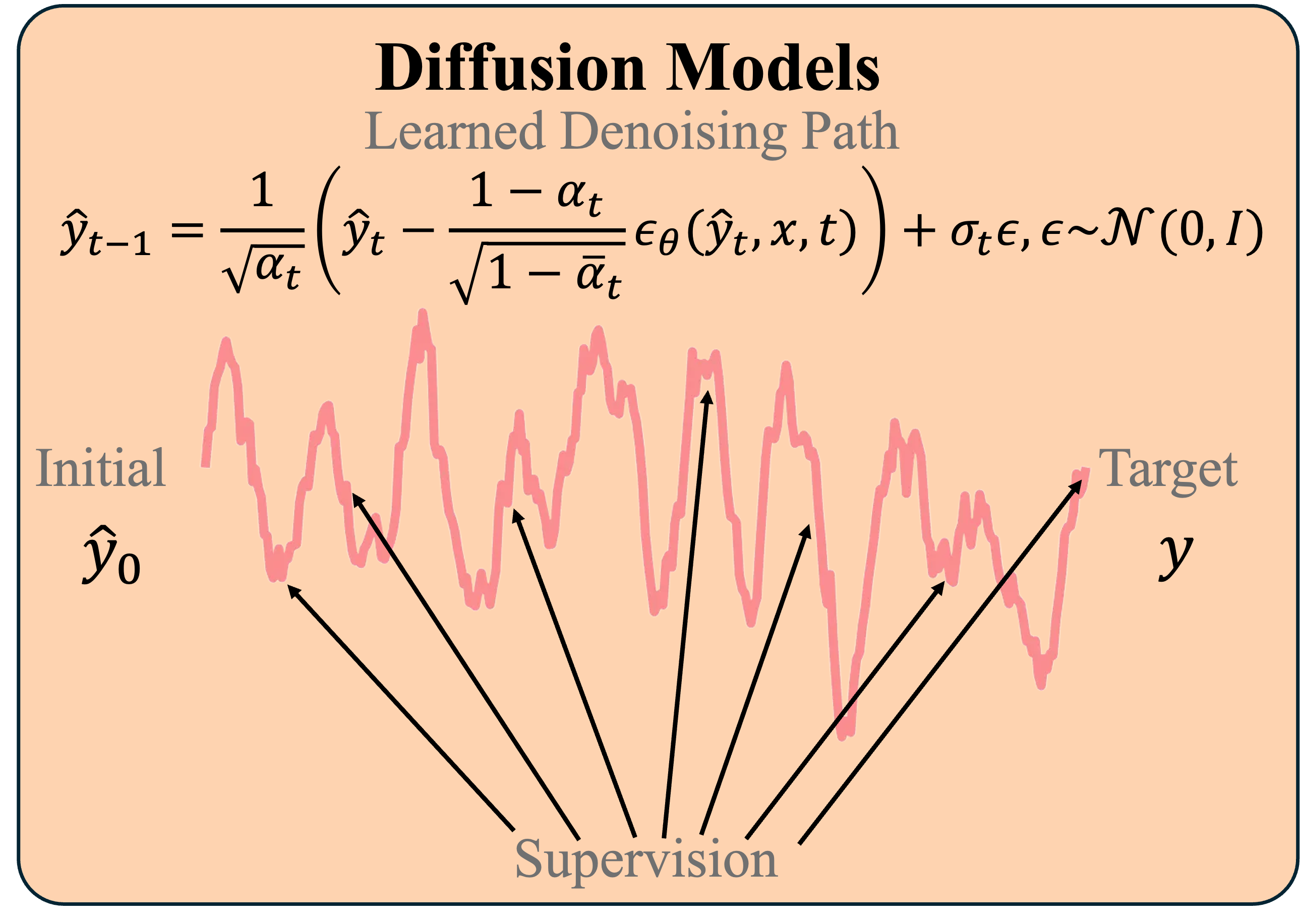}
      \caption{Diffusion Model}
    \end{subfigure}\hfill
    \begin{subfigure}[t]{0.48\columnwidth}
      \includegraphics[valign=t,width=\linewidth]{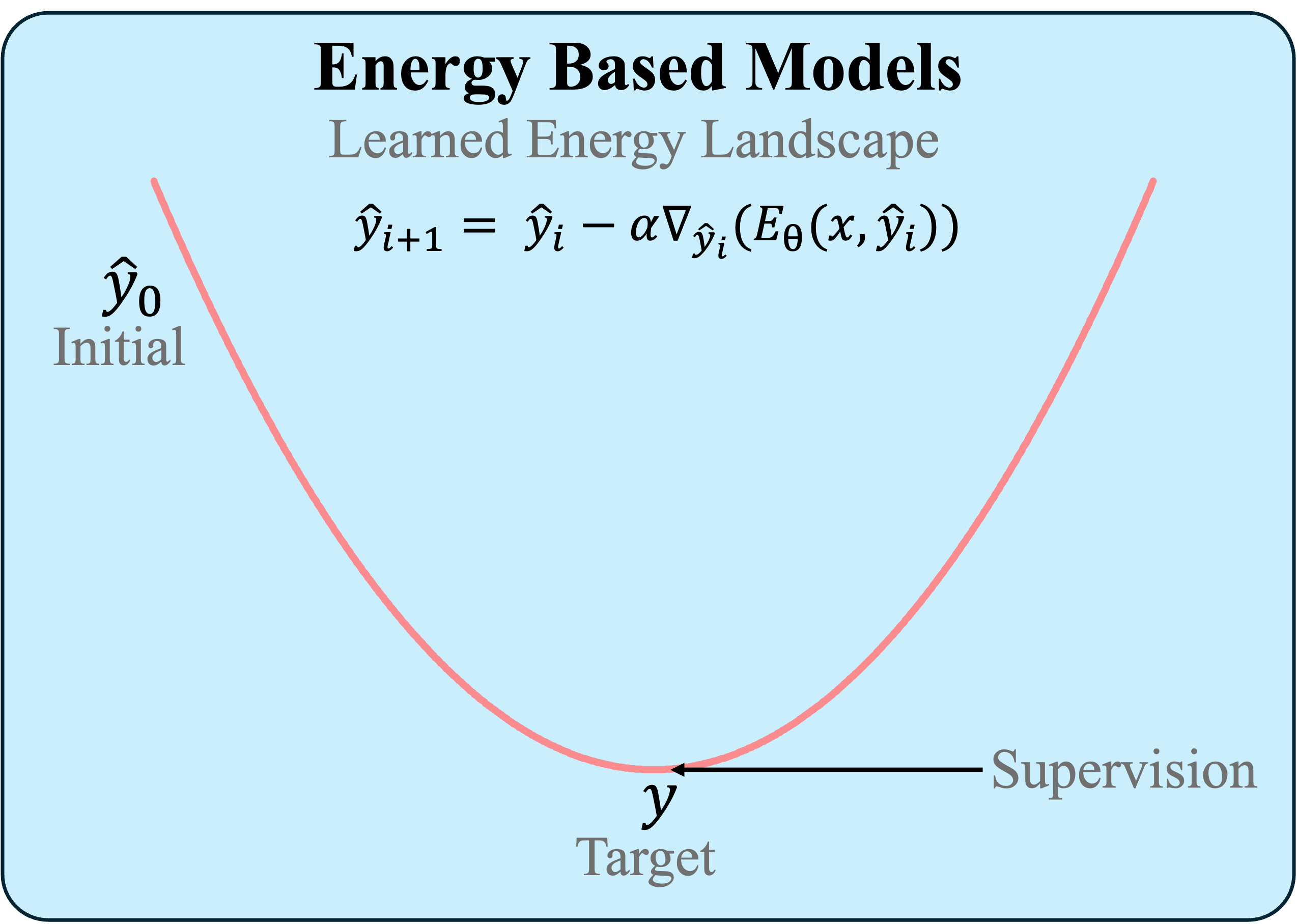}
      \caption{Energy-Based Model (EBM)}
    \end{subfigure}
    \caption{\textbf{EBM and Diffusion Comparison.} Diffusion models receive supervision at each step of the denoising process (e.g., for one thousand steps), whereas EBMs only receive supervision at the end of the optimization process. This training procedure allows EBMs to learn an entire Energy Landscape over predictions, associating a scalar energy for every prediction according to its likelihood. Learning landscapes in this manner can reduce ``error'' accumulation throughout the denoising process~\cite{du2024learning} and makes EBMs more flexible by allowing unnormalized likelihood estimation at each step of the denoising process. Additionally, diffusion models update predictions by predicting the noise at each timestep, meaning they must follow a set denoising schedule. On the other hand, EBMs update predictions by performing gradient descent with respect to the energy scalar, allowing for flexible inference where this optimization process can be performed for any number of steps. $x$ here refers to some condition (e.g., a class or text) whereas $y$ is the generated prediction.}
    \label{fig:diffusion_vs_ebm}
  \end{figure}

It's also worth noting that RNN, diffusion models, and EBMs need not be incompatible with one another. For example,~\cite{du2024learning} combines EBMs and diffusion to reason over challenging problems. This can increase stability of the learned energy landscape by adding explicit score supervision.

\subsection{In Depth EBM and Diffusion Model Comparison}
\label{sec:diffusion_vs_ebm}
Because of the similarity of diffusion models and EBMs, we present a side-by-side comparison of the training and inference approach for both in Figure~\ref{fig:diffusion_vs_ebm}, where the primary difference lie in the supervision they receive during training and the update rule for predictions. Additionally, we provide more information to compare these approaches.

Under the assumption that the energy landscape is well formed and that optimization is well behaved, EBMs offer several distinct advantages over diffusion models. In EBMs, the energy function is trained to represent a meaningful landscape where the energy value of a sample directly corresponds to its relative unnormalized likelihood. Consequently, two samples can be directly compared to determine which is more likely, in a single forward pass. On the other hand, diffusion models require running samples through the entire reverse diffusion process to get likelihoods, which often requires hundreds to thousands of steps, and rely on likelihood approximations such as ELBOs or numerical solvers for SDEs/ODEs. In practice, these result in incomparable likelihoods, as ELBOs only give likelihood lower bounds and numerical solvers result in high approximation error~\cite{ma2025inference}.

Furthermore, the learning of an energy landscape over predictions means that any approximation errors at each individual step of the Markov Chain (optimization process) do not result in cumulative error, as the minimum of the energy landscape can still be reached. This differs from diffusion models, where any approximation error at each step will result in increasing accumulated error across the entire Markov Chain~\cite{du2024learning} (demonstrated in Figure~\ref{fig:diffusion_vs_ebm}).

Lastly, EBMs, giving an unnormalized likelihood estimate at each step, are in practice much more flexible for generation than diffusion models. While diffusion models require running the entire reverse diffusion process with a specific denoising schedule to generate a sample, EBMs can be trained to directly predict the next sample in a single step, and give an unnormalized likelihood at each step indicating how likely they think this sample is. This better approximates human-like System 2 thinking, where humans naturally evaluate the strength of current predictions, and on the basis of knowing how good their predictions are, decide to dynamically allocate more or less computational resources. 

\vspace{-5pt}
\subsection{Additional Energy-Based Models Related Works}

One contribution of this work was the design of a custom architecture for EBMs called the Energy-Based Transformer ({\parch}). Roughly similar in naming is the work of the Energy Transformer~\cite{hoover2024energy}. However, despite similarity in the names of these architectures, they are very different---with the primary similarity in architectures being the usage of attention mechanisms as well as a global energy function. The existing work integrated ideas from Modern Hopfield Networks, including RNNs, whereas in our work the architecture is non-recurrent and does not use associative memories. Additionally, in this work with EBTs we focused on System 2 Thinking and scalability to high-dimensional problems, which the previous work did not experiment with.

Other loosely related approaches to EBTs involve autoregressive Energy-Based Models, including E-ARM~\cite{wang2022your}, EBR~\cite{bhattacharyya2020energy}, and Residual EBMs~\cite{bakhtin2021residual}. E-ARM involves adding an objective to the learning process to turn a traditional autoregressive model into an EBM, and as such does not achieve two of the cognitive facets discussed in Section~\ref{sec:intro}. EBR and Residual EBMs involve the training of an EBM on top of an already pretrained autoregressive language model. Both works, however, leverage a contrastive objective, which suffers from the curse of dimensionality. 

The optimization procedure used to train {\pa}s can be seen as a form of denoising score matching~\cite{vincent2011connection, wang2023energy}. Particularly, EBTs having predictions initialized at some Gaussian, and then optimizing predictions using the gradient of the energy function, can be seen as training EBMs to denoise by learning the score of the data starting from a Gaussian prior. However, we find the optimization perspective is more intuitive, and this denoising score matching perspective is more similar to the diffusion model training procedure than it is EBMs, involving multiple levels of noise rather than just one level.

%% file: arxiv/supp_sec/F_Additional_Facets.tex
\vspace{-5pt}
\section{Additional Cognitive Facets}
\label{sec:cognitive_facet_details}

In this section, we describe in more detail the different facets of cognition, in addition to introducing additional facets.

Facet~\ref{facet:dynamic-computation}, Dynamic Allocation of Computation, can be formalized as Turing completeness (assuming an infinite length external memory). Because standard feed-forward Transformers have finite length programs (i.e., they have finite depth), they are unable to be Turing complete at the granularity of each prediction being made. This also holds for common RNN variants that are only updated with new sequence information. Diffusion models and EBMs, being able to iteratively compute functions, have potentially infinite length programs, and therefore when augmented with an infinite external memory can be Turing complete.

\facet[facet:compositionality]{Compositional Reasoning and Systematicity}
Humans routinely solve novel tasks by recombining familiar primitives (e.g., verbs with new arguments or visual parts into unseen objects), a hallmark of compositional generalization well-documented in neuroscience~\cite{friederici2007mapping}. In contrast, state-of-the-art Transformers and diffusion models often fall short when evaluated on compositional generalization~\cite{kobayashi2024can, huang2023t2i}. Energy-Based Models (EBMs) seamlessly address these limitations: energies for individual factors are composable in several different manners~\cite{du2023reduce}, enabling zero-shot generation of novel combinations without retraining, where gradient-based sampling provides an intrinsic mechanism to verify and iteratively correct compositions~\cite{du2023reduce}. Thus, EBMs offer a promising path toward human-like systematicity that remains elusive for existing paradigms and architectures.

%% file: arxiv/supp_sec/G_Counterarguments.tex
\section{Counterarguments}
\label{sec:counterarguments}

\subsection{System 2 Thinking}
In this paper, strong claims were made regarding the capabilities of current models and their ability to perform System 2 Thinking. However, there are common counterarguments to our claims, which we address here.

\subsubsection{System 2 Thinking and Inference-Time Compute Term Usage}
Whether the computational effort spent at inference time fully captures what psychologists term System 2 Thinking is still actively debated. In Section~\ref{sec:thinking_formalization} we outline three reasons for preferring the broader terminology of System 2 Thinking: (i) it naturally extends to settings such as continual learning where terms such as ``inference-time compute'' becomes ambiguous, (ii) it connects our discussion to a substantial body of cognitive-science work, and (iii) it offers a conceptually straightforward entry point for readers beyond the machine-learning community.

We believe that Human System 2 Thinking encompasses a far greater depth and complexity than the specific approaches explored in this paper, such as ``Thinking Longer'' and ``Self-Verification.'' We wish to emphasize that our work does not claim current models replicate the full spectrum of human System 2 Thinking. Rather, we view the methods presented here as foundational steps toward that more ambitious long-term goal.

We propose that ``System 2 Thinking'' offers a useful umbrella term that can effectively encompass and generalize other existing terminologies, including ``inference-time compute,'' ``test-time compute,'' or ``reasoning." A parallel can be drawn with the term ``learning'' in our field. ``Learning'' itself has evolved to describe a wide array of processes, some of which, such as k-Nearest Neighbors (KNNs) or the specific mechanisms of weight matrix updates in Artificial Neural Networks (ANNs), represent distinct facets rather than the entirety of human-like learning, meaning these approaches may not encompass the complexity of true human-like learning. However, despite this breadth and these simplicities, ``learning'' has become a cornerstone term, upon which our entire field of machine learning is named upon.

In a similar vein, while the ``thinking'' exhibited by the models discussed in this paper may not yet capture the full nuance and intricacy of human cognition, we believe the term ``System 2 Thinking'' can still serve a valuable role. It offers a generalizing framework for existing vocabulary and can contribute to making complex concepts within the field more accessible and understandable to newcomers or experts from other fields. Our intention is to contribute to a constructive and unifying dialogue of intelligent systems.

\subsubsection{Is Chain-of-Thought (CoT) Sufficient for System 2 Thinking?}
CoT is commonly believed to be sufficient for advanced reasoning to emerge in LLMs. However, we argue that there are several flaws with CoT preventing advanced thinking capabilities.
First, Chain-of-Thought (CoT) involves reasoning over a discrete state space, which limits the granularity of ``thoughts.'' Second, CoT is not an intrinsic architectural capability but an external procedure applied to token sequences. Ideally, such reasoning should be embedded within the model and learned during training (unsupervised training, particularly). Third, each token is produced with a fixed computational budget, restricting the depth of reasoning per step. In contrast, humans allocate variable effort across steps when reasoning ``step by step''. Similarly, models should be able to spend a variable amount of computation per token, as enabled by {\pa}s. This aligns with the intuition behind the saying: ``a chain is only as strong as its weakest link''---\textit{each step in a chain of thought should receive sufficient computation to avoid failure points that result in bad reasoning}.



%% file: arxiv/supp_sec/H_EBM_Intro.tex
\section{Energy-Based Models (EBMs) Introduction}
\label{sec:ebm_intro}

\subsection{Simplified Energy-Based Model (EBM) Introduction}

\begin{figure}[!ht]
  \centering
  \scriptsize
  \begin{subfigure}[b]{0.25\columnwidth}
    \includegraphics[height=0.15\textheight]{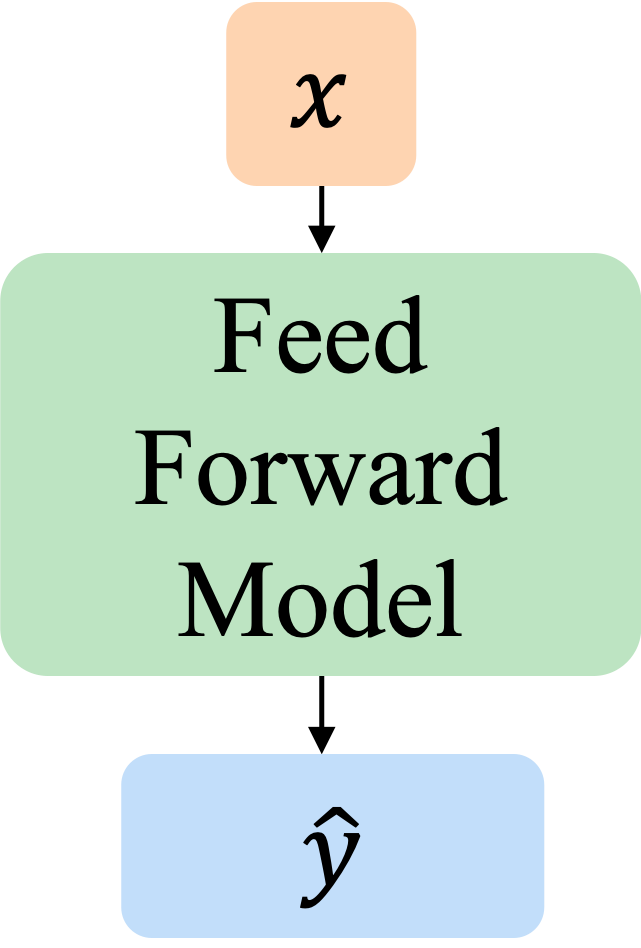}
    \caption{Feed Forward Model}
    \label{fig:general_feed_forward_model_arch}
  \end{subfigure}
  \begin{subfigure}[b]{0.25\columnwidth}
    \includegraphics[height=0.15\textheight]{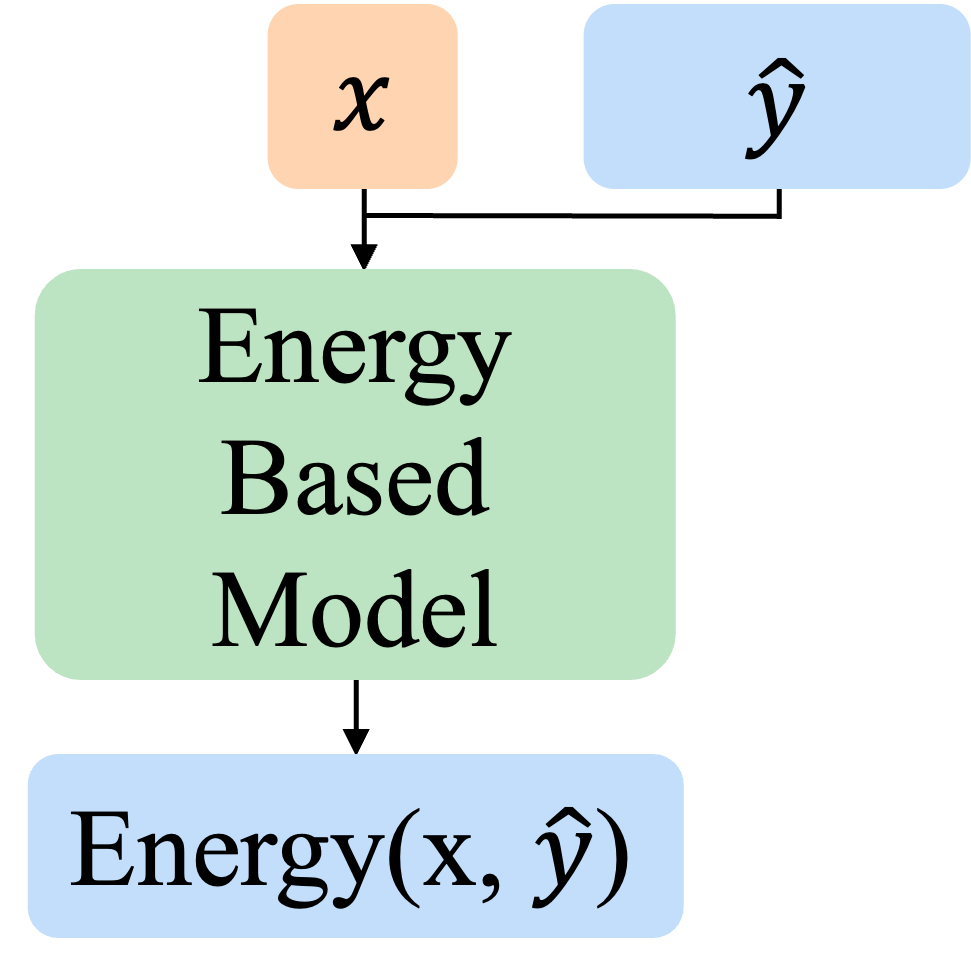}
    \caption{Energy-Based Model}
    \label{fig:general_ebm_arch}
  \end{subfigure}
  \caption{\textbf{Feed-Forward and Energy-Based Model Comparison.}  Feed-forward models (a), given an input $x$, directly try to predict $\hat{y}$. Instead of just getting $x$ as an input, EBMs receive both $x$ and $\hat{y}$ as an input and learn the \textbf{compatibility} of all possible values of $\hat{y}$ with $x$ by outputting a scalar energy value for each combination. Low energy corresponds to high probability, and high energy to low probability. In practice, $\hat{y}$ is often initialized as random.}
  \label{fig:simple_model_comparison}
\end{figure}

Feed-forward neural networks generally take the form of: given an $x$ predict $y$ (Fig.~\ref{fig:general_feed_forward_model_arch}). Energy-Based Models (EBMs) are a family of models that learn the compatibility (unnormalized probability) over all possible combinations of $x$ and $y$ (Fig.~\ref{fig:general_ebm_arch}). Intuitively, this can be seen as learning to verify the strength of $y$ as a prediction with $x$ as the input. Training models in this manner allows for representing multiple plausible versions of $y$ compatible with a given $x$. The differences between these models is visualized in Fig.~\ref{fig:simple_model_comparison}.

Formulating models in this manner ultimately brings about two primary questions: \\

\textbf{First question:} Assuming we still care about ultimately predicting $\hat{y}$ \textit{how do we use such an EBM to predict $\hat{y}$?} With feed forward models, generally we can just input $x$ and get the output of the model as $\hat{y}$, but we can't do this with EBMs? 

With EBMs, what happens is conceptually similar to diffusion models~\cite{rombach2022high}, where we (commonly) initialize $\hat{y}$ as random noise. Then, we input $x$ and $\hat{y}$ into the model, and get a single scalar energy output (our initial energy output) from the model. Now, because our entire model is differentiable, we can get the \textit{gradient from this energy scalar to $\hat{y}$} and perform gradient descent along the \textit{energy landscape} (energy landscapes are surfaces resulting from mapping all possible predictions to scalar values, visualized in Figure~\ref{fig:energy_landscape_minimization}) using this gradient (this is the key)! This process is visualized in Figures~\ref{fig:energy_landscape_minimization},~\ref{fig:model_architecture}. This gradient can be seen as the opposite of the noise (e.g., denoising) and therefore EBMs have strong relations with Diffusion models predicting the noise. EBMs can be seen as a generalization of diffusion models, where diffusion models are predicting the gradient of the energy function/scalar (more on this in Section~\ref{sec:additional_related_work}).

\textbf{Second question:} \textit{How do we train an EBM?} Generally, models are trained over a dataset of $x$ and $y$ pairs, but now there are several different possible $y$ values that can be associated with any given $x$ value---so how does that work?

It turns out that all the training techniques for EBMs boil down to two main categories: contrastive and regularizing approaches~\cite{dawid2024introduction}.

Contrastive approaches are more common for EBMs and are easier to rationalize about due to their similarity to GAN discriminators. The idea behind contrastive approaches is to push down on the energy of positive samples (i.e., the true data), and to push up on the energy of negative samples. While these positive samples are easy to rationalize about, as they are just the true data, the difficulty of contrastive EBMs is finding negative samples. Several approach exist, such as GANs, which use a generator to amortize negative sample generation, or running MCMC (similar to optimization) for some time. However, as discussed in Section~\ref{sec:learning_approach}, such approaches don't scale well due to the curse of dimensionality. 

Therefore, to achieve a scalable EBM approach, we train EBMs through an optimization procedure (which has strong resemblance to Langevin Dynamics). That is, EBMs are trained to, starting from an initial prediction, optimize predictions to the ground truth solution (shown in Figures~\ref{fig:energy_landscape_minimization},~\ref{fig:model_architecture}). This pushes the energy landscape to be locally convex surrounding the ground truth solution, thereby regularizing the energy landscape to have low energy only on the true data. As mentioned in~\cite{wang2023energy}, this can be seen as being similar to denoising score matching~\cite{vincent2011connection}.

Commonly, when people learn about EBMs and the iterative denoising/optimization procedure performed during training, they think of diffusion models, so we include a more in depth comparison between the two in Section~\ref{sec:diffusion_vs_ebm}.




\subsection{Energy-Based Model Types}
Because EBMs are a broad modeling framework, and can generalize many existing approaches, we aim to provide precise language to distinguish EBM types. We broadly classify the EBMs described throughout this paper as \textbf{explicit EBMs}, meaning they explicitly define an energy function over inputs as the entire function being learned. In other words, explicit EBMs directly map all variables (inputs) to a single scalar energy as the output of the neural network. We define these in contrast to \textbf{implicit EBMs}, or EBMs where the energy function is not the learned model but rather some implicit definition of the learned model. For example, with diffusion models, the energy function is implicitly defined by the learned score network ($s_\theta(x,t)$) as the following: \\
\begin{align*}
\nabla_x E(x,t) &= -\,s_\theta(x,t)\,\\
E(x,t) &= -\int^x s_\theta(u,t)\,\mathrm{d}u \;+\; C(t)\,.
\end{align*}
\\
Other notable examples of implicit EBMs include Hopfield Networks~\cite{hopfield1982neural}, RNNs~\cite{geiping2025scaling}, and Boltzmann machines.

\subsection{Energy-Based Model Frequently Asked Questions (FAQ)}
\begin{itemize}
  \item \textbf{What is energy/compatibility, what does it represent, and how is it learned? What energy corresponds to what probability?}\\
    Energy is just a learned compatibility score between $x$ and $y$ (lower means more likely). The EBMs described in this paper learn it implicitly as described in Section~\ref{sec:learning_approach} such that the true data (good pairs) have low energy and bad pairs (non-data) have higher energy. Probabilities follow: \\
    \begin{align*}
        p_\theta(x) = \frac{e^{-E_{\theta}(x, y)}}{Z(\theta)} \\
        p_\theta(x, y)\propto e^{-E_\theta(x,y)}
    \end{align*}
     
    so the energy $E$ is essentially the (unnormalized) negative log-likelihood up to an additive constant. The term compatibility is just a term used for intuition.

  \item \textbf{Does training EBMs require a full Hessian calculation?}\\
    No---the approach described in the paper only requires Hessian-vector products. That makes training only about a constant $1.66\times$ as expensive as a vanilla feed-forward model given everything else remains constant and you use a single step.

  \item \textbf{Why is low energy good (and high energy bad)? Why not just use probability?}\\
    Low energy is good because of the negative exponential. The reason we don't use probabilities is avoiding normalized probabilities makes the problem much more tractable in real-world high-dimensional continuous state spaces by removing the focus on explicit normalization via regularizing the partition function. EBMs come from a long line of work in statistical physics.

  \item \textbf{Is it okay that energies are unnormalized probabilities? }\\
    Yes, for most real-world applications, you only ever need sample \textbf{relative likelihood} comparison; it's significantly less common to need the exact likelihood of samples. An example of this is reward models, which can be seen as EBMs (just multiplying the reward by $-1$), where all that really matters is the relative reward for choosing which sample to use or which behavior to perform. 
    
  \item \textbf{Is it fine to not do Maximum Likelihood Training?} \\
    Contrary to what your intuition may say, the answer is \textbf{yes}! For most real-world distributions, data lies completely concentrated on a very thin manifold with no defined distribution outside of this manifold. Thus, directly doing Maximum Likelihood Estimation (MLE) training would push EBMs to have low energy on the true data manifold and then infinite energy off that manifold (as the probability of such samples is $0$). We don't want this as it would make the score (gradient of the energy function) undefined and the energy landscape untraversable---so not doing MLE makes the problem tractable. 
\end{itemize}

%% file: arxiv/supp_sec/I_How_to_Train.tex
\section{Energy-Based Transformers (EBTs) Tutorial}
\label{sec:tutorial}

\subsection{Improving Stability and Scalability}
Energy-Based Models are notorious for instability during training~\cite{du2019implicit, du2020improved, li2023learning, arbel2020generalized}. Therefore, we experiment with several different hyperparameters to increase the stability and scalability of {\pa}s and EBMs in general.

\subsubsection{Optimization Step Size and Stability}
We found that the step size for gradient descent updates of predictions ($\alpha$) was one of the primary factors affecting the stability of {\pa}s. Thus, for S1 models, we make the step size a learnable parameter (this is not the case for S2 models). We calculate its learning rate by multiplying the model's learning rate by the step size learning rate multiplier. We found that the values for the step size have a large effect on the magnitude of gradients generated for the optimization of predictions. This is because the step size is directly multiplied by the prediction gradients. Particularly, a smaller step size results in larger generated gradients, whereas a larger step size results in smaller gradients. Therefore, the step size needs to be tuned per modality, as the update required for predictions depends on data. It's also worth noting that we do not weight decay the step size in any of the models.

We also found that a relatively high optimization step size was necessary ($30,000$ for video and between $5$ and $500$ for text). Without a high optimization step size, gradient magnitudes continued to increase throughout training, resulting in unstable training dynamics.

\subsubsection{Architecture Stability}
For autoregressive models, we found that simply prepending a learnable ``step'' embedding to sequences significantly improved scalability and stability, especially for S1 models. This step embedding mapped a discrete step index (i.e., step 0, 1, etc.) of the current optimization step to an embedding the same dimension as the model's embedding. We believe this helped improve stability by enabling the accumulation of attention mass, as well as enabling less steep energy landscapes conditioned on the optimization step. 

Additionally, we experimented with adaptive layer normalization from the DiT architecture, but we found that the timestep embedding worked better. We also experimented with several different normalization and initialization approaches, where we found that the standard Llama2~\cite{touvron2023llama} architecture and initialization worked best. This involves using RMSNorm, Xavier init~\cite{glorot2010understanding}, SwiGLU MLP~\cite{shazeer2020glu}, and RoPE~\cite{su2024roformer}.

\subsubsection{System 2 Thinking Hyperparameter Stability}
For S2 models, we found certain hyperparameters to be essential for the stability and scalability of models. First, we found that using a lower number of optimization steps resulted in more stability, as using more optimization steps necessitates longer gradient chains. Additionally, we found that the strategy used for the randomization of $\alpha$ to be very important for stability. Particularly, we found that randomizing $\alpha$ with a single value for an entire batch resulted in issues with training convergence. We believe this is because using the same value for every element in the batch resulted in high-variance gradients. Thus, when randomizing $\alpha$ differently for every batch and sequence element, we found training convergence to be more stable. It's possible that randomizing the number of optimization steps would yield similar results in reducing gradient variance, however, we did not experiment with such a configuration.

\subsubsection{Clamping Optimization Gradients for Stability}
We found that clamping gradients of the energy function with respect to predictions (or prediction update gradients) could help improve training stability at the cost of some slight reductions in convergence speed. We do not conduct any experiments in the paper with clamped prediction gradients as we found that the other hyperparameters were sufficient for stable training, however, it's a potentially useful trick worth discussing.

\subsubsection{Normalizing Data for Stability}
We found that normalizing/standardizing input data was crucial for the stability of EBTs. An example of this was in our NLP experiments, where we ran experiments with and without normalizing probability distributions. The experiments with unnormalized distributions often had extreme activations as well as large loss spikes, whereas the experiments with normalized distributions (by applying softmax) were stable.

\subsection{How to Train Your EBT}
\label{sec:how_to_learn_ebt}
The hyperparameters used for training {\pa}s are \textit{extremely} important and can often be highly sensitive towards performance, so here we offer a guide toward hyperparameter tuning. First, we recommend starting off with training S1-EBTs, which are the easiest and most stable variant of EBTs not designed for System 2 Thinking. Then, once S1-EBTs can be trained succesfully and scaled, we recommend changing the hyperparameters gradually towards the hyperparameters used for S2 models (we say gradually here as occasionally these parameters can cause instability and require additional tuning). 

When training S1 EBTs, we recommend tuning hyperparameters in the following order: \\
\begin{itemize}
    \item First, tune standard hyperparameters such as Learning Rate (LR), batch size, etc. Having a high batch size helps with stability by making gradients less noisy (because you initialize predictions from random noise this makes gradients noisier). \\

    \item Second, start tuning S1-specific hyperparameters---primarily alpha and its LR multiplier (we recommend keeping its LR multiplier around $3x$ the value of alpha) and then tuning the number of optimization steps.

    \item Third, potentially tune whether the step size is learnable and try other EBT architectures (inspired by DiT~\cite{peebles2023scalable} we tried a time embedding as well as adaptive layer normalization).
\end{itemize}

Once you have tuned these, the model should be stable and fine for most use cases. At which point, if you are desiring System 2 capabilities, you can proceed to the S2 models~\ref{sec:how_to_think_ebt}. Some potential metrics to monitor and look out for include the gradient magnitudes (if these increase too much or spike a lot that's a bad sign) and the gap between the initial and final energy after optimization (if this is too high or low it could be a sign your model's alpha value needs to be adjusted).

Following~\cite{wang2023energy}, we give pseudocode for training {\pa}s in natural language for language modeling (Listing~\ref{lst:ebt_nlp_training}) as well as in computer vision for autoregressive video modeling (Listing~\ref{lst:ebt_cv_training}). The pseudocode is primarily for S2 models without any energy landscape regularization techniques. The first primary design decision in the presented pseudocode is whether or not to detach predictions in between steps. Not detaching predictions in between steps allows for more ``Thinking Time'' before making predictions, but makes the gradient computation graph longer and therefore increases the likelihood of stability issues with gradients. Similarly, calculating the loss at every step versus solely the last step enables model to ``think for longer'' before needing to make accurate predictions, and therefore affects the convexity and smoothness of the energy landscape. For S2 models, we found that \textbf{not} detaching between steps was best, and similarly that calculating the loss only at the last step was best. For S1 models, we found the opposite to be most stable. Generally, if one is calculating the loss only at the last step, then one should not detach between steps as it's best if the gradient propagates to previous steps in this case. For more details on these techniques, we refer the reader to the source code as well as Section~\ref{sec:how_to_think_ebt}.

\begin{lstlisting}[language=Python, caption={\textbf{Autoregressive Language Model Training Pseudocode in PyTorch}}, label={lst:ebt_nlp_training}]
# make sure to enable gradient tracking
with torch.set_grad_enabled(True):
    loss_fn = nn.CrossEntropyLoss(weight=None, ignore_index=tokenizer_pad_token_id)

    context_embeddings = self.embeddings(input_ids[:, :-1]) # B, S, D
    next_tokens = input_ids[:, 1:]
    next_embeddings = self.embeddings(next_tokens) # B, S, V; are just used for shaping next tensor
    predicted_distributions = torch.randn_like(next_embeddings) # B, S, V; initialize predictions as random

    for _ in range(num_steps):
        # Can optionally detach predicted distributions so that no gradient flows through past steps
        # predicted_distributions = predicted_distributions.detach()

        predicted_embeddings = self.vocab_to_embed(softmax(predicted_distributions)) # B, S, D; need to proj. to embed space for transformer to work in, use linear layer, weighted sum, etc

        all_embeddings = torch.cat((context_embeddings, predicted_embeddings), dim=1) # B, 2S, D
        predicted_energies = self.transformer(all_embeddings) # B, S, 1; this returns only energies for the predicted_embeddings
        
        # Compute the gradient of predicted energies w.r.t. predicted distributions
        predicted_distributions_grad = torch.autograd.grad(
            predicted_energies.sum(), 
            predicted_distributions, 
            create_graph=True
        )[0] # B, S, V
        
        # Perform gradient descent w.r.t. the energy function where self.alpha is the optimization step size
        predicted_distributions = predicted_distributions - self.alpha * predicted_distributions_grad
        
    # Calculate cce loss based on predicted and ground truth distributions, optionally at each optimization step or only at the end
    cce_loss = loss_fn(predicted_distributions, next_tokens)
\end{lstlisting}

\begin{lstlisting}[language=Python, caption={\textbf{Autoregressive Video Model Training Pseudocode in PyTorch}}, label={lst:ebt_cv_training}]
# make sure to enable gradient tracking
with torch.set_grad_enabled(True):
    loss_fn = torch.nn.SmoothL1Loss(beta=1.0) # use whichever loss function desired

    context_embeddings = embeddings[:, :-1] # B, S, D
    next_embeddings = embeddings[:, 1:] # B, S, D
    predicted_embeddings = torch.randn_like(next_embeddings) # B, S, D; initialize predictions as random

    for _ in range(num_steps):
        # Can optionally detach embeddings so that no gradient flows through past steps
        # predicted_embeddings = predicted_embeddings.detach()

        all_embeddings = torch.cat((context_embeddings, predicted_embeddings), dim=1) # B, 2S, D
        predicted_energies = self.transformer(all_embeddings) # this returns only energies for the predicted_embeddings # B, S, 1
        
        # Compute the gradient of predicted energies w.r.t. predicted embeddings
        predicted_embeddings_grad = torch.autograd.grad(
            predicted_energies.sum(), 
            predicted_embeddings, 
            create_graph=True
        )[0] # B, S, D
        
        # Perform gradient descent w.r.t. the energy function where self.alpha is the optimization step size
        predicted_embeddings = predicted_embeddings - self.alpha * predicted_embeddings_grad
        
    # Calculate reconstruction loss based on predicted and ground truth embeddings, optionally at each optimization step or only at the end
    reconstruction_loss = loss_fn(predicted_embeddings, next_embeddings)
\end{lstlisting}

\subsection{How to Think Using Your EBT}
\label{sec:how_to_think_ebt}

Once an S1 EBT has been trained, we recommend tuning the System 2 hyperparameters in the following manner: \\
\begin{itemize}
    \item Remove detaching tensors between optimization steps and add loss truncation so the loss is only calculated at the final step.
    \item Then, tune alpha, as it's by far the most important EBT-specific hyperparameter. But, do not make it learnable.
    \item Next, tune the number of optimization steps, including potentially a minimum and maximum number when randomizing the number of steps.
    \item Then add a replay buffer, Langevin Dynamics, and eventually a randomized alpha (step size). Tune all of these in tandem while tweaking the earlier parameters (particularly alpha).
\end{itemize}

It's possible that randomizing the number of steps for each sample within a batch would work better (similar to randomizing the alpha value within a batch). Additionally, it's worth mentioning that all optimization steps are performed along the same energy landscape (same time embedding condition), unlike with S1-EBTs using multiple time steps.

%% file: arxiv.bbl
\begin{thebibliography}{161}
\providecommand{\natexlab}[1]{#1}
\providecommand{\url}[1]{\texttt{#1}}
\expandafter\ifx\csname urlstyle\endcsname\relax
  \providecommand{\doi}[1]{doi: #1}\else
  \providecommand{\doi}{doi: \begingroup \urlstyle{rm}\Url}\fi

\bibitem[Kahneman(2011)]{kahneman2011thinking}
Daniel Kahneman.
\newblock \emph{Thinking, fast and slow}.
\newblock macmillan, 2011.

\bibitem[Evans(2011)]{evans2011dual}
Jonathan St~BT Evans.
\newblock Dual-process theories of reasoning: Contemporary issues and developmental applications.
\newblock \emph{Developmental review}, 31\penalty0 (2-3):\penalty0 86--102, 2011.

\bibitem[Kahneman et~al.(2002)Kahneman, Frederick, et~al.]{kahneman2002representativeness}
Daniel Kahneman, Shane Frederick, et~al.
\newblock Representativeness revisited: Attribute substitution in intuitive judgment.
\newblock \emph{Heuristics and biases: The psychology of intuitive judgment}, 49\penalty0 (49-81):\penalty0 74, 2002.

\bibitem[Frankish(2010)]{frankish2010dual}
Keith Frankish.
\newblock Dual-process and dual-system theories of reasoning.
\newblock \emph{Philosophy Compass}, 5\penalty0 (10):\penalty0 914--926, 2010.

\bibitem[Neys(2006)]{neys2006dual}
Wim~De Neys.
\newblock Dual processing in reasoning: Two systems but one reasoner.
\newblock \emph{Psychological science}, 17\penalty0 (5):\penalty0 428--433, 2006.

\bibitem[Goel et~al.(2000)Goel, Buchel, Frith, and Dolan]{goel2000dissociation}
Vinod Goel, Christian Buchel, Chris Frith, and Raymond~J Dolan.
\newblock Dissociation of mechanisms underlying syllogistic reasoning.
\newblock \emph{Neuroimage}, 12\penalty0 (5):\penalty0 504--514, 2000.

\bibitem[Li et~al.(2025{\natexlab{a}})Li, Zhang, Zhang, Zhang, Liu, Yao, Xu, Zheng, Wang, Chen, et~al.]{li2025system}
Zhong-Zhi Li, Duzhen Zhang, Ming-Liang Zhang, Jiaxin Zhang, Zengyan Liu, Yuxuan Yao, Haotian Xu, Junhao Zheng, Pei-Jie Wang, Xiuyi Chen, et~al.
\newblock From system 1 to system 2: A survey of reasoning large language models.
\newblock \emph{arXiv preprint arXiv:2502.17419}, 2025{\natexlab{a}}.

\bibitem[Mirzadeh et~al.(2024)Mirzadeh, Alizadeh, Shahrokhi, Tuzel, Bengio, and Farajtabar]{mirzadeh2024gsm}
Iman Mirzadeh, Keivan Alizadeh, Hooman Shahrokhi, Oncel Tuzel, Samy Bengio, and Mehrdad Farajtabar.
\newblock Gsm-symbolic: Understanding the limitations of mathematical reasoning in large language models.
\newblock \emph{arXiv preprint arXiv:2410.05229}, 2024.

\bibitem[Yan et~al.(2025)Yan, Lu, Xu, and Lan]{yan2025phd}
Yang Yan, Yu~Lu, Renjun Xu, and Zhenzhong Lan.
\newblock Do phd-level llms truly grasp elementary addition? probing rule learning vs. memorization in large language models.
\newblock \emph{arXiv preprint arXiv:2504.05262}, 2025.

\bibitem[Qian et~al.(2025)Qian, Han, Luo, He, Chen, Zhang, Du, Yao, Yang, Zhang, Li, and Ji]{EscapeBench2025}
Cheng Qian, Peixuan Han, Qinyu Luo, Bingxiang He, Xiusi Chen, Yuji Zhang, Hongyi Du, Jiarui Yao, Xiaocheng Yang, Denghui Zhang, Yunzhu Li, and Heng Ji.
\newblock Escapebench: Pushing language models to think outside the box.
\newblock In \emph{arxiv}, 2025.

\bibitem[Jaech et~al.(2024)Jaech, Kalai, Lerer, Richardson, El-Kishky, Low, Helyar, Madry, Beutel, Carney, et~al.]{jaech2024openai}
Aaron Jaech, Adam Kalai, Adam Lerer, Adam Richardson, Ahmed El-Kishky, Aiden Low, Alec Helyar, Aleksander Madry, Alex Beutel, Alex Carney, et~al.
\newblock Openai o1 system card.
\newblock \emph{arXiv preprint arXiv:2412.16720}, 2024.

\bibitem[Guo et~al.(2025)Guo, Yang, Zhang, Song, Zhang, Xu, Zhu, Ma, Wang, Bi, et~al.]{guo2025deepseek}
Daya Guo, Dejian Yang, Haowei Zhang, Junxiao Song, Ruoyu Zhang, Runxin Xu, Qihao Zhu, Shirong Ma, Peiyi Wang, Xiao Bi, et~al.
\newblock Deepseek-r1: Incentivizing reasoning capability in llms via reinforcement learning.
\newblock \emph{arXiv preprint arXiv:2501.12948}, 2025.

\bibitem[{xAI}(2025)]{xai2025grok3}
{xAI}.
\newblock {Grok 3 Beta — The Age of Reasoning Agents}, 2025.
\newblock URL \url{https://x.ai/blog/grok-3}.
\newblock Accessed: 2025-02-21.

\bibitem[Anthropic(2025)]{anthropic2025claude37sonnet}
Anthropic.
\newblock Claude 3.7 sonnet and claude code, 2025.
\newblock URL \url{https://www.anthropic.com/news/claude-3-7-sonnet}.
\newblock Accessed: 2025-02-21.

\bibitem[OpenAI(2024)]{openai2024learning}
OpenAI.
\newblock Learning to reason with llms, 2024.
\newblock URL \url{https://openai.com/index/learning-to-reason-with-llms/}.
\newblock Accessed: 2025-02-21.

\bibitem[Su et~al.(2025)Su, Yu, Song, Li, Mi, Tu, Zhang, and Yu]{su2025expanding}
Yi~Su, Dian Yu, Linfeng Song, Juntao Li, Haitao Mi, Zhaopeng Tu, Min Zhang, and Dong Yu.
\newblock Expanding rl with verifiable rewards across diverse domains.
\newblock \emph{arXiv preprint arXiv:2503.23829}, 2025.

\bibitem[Shojaee*† et~al.(2025)Shojaee*†, Mirzadeh*, Alizadeh, Horton, Bengio, and Farajtabar]{illusionofthinking}
Parshin Shojaee*†, Iman Mirzadeh*, Keivan Alizadeh, Maxwell Horton, Samy Bengio, and Mehrdad Farajtabar.
\newblock The illusion of thinking: Understanding the strengths and limitations of reasoning models via the lens of problem complexity, 2025.
\newblock URL \url{https://ml-site.cdn-apple.com/papers/the-illusion-of-thinking.pdf}.

\bibitem[Yue et~al.(2025)Yue, Chen, Lu, Zhao, Wang, Song, and Huang]{yue2025does}
Yang Yue, Zhiqi Chen, Rui Lu, Andrew Zhao, Zhaokai Wang, Shiji Song, and Gao Huang.
\newblock Does reinforcement learning really incentivize reasoning capacity in llms beyond the base model?
\newblock \emph{arXiv preprint arXiv:2504.13837}, 2025.

\bibitem[Ma et~al.(2025)Ma, Tong, Jia, Hu, Su, Zhang, Yang, Li, Jaakkola, Jia, et~al.]{ma2025inference}
Nanye Ma, Shangyuan Tong, Haolin Jia, Hexiang Hu, Yu-Chuan Su, Mingda Zhang, Xuan Yang, Yandong Li, Tommi Jaakkola, Xuhui Jia, et~al.
\newblock Inference-time scaling for diffusion models beyond scaling denoising steps.
\newblock \emph{arXiv preprint arXiv:2501.09732}, 2025.

\bibitem[Liu et~al.(2025)Liu, Wang, Cai, Zhang, Zhan, and Duan]{liu2025video}
Fangfu Liu, Hanyang Wang, Yimo Cai, Kaiyan Zhang, Xiaohang Zhan, and Yueqi Duan.
\newblock Video-t1: Test-time scaling for video generation.
\newblock \emph{arXiv preprint arXiv:2503.18942}, 2025.

\bibitem[Singhal et~al.(2025)Singhal, Horvitz, Teehan, Ren, Yu, McKeown, and Ranganath]{singhal2025general}
Raghav Singhal, Zachary Horvitz, Ryan Teehan, Mengye Ren, Zhou Yu, Kathleen McKeown, and Rajesh Ranganath.
\newblock A general framework for inference-time scaling and steering of diffusion models.
\newblock \emph{arXiv preprint arXiv:2501.06848}, 2025.

\bibitem[Gu and Dao(2023)]{gu2023mamba}
Albert Gu and Tri Dao.
\newblock Mamba: Linear-time sequence modeling with selective state spaces.
\newblock \emph{arXiv preprint arXiv:2312.00752}, 2023.

\bibitem[Peng et~al.(2023)Peng, Alcaide, Anthony, Albalak, Arcadinho, Cao, Cheng, Chung, Grella, GV, et~al.]{peng2023rwkv}
Bo~Peng, Eric Alcaide, Quentin Anthony, Alon Albalak, Samuel Arcadinho, Huanqi Cao, Xin Cheng, Michael Chung, Matteo Grella, Kranthi~Kiran GV, et~al.
\newblock Rwkv: Reinventing rnns for the transformer era.
\newblock \emph{arXiv preprint arXiv:2305.13048}, 2023.

\bibitem[Hochreiter and Schmidhuber(1997)]{hochreiter1997long}
Sepp Hochreiter and J{\"u}rgen Schmidhuber.
\newblock Long short-term memory.
\newblock \emph{Neural computation}, 9\penalty0 (8):\penalty0 1735--1780, 1997.

\bibitem[Geiping et~al.(2025)Geiping, McLeish, Jain, Kirchenbauer, Singh, Bartoldson, Kailkhura, Bhatele, and Goldstein]{geiping2025scaling}
Jonas Geiping, Sean McLeish, Neel Jain, John Kirchenbauer, Siddharth Singh, Brian~R Bartoldson, Bhavya Kailkhura, Abhinav Bhatele, and Tom Goldstein.
\newblock Scaling up test-time compute with latent reasoning: A recurrent depth approach.
\newblock \emph{arXiv preprint arXiv:2502.05171}, 2025.

\bibitem[Peebles and Xie(2023)]{peebles2023scalable}
William Peebles and Saining Xie.
\newblock Scalable diffusion models with transformers, 2023.

\bibitem[Li et~al.(2025{\natexlab{b}})Li, Tian, Li, Deng, and He]{li2025autoregressive}
Tianhong Li, Yonglong Tian, He~Li, Mingyang Deng, and Kaiming He.
\newblock Autoregressive image generation without vector quantization.
\newblock \emph{Advances in Neural Information Processing Systems}, 37:\penalty0 56424--56445, 2025{\natexlab{b}}.

\bibitem[Deng et~al.(2024{\natexlab{a}})Deng, Zhu, Li, Guang, and Fan]{deng2024causal}
Chaorui Deng, Deyao Zhu, Kunchang Li, Shi Guang, and Haoqi Fan.
\newblock Causal diffusion transformers for generative modeling.
\newblock \emph{arXiv preprint arXiv:2412.12095}, 2024{\natexlab{a}}.

\bibitem[Hao et~al.(2024)Hao, Sukhbaatar, Su, Li, Hu, Weston, and Tian]{hao2024training}
Shibo Hao, Sainbayar Sukhbaatar, DiJia Su, Xian Li, Zhiting Hu, Jason Weston, and Yuandong Tian.
\newblock Training large language models to reason in a continuous latent space.
\newblock \emph{arXiv preprint arXiv:2412.06769}, 2024.

\bibitem[Goyal et~al.(2023)Goyal, Ji, Rawat, Menon, Kumar, and Nagarajan]{goyal2023think}
Sachin Goyal, Ziwei Ji, Ankit~Singh Rawat, Aditya~Krishna Menon, Sanjiv Kumar, and Vaishnavh Nagarajan.
\newblock Think before you speak: Training language models with pause tokens.
\newblock \emph{arXiv preprint arXiv:2310.02226}, 2023.

\bibitem[Ditterich(2006)]{ditterich2006evidence}
Jochen Ditterich.
\newblock Evidence for time-variant decision making.
\newblock \emph{European Journal of Neuroscience}, 24\penalty0 (12):\penalty0 3628--3641, 2006.

\bibitem[Rougier et~al.(2005)Rougier, Noelle, Braver, Cohen, and O'Reilly]{rougier2005prefrontal}
Nicolas~P Rougier, David~C Noelle, Todd~S Braver, Jonathan~D Cohen, and Randall~C O'Reilly.
\newblock Prefrontal cortex and flexible cognitive control: Rules without symbols.
\newblock \emph{Proceedings of the National Academy of Sciences}, 102\penalty0 (20):\penalty0 7338--7343, 2005.

\bibitem[Tomani et~al.(2024)Tomani, Chaudhuri, Evtimov, Cremers, and Ibrahim]{tomani2024uncertainty}
Christian Tomani, Kamalika Chaudhuri, Ivan Evtimov, Daniel Cremers, and Mark Ibrahim.
\newblock Uncertainty-based abstention in llms improves safety and reduces hallucinations.
\newblock \emph{arXiv preprint arXiv:2404.10960}, 2024.

\bibitem[Van Den~Oord et~al.(2017)Van Den~Oord, Vinyals, et~al.]{van2017neural}
Aaron Van Den~Oord, Oriol Vinyals, et~al.
\newblock Neural discrete representation learning.
\newblock \emph{Advances in neural information processing systems}, 30, 2017.

\bibitem[Kingma et~al.(2013)Kingma, Welling, et~al.]{kingma2013auto}
Diederik~P Kingma, Max Welling, et~al.
\newblock Auto-encoding variational bayes, 2013.

\bibitem[Sankararaman et~al.(2022)Sankararaman, Wang, and Fang]{sankararaman2022bayesformer}
Karthik~Abinav Sankararaman, Sinong Wang, and Han Fang.
\newblock Bayesformer: Transformer with uncertainty estimation.
\newblock \emph{arXiv preprint arXiv:2206.00826}, 2022.

\bibitem[Heng et~al.(2024)Heng, Soh, et~al.]{heng2024out}
Alvin Heng, Harold Soh, et~al.
\newblock Out-of-distribution detection with a single unconditional diffusion model.
\newblock \emph{Advances in Neural Information Processing Systems}, 37:\penalty0 43952--43974, 2024.

\bibitem[Nalisnick et~al.(2018)Nalisnick, Matsukawa, Teh, Gorur, and Lakshminarayanan]{nalisnick2018deep}
Eric Nalisnick, Akihiro Matsukawa, Yee~Whye Teh, Dilan Gorur, and Balaji Lakshminarayanan.
\newblock Do deep generative models know what they don't know?
\newblock \emph{arXiv preprint arXiv:1810.09136}, 2018.

\bibitem[Serr{\`a} et~al.(2019)Serr{\`a}, {\'A}lvarez, G{\'o}mez, Slizovskaia, N{\'u}{\~n}ez, and Luque]{serra2019input}
Joan Serr{\`a}, David {\'A}lvarez, Vicen{\c{c}} G{\'o}mez, Olga Slizovskaia, Jos{\'e}~F N{\'u}{\~n}ez, and Jordi Luque.
\newblock Input complexity and out-of-distribution detection with likelihood-based generative models.
\newblock \emph{arXiv preprint arXiv:1909.11480}, 2019.

\bibitem[Bishop(1994)]{bishop1994mixture}
Christopher~M Bishop.
\newblock Mixture density networks.
\newblock 1994.

\bibitem[Song et~al.(2020)Song, Sohl-Dickstein, Kingma, Kumar, Ermon, and Poole]{song2020score}
Yang Song, Jascha Sohl-Dickstein, Diederik~P Kingma, Abhishek Kumar, Stefano Ermon, and Ben Poole.
\newblock Score-based generative modeling through stochastic differential equations.
\newblock \emph{arXiv preprint arXiv:2011.13456}, 2020.

\bibitem[Dawid and LeCun(2024)]{dawid2024introduction}
Anna Dawid and Yann LeCun.
\newblock Introduction to latent variable energy-based models: a path toward autonomous machine intelligence.
\newblock \emph{Journal of Statistical Mechanics: Theory and Experiment}, 2024\penalty0 (10):\penalty0 104011, 2024.

\bibitem[Peters et~al.(2017)Peters, McEwen, and Friston]{Peters2017Uncertainty}
A.~Peters, B.~McEwen, and Karl~J. Friston.
\newblock Uncertainty and stress: Why it causes diseases and how it is mastered by the brain.
\newblock \emph{Progress in Neurobiology}, 156:\penalty0 164--188, 2017.
\newblock \doi{10.1016/j.pneurobio.2017.05.004}.

\bibitem[Vilares et~al.(2012)Vilares, Howard, Fernandes, Gottfried, and Kording]{Vilares2012Differential}
I.~Vilares, J.~D. Howard, Hugo~L. Fernandes, J.~Gottfried, and Konrad~Paul Kording.
\newblock Differential representations of prior and likelihood uncertainty in the human brain.
\newblock \emph{Current Biology}, 22:\penalty0 1641--1648, 2012.
\newblock \doi{10.1016/j.cub.2012.07.010}.

\bibitem[Sarinopoulos et~al.(2010)Sarinopoulos, Grupe, Mackiewicz, Herrington, Lor, Steege, and Nitschke]{Sarinopoulos2010Uncertainty}
Issidoros~C. Sarinopoulos, D.~Grupe, Kristen~L. Mackiewicz, J.~Herrington, M.~Lor, E.~E. Steege, and J.~Nitschke.
\newblock Uncertainty during anticipation modulates neural responses to aversion in human insula and amygdala.
\newblock \emph{Cerebral cortex}, 20 4:\penalty0 929--40, 2010.
\newblock \doi{10.1093/cercor/bhp155}.

\bibitem[Loesche et~al.(2018)Loesche, Goslin, and Bugmann]{loesche2018paving}
Frank Loesche, Jeremy Goslin, and Guido Bugmann.
\newblock Paving the way to eureka—introducing “dira” as an experimental paradigm to observe the process of creative problem solving.
\newblock \emph{Frontiers in Psychology}, 9:\penalty0 1773, 2018.

\bibitem[Alkouri(2016)]{alkouri2016using}
Zaid Alkouri.
\newblock Using contents and containers to investigate problem solving strategies among toddlers.
\newblock 2016.

\bibitem[Du et~al.(2022)Du, Li, Tenenbaum, and Mordatch]{du2022learning}
Yilun Du, Shuang Li, Joshua Tenenbaum, and Igor Mordatch.
\newblock Learning iterative reasoning through energy minimization.
\newblock In \emph{International Conference on Machine Learning}, pages 5570--5582. PMLR, 2022.

\bibitem[Silver et~al.(2017)Silver, Schrittwieser, Simonyan, Antonoglou, Huang, Guez, Hubert, Baker, Lai, Bolton, et~al.]{silver2017mastering}
David Silver, Julian Schrittwieser, Karen Simonyan, Ioannis Antonoglou, Aja Huang, Arthur Guez, Thomas Hubert, Lucas Baker, Matthew Lai, Adrian Bolton, et~al.
\newblock Mastering the game of go without human knowledge.
\newblock \emph{nature}, 550\penalty0 (7676):\penalty0 354--359, 2017.

\bibitem[Lightman et~al.(2023)Lightman, Kosaraju, Burda, Edwards, Baker, Lee, Leike, Schulman, Sutskever, and Cobbe]{lightman2023let}
Hunter Lightman, Vineet Kosaraju, Yuri Burda, Harrison Edwards, Bowen Baker, Teddy Lee, Jan Leike, John Schulman, Ilya Sutskever, and Karl Cobbe.
\newblock Let's verify step by step.
\newblock In \emph{The Twelfth International Conference on Learning Representations}, 2023.

\bibitem[Du and Mordatch(2019)]{du2019implicit}
Yilun Du and Igor Mordatch.
\newblock Implicit generation and modeling with energy based models.
\newblock \emph{Advances in neural information processing systems}, 32, 2019.

\bibitem[Du et~al.(2020)Du, Li, Tenenbaum, and Mordatch]{du2020improved}
Yilun Du, Shuang Li, Joshua Tenenbaum, and Igor Mordatch.
\newblock Improved contrastive divergence training of energy based models.
\newblock \emph{arXiv preprint arXiv:2012.01316}, 2020.

\bibitem[Li et~al.(2023)Li, Chen, and Sommer]{li2023learning}
Zengyi Li, Yubei Chen, and Friedrich~T Sommer.
\newblock Learning energy-based models in high-dimensional spaces with multiscale denoising-score matching.
\newblock \emph{Entropy}, 25\penalty0 (10):\penalty0 1367, 2023.

\bibitem[Arbel et~al.(2020)Arbel, Zhou, and Gretton]{arbel2020generalized}
Michael Arbel, Liang Zhou, and Arthur Gretton.
\newblock Generalized energy based models.
\newblock \emph{arXiv preprint arXiv:2003.05033}, 2020.

\bibitem[Radford et~al.(2018)Radford, Narasimhan, Salimans, Sutskever, et~al.]{radford2018improving}
Alec Radford, Karthik Narasimhan, Tim Salimans, Ilya Sutskever, et~al.
\newblock Improving language understanding by generative pre-training.
\newblock 2018.

\bibitem[Devlin et~al.(2019)Devlin, Chang, Lee, and Toutanova]{devlin2019bert}
Jacob Devlin, Ming-Wei Chang, Kenton Lee, and Kristina Toutanova.
\newblock Bert: Pre-training of deep bidirectional transformers for language understanding, 2019.

\bibitem[Li et~al.(2018)Li, Xu, Taylor, Studer, and Goldstein]{li2018visualizing}
Hao Li, Zheng Xu, Gavin Taylor, Christoph Studer, and Tom Goldstein.
\newblock Visualizing the loss landscape of neural nets.
\newblock \emph{Advances in neural information processing systems}, 31, 2018.

\bibitem[Cook(2023)]{cook2023complexity}
Stephen~A Cook.
\newblock The complexity of theorem-proving procedures.
\newblock In \emph{Logic, automata, and computational complexity: The works of Stephen A. Cook}, pages 143--152. 2023.

\bibitem[Goldwasser et~al.(2019)Goldwasser, Micali, and Rackoff]{goldwasser2019knowledge}
Shafi Goldwasser, Silvio Micali, and Chales Rackoff.
\newblock The knowledge complexity of interactive proof-systems.
\newblock In \emph{Providing sound foundations for cryptography: On the work of shafi goldwasser and silvio micali}, pages 203--225. 2019.

\bibitem[Gödel(1956)]{godel1956letter}
Kurt Gödel.
\newblock Letter to john von neumann, 1956.
\newblock URL \url{https://ecommons.cornell.edu/server/api/core/bitstreams/46aef9c4-288b-457d-ab3e-bb6cb1a4b88e/content}.
\newblock Accessed: 2025-04-28.

\bibitem[Lavin et~al.(2024)Lavin, Liu, Mohanty, Norman, Zaarour, and Krishnamachari]{lavin2024survey}
Ryan Lavin, Xuekai Liu, Hardhik Mohanty, Logan Norman, Giovanni Zaarour, and Bhaskar Krishnamachari.
\newblock A survey on the applications of zero-knowledge proofs.
\newblock \emph{arXiv preprint arXiv:2408.00243}, 2024.

\bibitem[Rivest et~al.(1978)Rivest, Shamir, and Adleman]{rivest1978method}
Ronald~L Rivest, Adi Shamir, and Leonard Adleman.
\newblock A method for obtaining digital signatures and public-key cryptosystems.
\newblock \emph{Communications of the ACM}, 21\penalty0 (2):\penalty0 120--126, 1978.

\bibitem[Team(2023)]{Team_2023}
AlphaCode Team.
\newblock Alphacode 2 technical report.
\newblock December 2023.

\bibitem[Yao et~al.(2023)Yao, Yu, Zhao, Shafran, Griffiths, Cao, and Narasimhan]{yao2023tree}
Shunyu Yao, Dian Yu, Jeffrey Zhao, Izhak Shafran, Thomas~L. Griffiths, Yuan Cao, and Karthik Narasimhan.
\newblock Tree of thoughts: Deliberate problem solving with large language models, 2023.

\bibitem[Ouyang et~al.(2022)Ouyang, Wu, Jiang, Almeida, Wainwright, Mishkin, Zhang, Agarwal, Slama, Ray, et~al.]{ouyang2022training}
Long Ouyang, Jeffrey Wu, Xu~Jiang, Diogo Almeida, Carroll Wainwright, Pamela Mishkin, Chong Zhang, Sandhini Agarwal, Katarina Slama, Alex Ray, et~al.
\newblock Training language models to follow instructions with human feedback.
\newblock \emph{Advances in neural information processing systems}, 35:\penalty0 27730--27744, 2022.

\bibitem[Swamy et~al.(2025)Swamy, Choudhury, Sun, Wu, and Bagnell]{swamy2025all}
Gokul Swamy, Sanjiban Choudhury, Wen Sun, Zhiwei~Steven Wu, and J~Andrew Bagnell.
\newblock All roads lead to likelihood: The value of reinforcement learning in fine-tuning.
\newblock \emph{arXiv preprint arXiv:2503.01067}, 2025.

\bibitem[Du et~al.(2024)Du, Mao, and Tenenbaum]{du2024learning}
Yilun Du, Jiayuan Mao, and Joshua~B Tenenbaum.
\newblock Learning iterative reasoning through energy diffusion.
\newblock \emph{arXiv preprint arXiv:2406.11179}, 2024.

\bibitem[West et~al.(2023)West, Lu, Dziri, Brahman, Li, Hwang, Jiang, Fisher, Ravichander, Chandu, et~al.]{west2023generative}
Peter West, Ximing Lu, Nouha Dziri, Faeze Brahman, Linjie Li, Jena~D Hwang, Liwei Jiang, Jillian Fisher, Abhilasha Ravichander, Khyathi Chandu, et~al.
\newblock The generative ai paradox:" what it can create, it may not understand".
\newblock \emph{arXiv preprint arXiv:2311.00059}, 2023.

\bibitem[Stojni{\'c} et~al.(2023)Stojni{\'c}, Gandhi, Yasuda, Lake, and Dillon]{stojnic2023commonsense}
Gala Stojni{\'c}, Kanishk Gandhi, Shannon Yasuda, Brenden~M Lake, and Moira~R Dillon.
\newblock Commonsense psychology in human infants and machines.
\newblock \emph{Cognition}, 235:\penalty0 105406, 2023.

\bibitem[Kambhampati et~al.(2024)Kambhampati, Valmeekam, Guan, Verma, Stechly, Bhambri, Saldyt, and Murthy]{kambhampati2024position}
Subbarao Kambhampati, Karthik Valmeekam, Lin Guan, Mudit Verma, Kaya Stechly, Siddhant Bhambri, Lucas~Paul Saldyt, and Anil~B Murthy.
\newblock Position: Llms can’t plan, but can help planning in llm-modulo frameworks.
\newblock In \emph{Forty-first International Conference on Machine Learning}, 2024.

\bibitem[Wang et~al.(2023)Wang, Wang, Liu, and Qiu]{wang2023energy}
Ze~Wang, Jiang Wang, Zicheng Liu, and Qiang Qiu.
\newblock Energy-inspired self-supervised pretraining for vision models.
\newblock \emph{arXiv preprint arXiv:2302.01384}, 2023.

\bibitem[LeCun(2022)]{lecun2022path}
Yann LeCun.
\newblock A path towards autonomous machine intelligence version 0.9. 2, 2022-06-27.
\newblock \emph{Open Review}, 62, 2022.

\bibitem[Goodfellow et~al.(2014)Goodfellow, Pouget-Abadie, Mirza, Xu, Warde-Farley, Ozair, Courville, and Bengio]{goodfellow2014generative}
Ian~J Goodfellow, Jean Pouget-Abadie, Mehdi Mirza, Bing Xu, David Warde-Farley, Sherjil Ozair, Aaron Courville, and Yoshua Bengio.
\newblock Generative adversarial nets.
\newblock \emph{Advances in neural information processing systems}, 27, 2014.

\bibitem[Bengio et~al.(2003)Bengio, Ducharme, Vincent, and Jauvin]{bengio2003neural}
Yoshua Bengio, R{\'e}jean Ducharme, Pascal Vincent, and Christian Jauvin.
\newblock A neural probabilistic language model.
\newblock \emph{Journal of machine learning research}, 3\penalty0 (Feb):\penalty0 1137--1155, 2003.

\bibitem[Ho et~al.(2020)Ho, Jain, and Abbeel]{ho2020denoising}
Jonathan Ho, Ajay Jain, and Pieter Abbeel.
\newblock Denoising diffusion probabilistic models.
\newblock \emph{Advances in neural information processing systems}, 33:\penalty0 6840--6851, 2020.

\bibitem[Dagréou et~al.(2024)Dagréou, Ablin, Vaiter, and Moreau]{dagréou2024howtocompute}
Mathieu Dagréou, Pierre Ablin, Samuel Vaiter, and Thomas Moreau.
\newblock How to compute hessian-vector products?
\newblock In \emph{ICLR Blogposts 2024}, 2024.
\newblock URL \url{https://iclr-blogposts.github.io/2024/blog/bench-hvp/}.
\newblock https://iclr-blogposts.github.io/2024/blog/bench-hvp/.

\bibitem[Stiennon et~al.(2020)Stiennon, Ouyang, Wu, Ziegler, Lowe, Voss, Radford, Amodei, and Christiano]{stiennon2020learning}
Nisan Stiennon, Long Ouyang, Jeffrey Wu, Daniel Ziegler, Ryan Lowe, Chelsea Voss, Alec Radford, Dario Amodei, and Paul~F Christiano.
\newblock Learning to summarize with human feedback.
\newblock \emph{Advances in neural information processing systems}, 33:\penalty0 3008--3021, 2020.

\bibitem[OpenAI(2023)]{openai2023gpt4}
OpenAI.
\newblock Gpt-4 technical report, 2023.

\bibitem[Oquab et~al.(2023)Oquab, Darcet, Moutakanni, Vo, Szafraniec, Khalidov, Fernandez, Haziza, Massa, El-Nouby, Assran, Ballas, Galuba, Howes, Huang, Li, Misra, Rabbat, Sharma, Synnaeve, Xu, Jegou, Mairal, Labatut, Joulin, and Bojanowski]{oquab2023dinov2}
Maxime Oquab, Timothée Darcet, Théo Moutakanni, Huy Vo, Marc Szafraniec, Vasil Khalidov, Pierre Fernandez, Daniel Haziza, Francisco Massa, Alaaeldin El-Nouby, Mahmoud Assran, Nicolas Ballas, Wojciech Galuba, Russell Howes, Po-Yao Huang, Shang-Wen Li, Ishan Misra, Michael Rabbat, Vasu Sharma, Gabriel Synnaeve, Hu~Xu, Hervé Jegou, Julien Mairal, Patrick Labatut, Armand Joulin, and Piotr Bojanowski.
\newblock Dinov2: Learning robust visual features without supervision, 2023.

\bibitem[He et~al.(2021)He, Chen, Xie, Li, Dollár, and Girshick]{he2021masked}
Kaiming He, Xinlei Chen, Saining Xie, Yanghao Li, Piotr Dollár, and Ross Girshick.
\newblock Masked autoencoders are scalable vision learners, 2021.

\bibitem[Borsos et~al.(2023)Borsos, Marinier, Vincent, Kharitonov, Pietquin, Sharifi, Roblek, Teboul, Grangier, Tagliasacchi, and Zeghidour]{borsos2023audiolm}
Zalán Borsos, Raphaël Marinier, Damien Vincent, Eugene Kharitonov, Olivier Pietquin, Matt Sharifi, Dominik Roblek, Olivier Teboul, David Grangier, Marco Tagliasacchi, and Neil Zeghidour.
\newblock Audiolm: a language modeling approach to audio generation, 2023.

\bibitem[Radford et~al.(2019)Radford, Wu, Child, Luan, Amodei, Sutskever, et~al.]{radford2019language}
Alec Radford, Jeffrey Wu, Rewon Child, David Luan, Dario Amodei, Ilya Sutskever, et~al.
\newblock Language models are unsupervised multitask learners.
\newblock \emph{OpenAI blog}, 1\penalty0 (8):\penalty0 9, 2019.

\bibitem[Vaswani et~al.(2017)Vaswani, Shazeer, Parmar, Uszkoreit, Jones, Gomez, Kaiser, and Polosukhin]{vaswani2017attention}
Ashish Vaswani, Noam Shazeer, Niki Parmar, Jakob Uszkoreit, Llion Jones, Aidan~N Gomez, {\L}ukasz Kaiser, and Illia Polosukhin.
\newblock Attention is all you need.
\newblock \emph{Advances in neural information processing systems}, 30, 2017.

\bibitem[Grattafiori et~al.(2024)Grattafiori, Dubey, Jauhri, Pandey, Kadian, Al-Dahle, Letman, Mathur, Schelten, Vaughan, et~al.]{grattafiori2024llama}
Aaron Grattafiori, Abhimanyu Dubey, Abhinav Jauhri, Abhinav Pandey, Abhishek Kadian, Ahmad Al-Dahle, Aiesha Letman, Akhil Mathur, Alan Schelten, Alex Vaughan, et~al.
\newblock The llama 3 herd of models.
\newblock \emph{arXiv preprint arXiv:2407.21783}, 2024.

\bibitem[Kaplan et~al.(2020)Kaplan, McCandlish, Henighan, Brown, Chess, Child, Gray, Radford, Wu, and Amodei]{kaplan2020scaling}
Jared Kaplan, Sam McCandlish, Tom Henighan, Tom~B Brown, Benjamin Chess, Rewon Child, Scott Gray, Alec Radford, Jeffrey Wu, and Dario Amodei.
\newblock Scaling laws for neural language models.
\newblock \emph{arXiv preprint arXiv:2001.08361}, 2020.

\bibitem[Li et~al.(2025{\natexlab{c}})Li, Kudugunta, and Zettlemoyer]{li2025mis}
Margaret Li, Sneha Kudugunta, and Luke Zettlemoyer.
\newblock (mis) fitting: A survey of scaling laws.
\newblock \emph{arXiv preprint arXiv:2502.18969}, 2025{\natexlab{c}}.

\bibitem[Touvron et~al.(2023)Touvron, Martin, Stone, Albert, Almahairi, Babaei, Bashlykov, Batra, Bhargava, Bhosale, Bikel, Blecher, Ferrer, Chen, Cucurull, Esiobu, Fernandes, Fu, Fu, Fuller, Gao, Goswami, Goyal, Hartshorn, Hosseini, Hou, Inan, Kardas, Kerkez, Khabsa, Kloumann, Korenev, Koura, Lachaux, Lavril, Lee, Liskovich, Lu, Mao, Martinet, Mihaylov, Mishra, Molybog, Nie, Poulton, Reizenstein, Rungta, Saladi, Schelten, Silva, Smith, Subramanian, Tan, Tang, Taylor, Williams, Kuan, Xu, Yan, Zarov, Zhang, Fan, Kambadur, Narang, Rodriguez, Stojnic, Edunov, and Scialom]{touvron2023llama}
Hugo Touvron, Louis Martin, Kevin Stone, Peter Albert, Amjad Almahairi, Yasmine Babaei, Nikolay Bashlykov, Soumya Batra, Prajjwal Bhargava, Shruti Bhosale, Dan Bikel, Lukas Blecher, Cristian~Canton Ferrer, Moya Chen, Guillem Cucurull, David Esiobu, Jude Fernandes, Jeremy Fu, Wenyin Fu, Brian Fuller, Cynthia Gao, Vedanuj Goswami, Naman Goyal, Anthony Hartshorn, Saghar Hosseini, Rui Hou, Hakan Inan, Marcin Kardas, Viktor Kerkez, Madian Khabsa, Isabel Kloumann, Artem Korenev, Punit~Singh Koura, Marie-Anne Lachaux, Thibaut Lavril, Jenya Lee, Diana Liskovich, Yinghai Lu, Yuning Mao, Xavier Martinet, Todor Mihaylov, Pushkar Mishra, Igor Molybog, Yixin Nie, Andrew Poulton, Jeremy Reizenstein, Rashi Rungta, Kalyan Saladi, Alan Schelten, Ruan Silva, Eric~Michael Smith, Ranjan Subramanian, Xiaoqing~Ellen Tan, Binh Tang, Ross Taylor, Adina Williams, Jian~Xiang Kuan, Puxin Xu, Zheng Yan, Iliyan Zarov, Yuchen Zhang, Angela Fan, Melanie Kambadur, Sharan Narang, Aurelien Rodriguez, Robert Stojnic, Sergey Edunov, and Thomas
  Scialom.
\newblock Llama 2: Open foundation and fine-tuned chat models, 2023.

\bibitem[Alabdulmohsin et~al.(2022)Alabdulmohsin, Neyshabur, and Zhai]{alabdulmohsin2022revisiting}
Ibrahim~M Alabdulmohsin, Behnam Neyshabur, and Xiaohua Zhai.
\newblock Revisiting neural scaling laws in language and vision.
\newblock \emph{Advances in Neural Information Processing Systems}, 35:\penalty0 22300--22312, 2022.

\bibitem[Henighan et~al.(2020)Henighan, Kaplan, Katz, Chen, Hesse, Jackson, Jun, Brown, Dhariwal, Gray, et~al.]{henighan2020scaling}
Tom Henighan, Jared Kaplan, Mor Katz, Mark Chen, Christopher Hesse, Jacob Jackson, Heewoo Jun, Tom~B Brown, Prafulla Dhariwal, Scott Gray, et~al.
\newblock Scaling laws for autoregressive generative modeling.
\newblock \emph{arXiv preprint arXiv:2010.14701}, 2020.

\bibitem[Hoffmann et~al.(2022)Hoffmann, Borgeaud, Mensch, Buchatskaya, Cai, Rutherford, Casas, Hendricks, Welbl, Clark, et~al.]{hoffmann2022training}
Jordan Hoffmann, Sebastian Borgeaud, Arthur Mensch, Elena Buchatskaya, Trevor Cai, Eliza Rutherford, Diego de~Las Casas, Lisa~Anne Hendricks, Johannes Welbl, Aidan Clark, et~al.
\newblock Training compute-optimal large language models.
\newblock \emph{arXiv preprint arXiv:2203.15556}, 2022.

\bibitem[Chen et~al.(2018)Chen, Rubanova, Bettencourt, and Duvenaud]{chen2018neural}
Ricky~TQ Chen, Yulia Rubanova, Jesse Bettencourt, and David~K Duvenaud.
\newblock Neural ordinary differential equations.
\newblock \emph{Advances in neural information processing systems}, 31, 2018.

\bibitem[Weber et~al.(2024)Weber, Fu, Anthony, Oren, Adams, Alexandrov, Lyu, Nguyen, Yao, Adams, et~al.]{weber2024redpajama}
Maurice Weber, Dan Fu, Quentin Anthony, Yonatan Oren, Shane Adams, Anton Alexandrov, Xiaozhong Lyu, Huu Nguyen, Xiaozhe Yao, Virginia Adams, et~al.
\newblock Redpajama: an open dataset for training large language models.
\newblock \emph{Advances in neural information processing systems}, 37:\penalty0 116462--116492, 2024.

\bibitem[Computer(2023)]{together2023redpajama}
Together Computer.
\newblock Redpajama: an open dataset for training large language models, 2023.
\newblock URL \url{https://github.com/togethercomputer/RedPajama-Data}.

\bibitem[Black et~al.(2022{\natexlab{a}})Black, Biderman, Hallahan, Anthony, Gao, Golding, He, Leahy, McDonell, Phang, et~al.]{black2022gpt}
Sid Black, Stella Biderman, Eric Hallahan, Quentin Anthony, Leo Gao, Laurence Golding, Horace He, Connor Leahy, Kyle McDonell, Jason Phang, et~al.
\newblock Gpt-neox-20b: An open-source autoregressive language model.
\newblock \emph{arXiv preprint arXiv:2204.06745}, 2022{\natexlab{a}}.

\bibitem[Sun et~al.(2024)Sun, Li, Dalal, Xu, Vikram, Zhang, Dubois, Chen, Wang, Koyejo, et~al.]{sun2024learning}
Yu~Sun, Xinhao Li, Karan Dalal, Jiarui Xu, Arjun Vikram, Genghan Zhang, Yann Dubois, Xinlei Chen, Xiaolong Wang, Sanmi Koyejo, et~al.
\newblock Learning to (learn at test time): Rnns with expressive hidden states.
\newblock \emph{arXiv preprint arXiv:2407.04620}, 2024.

\bibitem[Ye et~al.(2024)Ye, Xu, Li, and Allen-Zhu]{ye2024physics}
Tian Ye, Zicheng Xu, Yuanzhi Li, and Zeyuan Allen-Zhu.
\newblock Physics of language models: Part 2.1, grade-school math and the hidden reasoning process.
\newblock In \emph{The Thirteenth International Conference on Learning Representations}, 2024.

\bibitem[Cobbe et~al.(2021)Cobbe, Kosaraju, Bavarian, Chen, Jun, Kaiser, Plappert, Tworek, Hilton, Nakano, et~al.]{cobbe2021training}
Karl Cobbe, Vineet Kosaraju, Mohammad Bavarian, Mark Chen, Heewoo Jun, Lukasz Kaiser, Matthias Plappert, Jerry Tworek, Jacob Hilton, Reiichiro Nakano, et~al.
\newblock Training verifiers to solve math word problems.
\newblock \emph{arXiv preprint arXiv:2110.14168}, 2021.

\bibitem[Rajpurkar et~al.(2016)Rajpurkar, Zhang, Lopyrev, and Liang]{rajpurkar2016squad}
Pranav Rajpurkar, Jian Zhang, Konstantin Lopyrev, and Percy Liang.
\newblock Squad: 100,000+ questions for machine comprehension of text.
\newblock \emph{arXiv preprint arXiv:1606.05250}, 2016.

\bibitem[Srivastava et~al.(2022)Srivastava, Rastogi, Rao, Shoeb, Abid, Fisch, Brown, Santoro, Gupta, Garriga-Alonso, et~al.]{srivastava2022beyond}
Aarohi Srivastava, Abhinav Rastogi, Abhishek Rao, Abu Awal~Md Shoeb, Abubakar Abid, Adam Fisch, Adam~R Brown, Adam Santoro, Aditya Gupta, Adri{\`a} Garriga-Alonso, et~al.
\newblock Beyond the imitation game: Quantifying and extrapolating the capabilities of language models.
\newblock \emph{arXiv preprint arXiv:2206.04615}, 2022.

\bibitem[Schaeffer et~al.(2023)Schaeffer, Miranda, and Koyejo]{schaeffer2023emergent}
Rylan Schaeffer, Brando Miranda, and Sanmi Koyejo.
\newblock Are emergent abilities of large language models a mirage?
\newblock \emph{Advances in Neural Information Processing Systems}, 36:\penalty0 55565--55581, 2023.

\bibitem[Pagnoni et~al.(2024)Pagnoni, Pasunuru, Rodriguez, Nguyen, Muller, Li, Zhou, Yu, Weston, Zettlemoyer, et~al.]{pagnoni2024byte}
Artidoro Pagnoni, Ram Pasunuru, Pedro Rodriguez, John Nguyen, Benjamin Muller, Margaret Li, Chunting Zhou, Lili Yu, Jason Weston, Luke Zettlemoyer, et~al.
\newblock Byte latent transformer: Patches scale better than tokens.
\newblock \emph{arXiv preprint arXiv:2412.09871}, 2024.

\bibitem[Gadre et~al.(2024)Gadre, Smyrnis, Shankar, Gururangan, Wortsman, Shao, Mercat, Fang, Li, Keh, et~al.]{gadre2024language}
Samir~Yitzhak Gadre, Georgios Smyrnis, Vaishaal Shankar, Suchin Gururangan, Mitchell Wortsman, Rulin Shao, Jean Mercat, Alex Fang, Jeffrey Li, Sedrick Keh, et~al.
\newblock Language models scale reliably with over-training and on downstream tasks.
\newblock \emph{arXiv preprint arXiv:2403.08540}, 2024.

\bibitem[Gururangan et~al.(2020)Gururangan, Marasovi{\'c}, Swayamdipta, Lo, Beltagy, Downey, and Smith]{gururangan2020don}
Suchin Gururangan, Ana Marasovi{\'c}, Swabha Swayamdipta, Kyle Lo, Iz~Beltagy, Doug Downey, and Noah~A Smith.
\newblock Don't stop pretraining: Adapt language models to domains and tasks.
\newblock \emph{arXiv preprint arXiv:2004.10964}, 2020.

\bibitem[Thrush et~al.(2024)Thrush, Potts, and Hashimoto]{thrush2024improving}
Tristan Thrush, Christopher Potts, and Tatsunori Hashimoto.
\newblock Improving pretraining data using perplexity correlations.
\newblock \emph{arXiv preprint arXiv:2409.05816}, 2024.

\bibitem[Chen et~al.(2024)Chen, Huang, Gao, Wang, Yang, and Ji]{chen2024scaling}
Yangyi Chen, Binxuan Huang, Yifan Gao, Zhengyang Wang, Jingfeng Yang, and Heng Ji.
\newblock Scaling laws for predicting downstream performance in llms.
\newblock \emph{arXiv preprint arXiv:2410.08527}, 2024.

\bibitem[Isik et~al.(2024)Isik, Ponomareva, Hazimeh, Paparas, Vassilvitskii, and Koyejo]{isik2024scaling}
Berivan Isik, Natalia Ponomareva, Hussein Hazimeh, Dimitris Paparas, Sergei Vassilvitskii, and Sanmi Koyejo.
\newblock Scaling laws for downstream task performance of large language models.
\newblock In \emph{ICLR 2024 Workshop on Mathematical and Empirical Understanding of Foundation Models}, 2024.

\bibitem[Deng et~al.(2024{\natexlab{b}})Deng, Pan, Diao, Luo, Cui, Lu, Shan, Qi, and Wang]{deng2024autoregressive}
Haoge Deng, Ting Pan, Haiwen Diao, Zhengxiong Luo, Yufeng Cui, Huchuan Lu, Shiguang Shan, Yonggang Qi, and Xinlong Wang.
\newblock Autoregressive video generation without vector quantization.
\newblock \emph{arXiv preprint arXiv:2412.14169}, 2024{\natexlab{b}}.

\bibitem[Weissenborn et~al.(2019)Weissenborn, T{\"a}ckstr{\"o}m, and Uszkoreit]{weissenborn2019scaling}
Dirk Weissenborn, Oscar T{\"a}ckstr{\"o}m, and Jakob Uszkoreit.
\newblock Scaling autoregressive video models.
\newblock \emph{arXiv preprint arXiv:1906.02634}, 2019.

\bibitem[Rakhimov et~al.(2020)Rakhimov, Volkhonskiy, Artemov, Zorin, and Burnaev]{rakhimov2020latent}
Ruslan Rakhimov, Denis Volkhonskiy, Alexey Artemov, Denis Zorin, and Evgeny Burnaev.
\newblock Latent video transformer.
\newblock \emph{arXiv preprint arXiv:2006.10704}, 2020.

\bibitem[Ye and Bilodeau(2023)]{ye2023video}
Xi~Ye and Guillaume-Alexandre Bilodeau.
\newblock Video prediction by efficient transformers.
\newblock \emph{Image and Vision Computing}, 130:\penalty0 104612, 2023.

\bibitem[Gu et~al.(2025)Gu, Mao, and Shou]{gu2025long}
Yuchao Gu, Weijia Mao, and Mike~Zheng Shou.
\newblock Long-context autoregressive video modeling with next-frame prediction.
\newblock \emph{arXiv preprint arXiv:2503.19325}, 2025.

\bibitem[Villalobos et~al.(2022)Villalobos, Sevilla, Heim, Besiroglu, Hobbhahn, and Ho]{villalobos2022will}
Pablo Villalobos, Jaime Sevilla, Lennart Heim, Tamay Besiroglu, Marius Hobbhahn, and Anson Ho.
\newblock Will we run out of data? an analysis of the limits of scaling datasets in machine learning.
\newblock \emph{arXiv preprint arXiv:2211.04325}, 2022.

\bibitem[Dat()]{Data-centricML-benchmarking}
URL \url{https://research.google/blog/data-centric-ml-benchmarking-announcing-dataperfs-2023-challenges/}.

\bibitem[Rombach et~al.(2022)Rombach, Blattmann, Lorenz, Esser, and Ommer]{rombach2022high}
Robin Rombach, Andreas Blattmann, Dominik Lorenz, Patrick Esser, and Bj{\"o}rn Ommer.
\newblock High-resolution image synthesis with latent diffusion models.
\newblock In \emph{Proceedings of the IEEE/CVF conference on computer vision and pattern recognition}, pages 10684--10695, 2022.

\bibitem[sta()]{stabilityai/sd-vae-ft-mse}
URL \url{https://huggingface.co/stabilityai/sd-vae-ft-mse}.

\bibitem[Goyal et~al.(2017)Goyal, Ebrahimi~Kahou, Michalski, Materzynska, Westphal, Kim, Haenel, Fruend, Yianilos, Mueller-Freitag, et~al.]{goyal2017something}
Raghav Goyal, Samira Ebrahimi~Kahou, Vincent Michalski, Joanna Materzynska, Susanne Westphal, Heuna Kim, Valentin Haenel, Ingo Fruend, Peter Yianilos, Moritz Mueller-Freitag, et~al.
\newblock The" something something" video database for learning and evaluating visual common sense.
\newblock In \emph{Proceedings of the IEEE international conference on computer vision}, pages 5842--5850, 2017.

\bibitem[Islam et~al.(2023)Islam, Gladstone, Islam, and Iqbal]{islam2023eqa}
Md~Mofijul Islam, Alexi Gladstone, Riashat Islam, and Tariq Iqbal.
\newblock Eqa-mx: Embodied question answering using multimodal expression.
\newblock In \emph{The Twelfth International Conference on Learning Representations}, 2023.

\bibitem[Chen et~al.(2020)Chen, Dai, Li, Gao, and Song]{chen2020learning}
Xinshi Chen, Hanjun Dai, Yu~Li, Xin Gao, and Le~Song.
\newblock Learning to stop while learning to predict.
\newblock In \emph{International conference on machine learning}, pages 1520--1530. PMLR, 2020.

\bibitem[Lin et~al.(2014)Lin, Maire, Belongie, Hays, Perona, Ramanan, Doll{\'a}r, and Zitnick]{lin2014microsoft}
Tsung-Yi Lin, Michael Maire, Serge Belongie, James Hays, Pietro Perona, Deva Ramanan, Piotr Doll{\'a}r, and C~Lawrence Zitnick.
\newblock Microsoft coco: Common objects in context.
\newblock In \emph{Computer vision--ECCV 2014: 13th European conference, zurich, Switzerland, September 6-12, 2014, proceedings, part v 13}, pages 740--755. Springer, 2014.

\bibitem[Abd()]{AbdoTW/COCO_2014DatasetsHF}
URL \url{https://huggingface.co/datasets/AbdoTW/COCO_2014}.

\bibitem[Pei et~al.(2022)Pei, Wang, and Szarvas]{pei2022transformer}
Jiahuan Pei, Cheng Wang, and Gy{\"o}rgy Szarvas.
\newblock Transformer uncertainty estimation with hierarchical stochastic attention.
\newblock In \emph{Proceedings of the AAAI Conference on Artificial Intelligence}, volume~36, pages 11147--11155, 2022.

\bibitem[Russakovsky et~al.(2015)Russakovsky, Deng, Su, Krause, Satheesh, Ma, Huang, Karpathy, Khosla, Bernstein, et~al.]{russakovsky2015imagenet}
Olga Russakovsky, Jia Deng, Hao Su, Jonathan Krause, Sanjeev Satheesh, Sean Ma, Zhiheng Huang, Andrej Karpathy, Aditya Khosla, Michael Bernstein, et~al.
\newblock Imagenet large scale visual recognition challenge.
\newblock \emph{International journal of computer vision}, 115:\penalty0 211--252, 2015.

\bibitem[ope()]{openaiYouTubegpt4_5}
URL \url{https://www.youtube.com/watch?v=6nJZopACRuQ&ab_channel=OpenAI}.

\bibitem[Huang et~al.(2025)Huang, Block, Liu, Jiang, Foster, and Krishnamurthy]{huang2025best}
Audrey Huang, Adam Block, Qinghua Liu, Nan Jiang, Dylan~J Foster, and Akshay Krishnamurthy.
\newblock Is best-of-n the best of them? coverage, scaling, and optimality in inference-time alignment.
\newblock \emph{arXiv preprint arXiv:2503.21878}, 2025.

\bibitem[Latif et~al.(2023)Latif, Zaidi, Cuayahuitl, Shamshad, Shoukat, and Qadir]{latif2023transformers}
Siddique Latif, Aun Zaidi, Heriberto Cuayahuitl, Fahad Shamshad, Moazzam Shoukat, and Junaid Qadir.
\newblock Transformers in speech processing: A survey.
\newblock \emph{arXiv preprint arXiv:2303.11607}, 2023.

\bibitem[Dehghani et~al.(2018)Dehghani, Gouws, Vinyals, Uszkoreit, and Kaiser]{dehghani2018universal}
Mostafa Dehghani, Stephan Gouws, Oriol Vinyals, Jakob Uszkoreit, and {\L}ukasz Kaiser.
\newblock Universal transformers.
\newblock \emph{arXiv preprint arXiv:1807.03819}, 2018.

\bibitem[Saunshi et~al.(2025)Saunshi, Dikkala, Li, Kumar, and Reddi]{saunshi2025reasoning}
Nikunj Saunshi, Nishanth Dikkala, Zhiyuan Li, Sanjiv Kumar, and Sashank~J Reddi.
\newblock Reasoning with latent thoughts: On the power of looped transformers.
\newblock \emph{arXiv preprint arXiv:2502.17416}, 2025.

\bibitem[Wei et~al.(2022)Wei, Wang, Schuurmans, Bosma, Xia, Chi, Le, Zhou, et~al.]{wei2022chain}
Jason Wei, Xuezhi Wang, Dale Schuurmans, Maarten Bosma, Fei Xia, Ed~Chi, Quoc~V Le, Denny Zhou, et~al.
\newblock Chain-of-thought prompting elicits reasoning in large language models.
\newblock \emph{Advances in Neural Information Processing Systems}, 35:\penalty0 24824--24837, 2022.

\bibitem[Lin et~al.(2025)Lin, Xie, Yuan, and Yang]{lin2025implicit}
Tianhe Lin, Jian Xie, Siyu Yuan, and Deqing Yang.
\newblock Implicit reasoning in transformers is reasoning through shortcuts.
\newblock \emph{arXiv preprint arXiv:2503.07604}, 2025.

\bibitem[Turpin et~al.(2023)Turpin, Michael, Perez, and Bowman]{turpin2023language}
Miles Turpin, Julian Michael, Ethan Perez, and Samuel Bowman.
\newblock Language models don't always say what they think: Unfaithful explanations in chain-of-thought prompting.
\newblock \emph{Advances in Neural Information Processing Systems}, 36:\penalty0 74952--74965, 2023.

\bibitem[Agarwal et~al.(2024)Agarwal, Tanneru, and Lakkaraju]{agarwal2024faithfulness}
Chirag Agarwal, Sree~Harsha Tanneru, and Himabindu Lakkaraju.
\newblock Faithfulness vs. plausibility: On the (un) reliability of explanations from large language models.
\newblock \emph{arXiv preprint arXiv:2402.04614}, 2024.

\bibitem[Höppe et~al.(2022)Höppe, Mehrjou, Bauer, Nielsen, and Dittadi]{höppe2022diffusion}
Tobias Höppe, Arash Mehrjou, Stefan Bauer, Didrik Nielsen, and Andrea Dittadi.
\newblock Diffusion models for video prediction and infilling, 2022.

\bibitem[Du et~al.(2023)Du, Durkan, Strudel, Tenenbaum, Dieleman, Fergus, Sohl-Dickstein, Doucet, and Grathwohl]{du2023reduce}
Yilun Du, Conor Durkan, Robin Strudel, Joshua~B Tenenbaum, Sander Dieleman, Rob Fergus, Jascha Sohl-Dickstein, Arnaud Doucet, and Will~Sussman Grathwohl.
\newblock Reduce, reuse, recycle: Compositional generation with energy-based diffusion models and mcmc.
\newblock In \emph{International conference on machine learning}, pages 8489--8510. PMLR, 2023.

\bibitem[LeCun et~al.(2006)LeCun, Chopra, Hadsell, Ranzato, Huang, et~al.]{lecun2006tutorial}
Yann LeCun, Sumit Chopra, Raia Hadsell, M~Ranzato, Fujie Huang, et~al.
\newblock A tutorial on energy-based learning.
\newblock \emph{Predicting structured data}, 1\penalty0 (0), 2006.

\bibitem[Berglund et~al.(2023)Berglund, Tong, Kaufmann, Balesni, Stickland, Korbak, and Evans]{berglund2023reversal}
Lukas Berglund, Meg Tong, Max Kaufmann, Mikita Balesni, Asa~Cooper Stickland, Tomasz Korbak, and Owain Evans.
\newblock The reversal curse: Llms trained on" a is b" fail to learn" b is a".
\newblock \emph{arXiv preprint arXiv:2309.12288}, 2023.

\bibitem[Janner et~al.(2022)Janner, Du, Tenenbaum, and Levine]{janner2022planning}
Michael Janner, Yilun Du, Joshua~B Tenenbaum, and Sergey Levine.
\newblock Planning with diffusion for flexible behavior synthesis.
\newblock \emph{arXiv preprint arXiv:2205.09991}, 2022.

\bibitem[Chi et~al.(2023)Chi, Xu, Feng, Cousineau, Du, Burchfiel, Tedrake, and Song]{chi2023diffusion}
Cheng Chi, Zhenjia Xu, Siyuan Feng, Eric Cousineau, Yilun Du, Benjamin Burchfiel, Russ Tedrake, and Shuran Song.
\newblock Diffusion policy: Visuomotor policy learning via action diffusion.
\newblock \emph{The International Journal of Robotics Research}, page 02783649241273668, 2023.

\bibitem[Zhou et~al.(2024)Zhou, Pan, LeCun, and Pinto]{zhou2024dino}
Gaoyue Zhou, Hengkai Pan, Yann LeCun, and Lerrel Pinto.
\newblock Dino-wm: World models on pre-trained visual features enable zero-shot planning.
\newblock \emph{arXiv preprint arXiv:2411.04983}, 2024.

\bibitem[Bakhtin et~al.(2021)Bakhtin, Deng, Gross, Ott, Ranzato, and Szlam]{bakhtin2021residual}
Anton Bakhtin, Yuntian Deng, Sam Gross, Myle Ott, Marc'Aurelio Ranzato, and Arthur Szlam.
\newblock Residual energy-based models for text.
\newblock \emph{Journal of Machine Learning Research}, 22\penalty0 (40):\penalty0 1--41, 2021.

\bibitem[Bhattacharyya et~al.(2020)Bhattacharyya, Rooshenas, Naskar, Sun, Iyyer, and McCallum]{bhattacharyya2020energy}
Sumanta Bhattacharyya, Amirmohammad Rooshenas, Subhajit Naskar, Simeng Sun, Mohit Iyyer, and Andrew McCallum.
\newblock Energy-based reranking: Improving neural machine translation using energy-based models.
\newblock \emph{arXiv preprint arXiv:2009.13267}, 2020.

\bibitem[Douglas and Martin(2007)]{douglas2007recurrent}
Rodney~J Douglas and Kevan~AC Martin.
\newblock Recurrent neuronal circuits in the neocortex.
\newblock \emph{Current biology}, 17\penalty0 (13):\penalty0 R496--R500, 2007.

\bibitem[Betancourt(2017)]{betancourt2017conceptual}
Michael Betancourt.
\newblock A conceptual introduction to hamiltonian monte carlo.
\newblock \emph{arXiv preprint arXiv:1701.02434}, 2017.

\bibitem[Penedo et~al.(2024)Penedo, Kydl{\'\i}{\v{c}}ek, Lozhkov, Mitchell, Raffel, Von~Werra, Wolf, et~al.]{penedo2024fineweb}
Guilherme Penedo, Hynek Kydl{\'\i}{\v{c}}ek, Anton Lozhkov, Margaret Mitchell, Colin~A Raffel, Leandro Von~Werra, Thomas Wolf, et~al.
\newblock The fineweb datasets: Decanting the web for the finest text data at scale.
\newblock \emph{Advances in Neural Information Processing Systems}, 37:\penalty0 30811--30849, 2024.

\bibitem[Manvi et~al.(2024)Manvi, Singh, and Ermon]{manvi2024adaptive}
Rohin Manvi, Anikait Singh, and Stefano Ermon.
\newblock Adaptive inference-time compute: Llms can predict if they can do better, even mid-generation.
\newblock \emph{arXiv preprint arXiv:2410.02725}, 2024.

\bibitem[Snell et~al.(2024)Snell, Lee, Xu, and Kumar]{snell2024scaling}
Charlie Snell, Jaehoon Lee, Kelvin Xu, and Aviral Kumar.
\newblock Scaling llm test-time compute optimally can be more effective than scaling model parameters.
\newblock \emph{arXiv preprint arXiv:2408.03314}, 2024.

\bibitem[Parr et~al.(2022)Parr, Pezzulo, and Friston]{parr2022active}
Thomas Parr, Giovanni Pezzulo, and Karl~J Friston.
\newblock \emph{Active inference: the free energy principle in mind, brain, and behavior}.
\newblock MIT Press, 2022.

\bibitem[Falcon(2019)]{falcon2019pytorch}
William~A Falcon.
\newblock Pytorch lightning.
\newblock \emph{GitHub}, 3, 2019.

\bibitem[Black et~al.(2022{\natexlab{b}})Black, Biderman, Hallahan, Anthony, Gao, Golding, He, Leahy, McDonell, Phang, Pieler, Prashanth, Purohit, Reynolds, Tow, Wang, and Weinbach]{black2022gptneox20b}
Sid Black, Stella Biderman, Eric Hallahan, Quentin Anthony, Leo Gao, Laurence Golding, Horace He, Connor Leahy, Kyle McDonell, Jason Phang, Michael Pieler, USVSN~Sai Prashanth, Shivanshu Purohit, Laria Reynolds, Jonathan Tow, Ben Wang, and Samuel Weinbach.
\newblock Gpt-neox-20b: An open-source autoregressive language model, 2022{\natexlab{b}}.

\bibitem[Hu et~al.(2023)Hu, Liu, Han, Zhang, He, Zhao, Lin, Ding, Ou, Zeng, et~al.]{hu2023predicting}
Shengding Hu, Xin Liu, Xu~Han, Xinrong Zhang, Chaoqun He, Weilin Zhao, Yankai Lin, Ning Ding, Zebin Ou, Guoyang Zeng, et~al.
\newblock Predicting emergent abilities with infinite resolution evaluation.
\newblock \emph{arXiv preprint arXiv:2310.03262}, 2023.

\bibitem[Casson(2023)]{casson2023transformerflops}
Adam Casson.
\newblock Transformer flops.
\newblock 2023.
\newblock URL \url{https://adamcasson.com/posts/transformer-flops}.

\bibitem[Liu et~al.(2022)Liu, Li, Du, Torralba, and Tenenbaum]{liu2022compositional}
Nan Liu, Shuang Li, Yilun Du, Antonio Torralba, and Joshua~B Tenenbaum.
\newblock Compositional visual generation with composable diffusion models.
\newblock In \emph{European Conference on Computer Vision}, pages 423--439. Springer, 2022.

\bibitem[Hoover et~al.(2024)Hoover, Liang, Pham, Panda, Strobelt, Chau, Zaki, and Krotov]{hoover2024energy}
Benjamin Hoover, Yuchen Liang, Bao Pham, Rameswar Panda, Hendrik Strobelt, Duen~Horng Chau, Mohammed Zaki, and Dmitry Krotov.
\newblock Energy transformer.
\newblock \emph{Advances in Neural Information Processing Systems}, 36, 2024.

\bibitem[Wang et~al.(2022)Wang, Che, Li, Song, Pei, Bengio, and Li]{wang2022your}
Yezhen Wang, Tong Che, Bo~Li, Kaitao Song, Hengzhi Pei, Yoshua Bengio, and Dongsheng Li.
\newblock Your autoregressive generative model can be better if you treat it as an energy-based one.
\newblock \emph{arXiv preprint arXiv:2206.12840}, 2022.

\bibitem[Vincent(2011)]{vincent2011connection}
Pascal Vincent.
\newblock A connection between score matching and denoising autoencoders.
\newblock \emph{Neural computation}, 23\penalty0 (7):\penalty0 1661--1674, 2011.

\bibitem[Friederici and Weissenborn(2007)]{friederici2007mapping}
Angela~D Friederici and J{\"u}rgen Weissenborn.
\newblock Mapping sentence form onto meaning: The syntax--semantic interface.
\newblock \emph{Brain research}, 1146:\penalty0 50--58, 2007.

\bibitem[Kobayashi et~al.(2024)Kobayashi, Schug, Akram, Redhardt, von Oswald, Pascanu, Lajoie, and Sacramento]{kobayashi2024can}
Seijin Kobayashi, Simon Schug, Yassir Akram, Florian Redhardt, Johannes von Oswald, Razvan Pascanu, Guillaume Lajoie, and Jo{\~a}o Sacramento.
\newblock When can transformers compositionally generalize in-context?
\newblock \emph{arXiv preprint arXiv:2407.12275}, 2024.

\bibitem[Huang et~al.(2023)Huang, Sun, Xie, Li, and Liu]{huang2023t2i}
Kaiyi Huang, Kaiyue Sun, Enze Xie, Zhenguo Li, and Xihui Liu.
\newblock T2i-compbench: A comprehensive benchmark for open-world compositional text-to-image generation.
\newblock \emph{Advances in Neural Information Processing Systems}, 36:\penalty0 78723--78747, 2023.

\bibitem[Hopfield(1982)]{hopfield1982neural}
John~J Hopfield.
\newblock Neural networks and physical systems with emergent collective computational abilities.
\newblock \emph{Proceedings of the national academy of sciences}, 79\penalty0 (8):\penalty0 2554--2558, 1982.

\bibitem[Glorot and Bengio(2010)]{glorot2010understanding}
Xavier Glorot and Yoshua Bengio.
\newblock Understanding the difficulty of training deep feedforward neural networks.
\newblock In \emph{Proceedings of the thirteenth international conference on artificial intelligence and statistics}, pages 249--256. JMLR Workshop and Conference Proceedings, 2010.

\bibitem[Shazeer(2020)]{shazeer2020glu}
Noam Shazeer.
\newblock Glu variants improve transformer, 2020.

\bibitem[Su et~al.(2024)Su, Ahmed, Lu, Pan, Bo, and Liu]{su2024roformer}
Jianlin Su, Murtadha Ahmed, Yu~Lu, Shengfeng Pan, Wen Bo, and Yunfeng Liu.
\newblock Roformer: Enhanced transformer with rotary position embedding.
\newblock \emph{Neurocomputing}, 568:\penalty0 127063, 2024.

\end{thebibliography}
